\documentclass[11pt]{article}

\usepackage[preprint]{acl}

\usepackage{times}
\usepackage{latexsym}

\usepackage[T1]{fontenc}

\usepackage[utf8]{inputenc}

\usepackage{microtype}
\usepackage{amsmath,amssymb}
\usepackage{inconsolata}
\usepackage{float}
\usepackage{graphicx}
\usepackage{booktabs}
\usepackage{subcaption}
\usepackage{hyperref}
\usepackage{pifont}
\newcommand{\cmark}{\ding{51}}

\title{Unveiling the Reasoning Process of Large Language Models}

\author{
 \textbf{Junjie Zhang\textsuperscript{1,2}},
 \textbf{Zhen Shen\textsuperscript{1,2}},
 \textbf{Xisong Dong\textsuperscript{1,2}},
 \textbf{Gang Xiong\textsuperscript{1,2}}
\\
 \textsuperscript{1}Institute of Automation, Chinese Academy of Sciences,
 \\
 \textsuperscript{2}School of Artificial Intelligence, University of Chinese Academy of Sciences
\\
 \small{
   \textbf{Correspondence:} \href{mailto:email@domain}{Zhen.Shen@ia.ac.cn}
 }
}

\begin{document}
\maketitle

\begin{abstract}
Large language models often reason beyond surface tokens, but the internal stage at which token-level information becomes abstract relational structure remains unclear. We investigate this question by analyzing how attention heads and layers transform information during autoregressive reasoning. Across mathematical and symbolic reasoning tasks, we observe a consistent layer-wise division of labor: outer layers mainly preserve and route input-related features, whereas middle layers reorganize them into more transferable rule-level representations. This interpretation is supported by representation geometry: middle-layer states occupy lower-dimensional manifolds and show stronger alignment across disjoint vocabularies that instantiate the same symbolic rules. It is further supported by causal interventions: removing middle-layer components identified by our interaction-based criterion produces substantially larger downstream changes and accuracy drops than removing components from other regions or at random. Together, these results suggest that abstract reasoning is not uniformly distributed across transformer layers, but is preferentially formed in a middle-layer computation stage that converts token-level information into reusable relational structure.
\end{abstract}

\section{Introduction}
\label{Introduction}

Large Language Models (LLMs) have developed rapidly in recent years \citep{gemini-2.5,deepseek,llama3,qwen3}, and have shown strong abilities in zero-shot generalization and abstract reasoning \citep{language-is-learner}. However, a central question remains: how are such abstract capabilities organized inside the model \citep{emergent_abilities}? In particular, it is still unclear whether abstraction-related computations are distributed uniformly across transformer layers, or whether they are preferentially formed in specific stages of the network. Understanding this internal organization is important for explaining how LLMs transform token-level information into reasoning-relevant relational structure.


Several lines of prior work suggest that LLMs are not homogeneous information-processing systems. Studies comparing LLM representations with biological or cognitive systems have reported structural and functional correspondences \citep{geometry,lossless_by}, and related analyses have used LLM mechanisms to motivate new perspectives on neural computation and abstraction \citep{lossless_by,emergent_symbolic,domain}. At the same time, other work shows that current models contain substantial redundancy \citep{weight_sparse}, and that simply increasing depth or parameter scale does not necessarily yield more efficient high-level feature composition or multi-step reasoning \citep{do_language}. More broadly, the difficulty of continual adaptation and forgetting in artificial systems has often been discussed through analogy with memory formation and disruption in biological systems \citep{anterograde_amensia}. These results motivate a more specific mechanistic question: which components mainly preserve and route existing information, and which components transform it into abstract representations that can support reasoning?

In this work, inspired by Shannon information theory \citep{shannon1948mathematical} and prior interaction-based analyses of information integration in LLMs \citep{brain-like}, we view an LLM as a multi-node information-processing system. We use information reorganization analysis \citep{IID} to characterize how attention heads interact during autoregressive computation. This allows us to distinguish components that mainly preserve information already available in individual parts from components that combine information across multiple parts. We use this distinction only as an operational criterion for analyzing transformer computation. Similarly, when discussing robustness under intervention, we use the notion of plasticity of neuron in human brain, referring to whether the model can tolerate the removal of certain components without severe performance degradation \citep{plasticity}.

We focus on advanced open-source models \citep{qwen3,llama3,gemma3technicalreport}, with Qwen3 as the main case study. Following prior work on symbolic and mathematical reasoning in LLMs \citep{emergent_symbolic,do_language}, we mainly analyze tasks that require multi-step calculation, rule abstraction, and transfer across surface forms. Our results reveal a consistent layer-wise organization: outer layers are more strongly associated with preserving and routing input-related features, whereas middle layers show stronger evidence of information composition and representation transformation. We further show that this middle computation stage aligns with low-dimensional, vocabulary-invariant rule representations, and that interventions on these components produce larger downstream changes and accuracy degradation.

Our contributions are threefold:

\begin{itemize}
    \item We provide a layer-wise analysis of how LLMs components preserve, route, and compose information during autoregressive reasoning.

    \item We show that abstraction-related signals concentrate in middle layers, where token-level information is transformed into lower-dimensional and more vocabulary-invariant rule representations.

    \item We validate this interpretation through causal interventions at both the layer and attention-head levels, and summarize the same qualitative pattern across multiple model families.
\end{itemize}

Together, these findings suggest that abstract reasoning in LLMs is not uniformly distributed across all layers. Instead, it is preferentially formed in a middle computation stage that converts token-level information into reusable relational structure for later prediction.

\begin{figure*}[t]
\centering

\begin{subfigure}[t]{\linewidth}
    \centering
    \includegraphics[width=\linewidth]{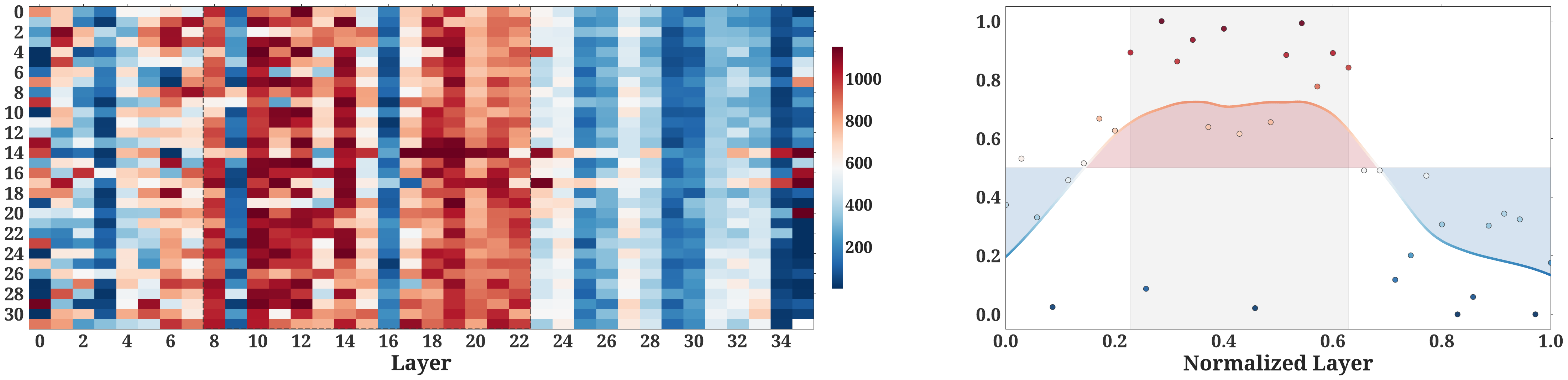}
    \caption{Distribution of composition- and preservation-oriented heads}
    \label{fig:synergy_redundancy_a}
\end{subfigure}

\vspace{1em}

\begin{subfigure}[t]{\linewidth}
    \centering
    \includegraphics[width=\linewidth]{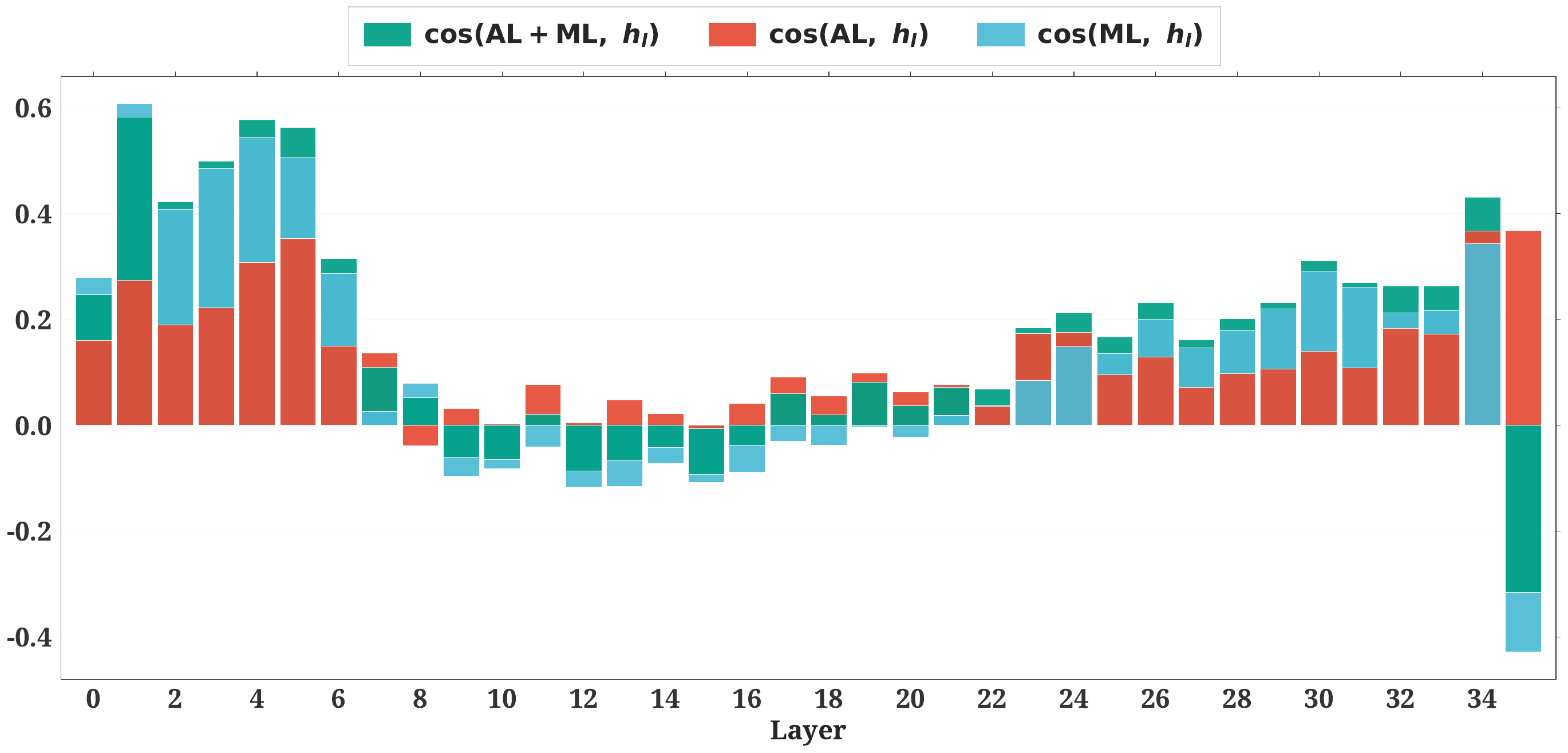}
    \caption{Cosine similarity of layer contributions}
    \label{fig:synergy_redundancy_b}
\end{subfigure}

\caption{Layer-wise localization of composition- and preservation-oriented components in Qwen3-8B-Base.
(a) Attention heads are ranked by the difference between interaction-based composition and preservation scores. Warmer colors indicate heads that more strongly combine information across components, whereas cooler colors indicate heads that more strongly preserve information. The highest concentration of composition-oriented heads appears in middle layers.
(b) Cosine similarity of residual updates across layers. Outer layers tend to reinforce existing directions in the residual stream, whereas middle layers introduce more orthogonal or suppressive updates, consistent with a stronger role in representation transformation.}
\label{fig:synergy_redundancy}
\end{figure*}

\begin{figure}[t]
    \centering

    \begin{subfigure}[t]{\columnwidth}
        \centering
        \includegraphics[width=\linewidth]{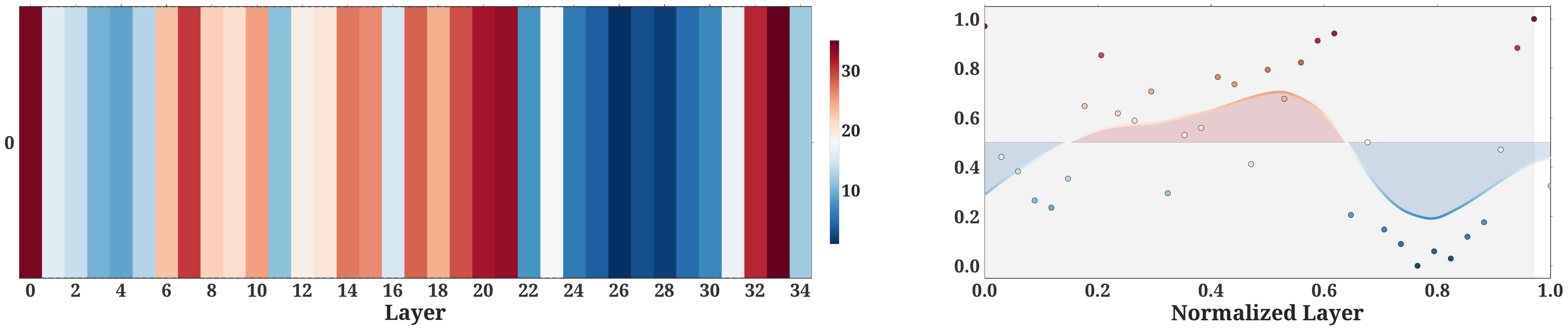}
        \caption{$ML$}
        \label{fig:layer_info_richness_a}
    \end{subfigure}
    
    \vspace{0.5em}
    
    \begin{subfigure}[t]{\columnwidth}
        \centering
        \includegraphics[width=\linewidth]{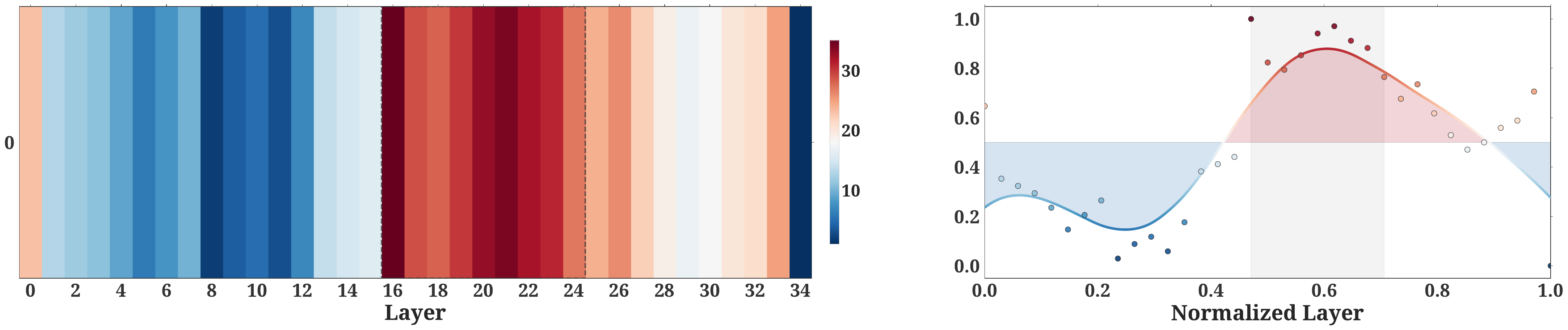}
        \caption{$ML+AL$}
        \label{fig:layer_info_richness_b}
    \end{subfigure}

    \caption{
    Layer-wise composition and preservation scores under different interaction settings.
    (a) Scores computed from the MLP contribution alone. In this setting, the layer-wise pattern shows a gradual transition, with preservation-oriented signals appearing more prominently in earlier layers and composition-oriented signals becoming stronger toward later intermediate layers.
    (b) Scores computed from the combined MLP and attention contributions. Adding attention interactions sharpens the middle-layer composition pattern, with the strongest composition-oriented region appearing around the intermediate-to-late middle layers.
    Together, these results suggest that the composition-preservation organization depends on the type of sublayer interaction being measured, while still supporting the broader observation that representation transformation is concentrated away from the outermost layers.
    }
    
    \label{fig:layer_info_richness}
\end{figure}

\section{Related Work}

Research on the internal mechanisms and generalization abilities of LLMs has shown that model behavior is supported by structured computations rather than uniformly distributed representations. Prior work has found that neurons and attention heads can organize into specialized units that contribute to task execution \citep{towards_representation_learning,emergent_analogical_reasoning,inductive_heads,language_modeling_is_compression}. In particular, studies of emergent symbolic mechanisms suggest that transformer models can develop internal structures that support abstraction and rule-like computation \citep{emergent_symbolic}. Related work on grokking in modular arithmetic has provided a controlled setting for studying how such internal structure emerges during learning \citep{grokking_implicit_reasoning}. Subsequent analyses further indicate that distinct circuits for memory and abstraction can appear within models during these tasks \citep{explaining_circuit_grokking,language_model_circuit}. These findings motivate our focus on whether abstraction-related computation can be localized within particular regions of large pretrained models.

A complementary line of work studies redundancy in model depth and width. Recent analyses suggest that current transformer architectures contain substantial functional redundancy, and that increasing depth does not always lead to more efficient use of intermediate computation \citep{do_language,nested_learning}. Sparse-model studies also show that many parameters can be removed while preserving meaningful task circuits, suggesting that redundancy may coexist with specialized computational structure \citep{weight_sparse}. These results raise an important question for mechanistic analysis: if many components are redundant, which components are actually responsible for transforming information into representations useful for reasoning?

Other studies have examined how individual units or learned features support reusable computation. Work inspired by neuroscience and representation analysis has identified phenomena such as cross-region-like communication patterns and reusable feature associations in LLMs \citep{nested_learning}. Sparse autoencoder studies further show that individual features or feature groups can support task-specific functions such as abstraction, induction, and retrieval \citep{sae_circuit,gemma_sae,igated_sae,rethinking_sae}. These results suggest that computational roles in LLMs are differentiated rather than homogeneous. However, it remains less clear how such differentiated components are organized across layers during abstract reasoning.

Layer-level interventions provide another perspective on this question. Prior work has shown that transformer layers can often be skipped, folded, or merged with limited impact on many tasks \citep{unreasonable_ineffectiveness,ransformer_as_painter,remarkable}. At the same time, reasoning-intensive benchmarks such as GSM8K and MATH are more sensitive to interventions that disrupt multi-step computation \citep{gsm8k,MATH}. This suggests that abstract reasoning may rely on specific intermediate transformations rather than on all layers equally. Our work builds on this observation by analyzing how information-preserving and information-composing components are distributed across layers, and by testing whether the identified middle-stage components align with rule-level representations and causal sensitivity under intervention.

\section{Preliminaries}

\subsection{Information Decomposition}

We briefly introduce the information-theoretic quantities used in our analysis. For two random variables $X$ and $Y$, mutual information measures the statistical dependence between them:
\begin{equation}
I(X;Y)=
\sum_{x\in X,y\in Y}
p(x,y)\log\frac{p(x,y)}{p(x)p(y)} .
\label{mi}
\end{equation}

While mutual information measures how much information two variables share, it does not specify how information from multiple sources is organized. Partial Information Decomposition (PID) \citep{PID} addresses this issue by decomposing the information that two sources $X_1$ and $X_2$ provide about a target $Y$ into redundant, unique, and joint components. In this work, we use this perspective to distinguish information that is already available from individual components from information that only becomes available through their interaction.

We further use Integrated Information Decomposition ($\Phi$ID) \citep{IID}, which extends PID to temporal dynamics. Given two components at consecutive time steps, $\Phi$ID decomposes the time-delayed mutual information \citep{TDMI}
\begin{equation}
I(X_t^1,X_t^2;X_{t+1}^1,X_{t+1}^2)
\label{tdmi}
\end{equation}
into atoms that describe how information changes from source states at time $t$ to target states at time $t+1$. This framework allows us to separate information that is preserved across time from information that is constructed through interactions between components.

Operationally, we use two temporal atoms throughout the paper. The first measures information that is jointly expressed by a pair of components and remains jointly expressed at the next step. We use this quantity as an interaction-based composition score. The second measures information that is shared by individual components and remains shared at the next step. We use this quantity as a preservation score. Their difference provides a layer- and head-level indicator of whether a component is more composition-oriented or preservation-oriented. Full details of the lattice construction and Möbius inversion are provided in Appendix~\ref{app:info_decomposition}.

\subsection{Transformer Residual Computation}

The models analyzed in this study are decoder-only transformers with pre-layer normalization. For layer $l$, let $h_l\in\mathbb{R}^{T\times d_{\mathrm{model}}}$ denote the residual stream entering the layer. The attention and MLP updates are
\begin{equation}
\begin{gathered}
a_l = \mathrm{SelfAttention}_l(\mathrm{Norm}(h_l)), \\
\hat{h}_l = h_l + a_l, \\
m_l = \mathrm{MLP}_l(\mathrm{Norm}(\hat{h}_l)), \\
h_{l+1} = \hat{h}_l + m_l .
\end{gathered}
\label{eq:transformer_layer}
\end{equation}
Thus, the total contribution written by layer $l$ into the residual stream is
\begin{equation}
\Delta_l = h_{l+1}-h_l = a_l+m_l .
\label{eq:layer_delta}
\end{equation}
This additive structure lets us analyze how each layer modifies the residual stream and how later layers depend on information written by earlier layers. In later experiments, we use $\Delta_l$ both for representation probing and for causal interventions.

\section{Methods and Experiments}

\subsection{Layer-wise Composition and Preservation in LLMs}
\label{sec:composition_preservation}

We first identify which attention heads are more associated with information composition and which are more associated with information preservation during autoregressive reasoning. We treat each attention head as an elementary computational component and summarize its output at each generation step by the $L_2$ norm of the head output:
\begin{equation}
a(h_i,t)
=
\left\|
\mathrm{softmax}
\left(
\frac{q_t^i(K_t^i)^\top}{\sqrt{d_k}}
\right)
V_t^i
\right\|_2 .
\label{eq:head_activation}
\end{equation}
Here, $q_t^i$ is the query vector of head $h_i$ at time step $t$, and $K_t^i$ and $V_t^i$ are the corresponding key and value matrices.

For each pair of attention heads $h_i$ and $h_j$, we construct two temporal activation sequences over a generation trajectory:
\begin{equation}
\begin{gathered}
A^1 = [a(h_i,1),a(h_i,2),\dots,a(h_i,T)],\\
A^2 = [a(h_j,1),a(h_j,2),\dots,a(h_j,T)] .
\end{gathered}
\label{eq:head_time_series}
\end{equation}
These two sequences are treated as a two-component temporal system. We then compute the time-delayed mutual information
\begin{equation}
\mathrm{TDMI}
=
I(A_t^1,A_t^2;A_{t+1}^1,A_{t+1}^2).
\label{eq:tdmi_heads}
\end{equation}

Using the decomposition described in Appendix~\ref{app:info_decomposition}, we extract two operational scores. The first is an interaction-based composition score, defined by the temporal atom in which jointly expressed information remains jointly expressed at the next step:
\begin{equation}
S_{\mathrm{comp}}(h_i,h_j)
=
I_{\partial}(\alpha_{\mathrm{syn}}\to\beta_{\mathrm{syn}})_{(h_i,h_j)} .
\label{eq:pair_comp}
\end{equation}
The second is a preservation score, defined by the atom in which shared information remains shared at the next step:
\begin{equation}
S_{\mathrm{pres}}(h_i,h_j)
=
I_{\partial}(\alpha_{\mathrm{red}}\to\beta_{\mathrm{red}})_{(h_i,h_j)} .
\label{eq:pair_pres}
\end{equation}

Before computing these scores, we standardize each temporal activation sequence as
\begin{equation}
\widetilde{A}
=
\frac{A-\mathbb{E}[A]}{\sqrt{\mathrm{Var}(A)}} ,
\label{eq:standardization}
\end{equation}
so that the resulting quantities reflect temporal dependency patterns rather than absolute activation scale.

We aggregate pairwise scores for each head by averaging its interactions with all other heads:
\begin{equation}
S_{\mathrm{comp}}(h_i)
=
\frac{1}{N-1}
\sum_{j\neq i}
S_{\mathrm{comp}}(h_i,h_j),
\label{eq:head_comp}
\end{equation}
and compute $S_{\mathrm{pres}}(h_i)$ analogously. The head-level orientation score is then
\begin{equation}
S_{\mathrm{orient}}(h_i)
=
S_{\mathrm{comp}}(h_i)
-
S_{\mathrm{pres}}(h_i).
\label{eq:head_orientation}
\end{equation}
Positive values indicate that a head is more composition-oriented under this criterion, whereas negative values indicate a stronger preservation-oriented tendency.

Figure~\ref{fig:synergy_redundancy} shows the resulting layer-wise distribution in Qwen3-8B-Base on GSM8K. Composition-oriented heads are concentrated in middle layers, whereas preservation-oriented heads are more common in outer layers. This pattern suggests a division of labor in which outer layers primarily preserve and route information, while middle layers more strongly transform representations. This is consistent with prior interpretability findings that early layers are associated with detokenization and late layers with tokenization \citep{remarkable}.

We also aggregate the head-level scores into layer-level quantities under different sublayer interaction settings. As shown in Figure~\ref{fig:layer_info_richness}, the MLP-only setting exhibits a gradual transition across depth, whereas the combined MLP-and-attention setting produces a more pronounced composition-oriented region in the intermediate-to-late middle layers. Thus, the exact layer-wise profile depends on which sublayer interactions are included. Nevertheless, both views suggest that the strongest representation transformation is not uniformly distributed across the network, but is concentrated in a restricted depth range rather than in the outermost layers.

\begin{figure}[t]
    \centering

    \begin{subfigure}[t]{0.48\columnwidth}
        \centering
        \includegraphics[width=\linewidth]{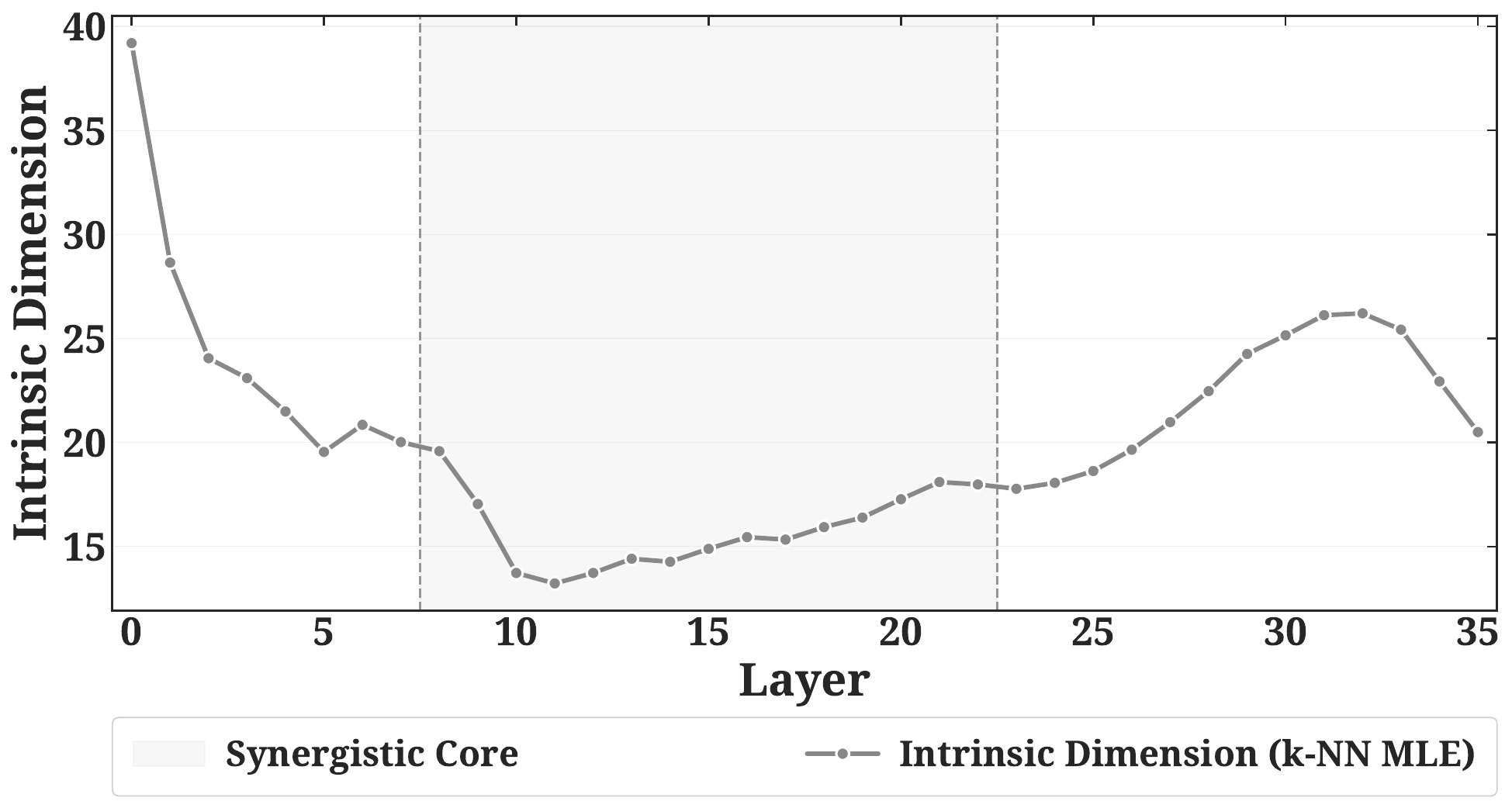}
        \caption{Intrinsic dimension across layers}
        \label{fig:lowdim_a}
    \end{subfigure}
    \hfill
    \begin{subfigure}[t]{0.48\columnwidth}
        \centering
        \includegraphics[width=\linewidth]{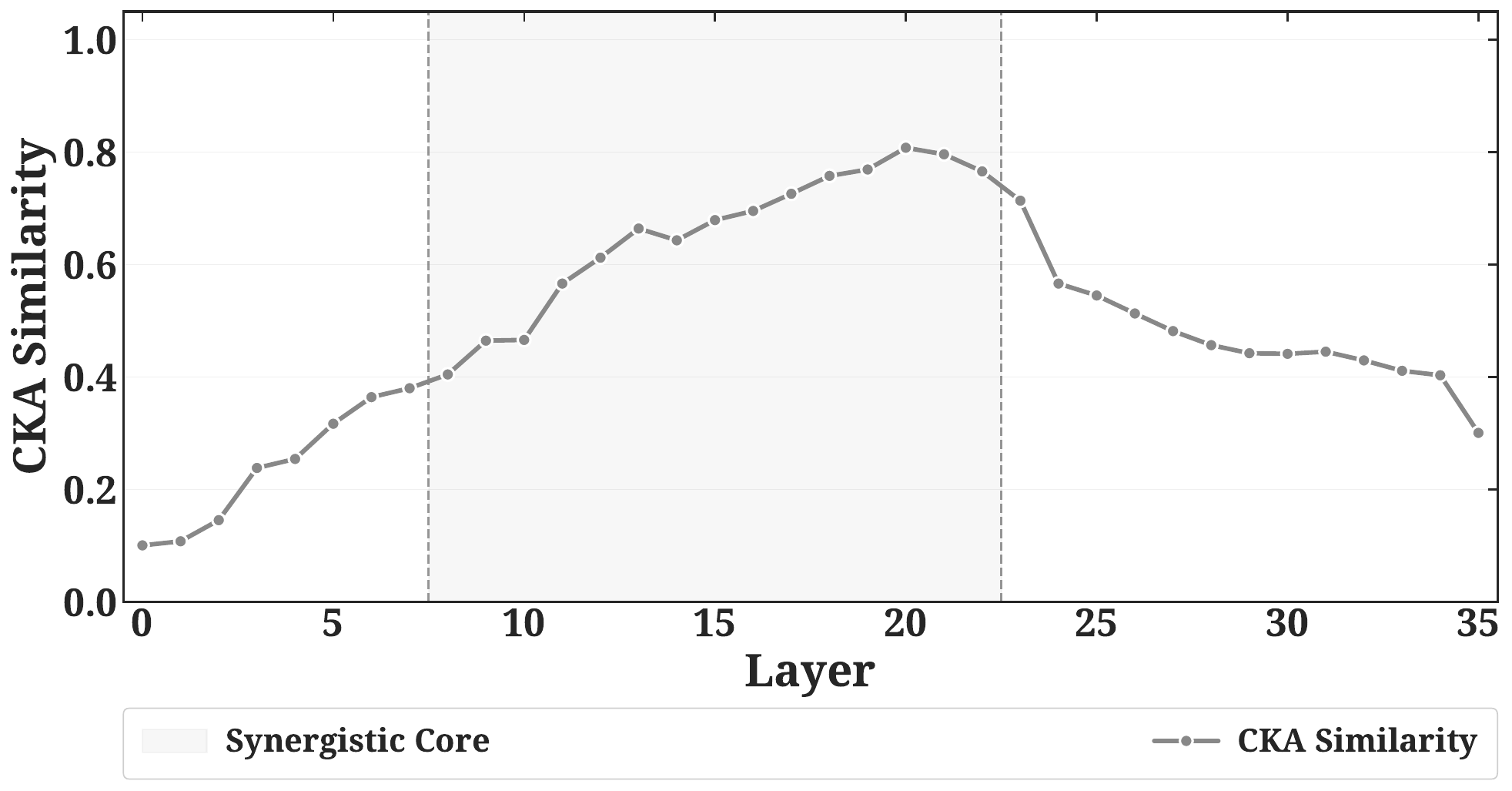}
        \caption{Cross-vocabulary CKA similarity across layers}
        \label{fig:lowdim_b}
    \end{subfigure}

    \vspace{0.5em}

    \begin{subfigure}[t]{0.48\columnwidth}
        \centering
        \includegraphics[width=\linewidth]{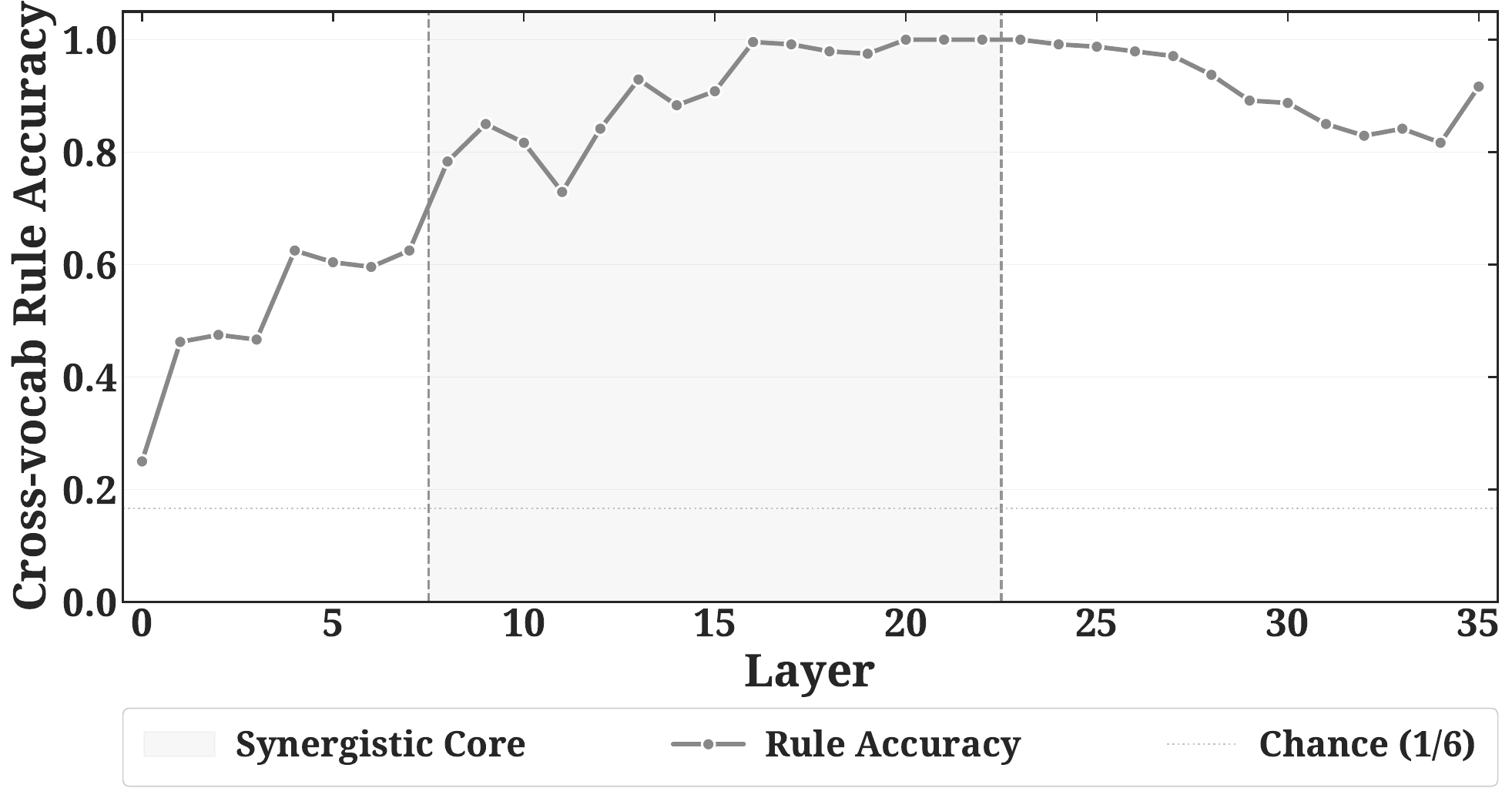}
        \caption{Cross-vocabulary abstract-rule probe accuracy}
        \label{fig:lowdim_c}
    \end{subfigure}
    \hfill
    \begin{subfigure}[t]{0.48\columnwidth}
        \centering
        \includegraphics[width=\linewidth]{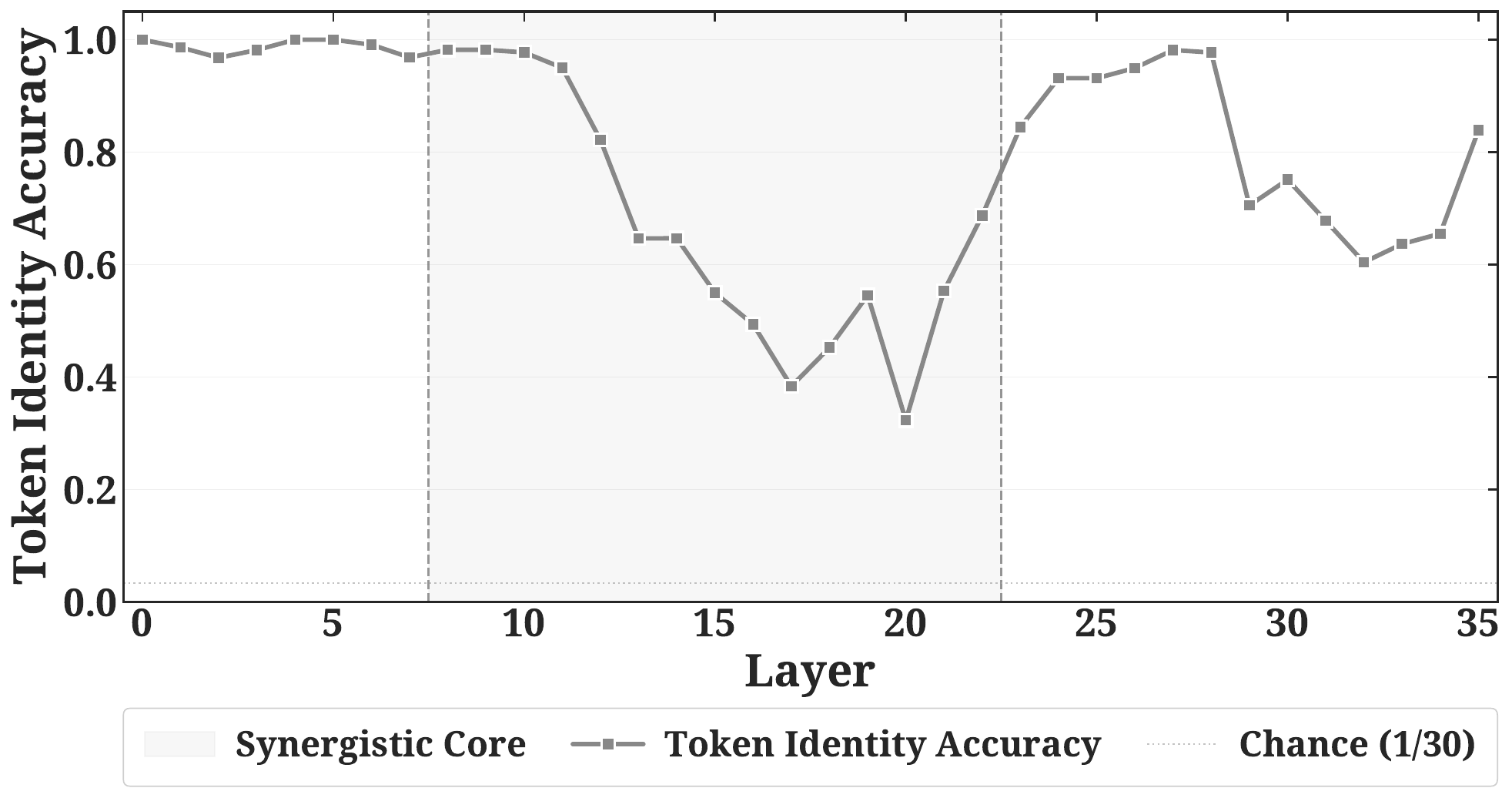}
        \caption{Visible token-identity probe accuracy}
        \label{fig:lowdim_d}
    \end{subfigure}

    \caption{
    Low-dimensional rule-level representations in middle layers.
    (a) Intrinsic dimension of hidden representations across layers. The estimated dimension decreases in the middle computation stage, suggesting a more compact representation geometry.
    (b) Cross-vocabulary CKA similarity between representations generated from disjoint vocabularies that instantiate the same symbolic rules. Higher similarity indicates stronger vocabulary-invariant rule alignment.
    (c) Cross-vocabulary rule-probe accuracy using residual updates $\Delta_l=h_{l+1}-h_l$. Middle layers achieve higher transfer accuracy, indicating that rule information is more recoverable from their updates.
    (d) Visible token-identity probe accuracy. Surface token identity becomes less recoverable in the same depth range, suggesting a shift from lexical information toward relational structure.
    }
    
    \label{fig:low_dimensional_subspace}
\end{figure}


\begin{figure}[t]
    \centering

    \begin{subfigure}[t]{0.48\columnwidth}
        \centering
        \includegraphics[width=\linewidth]{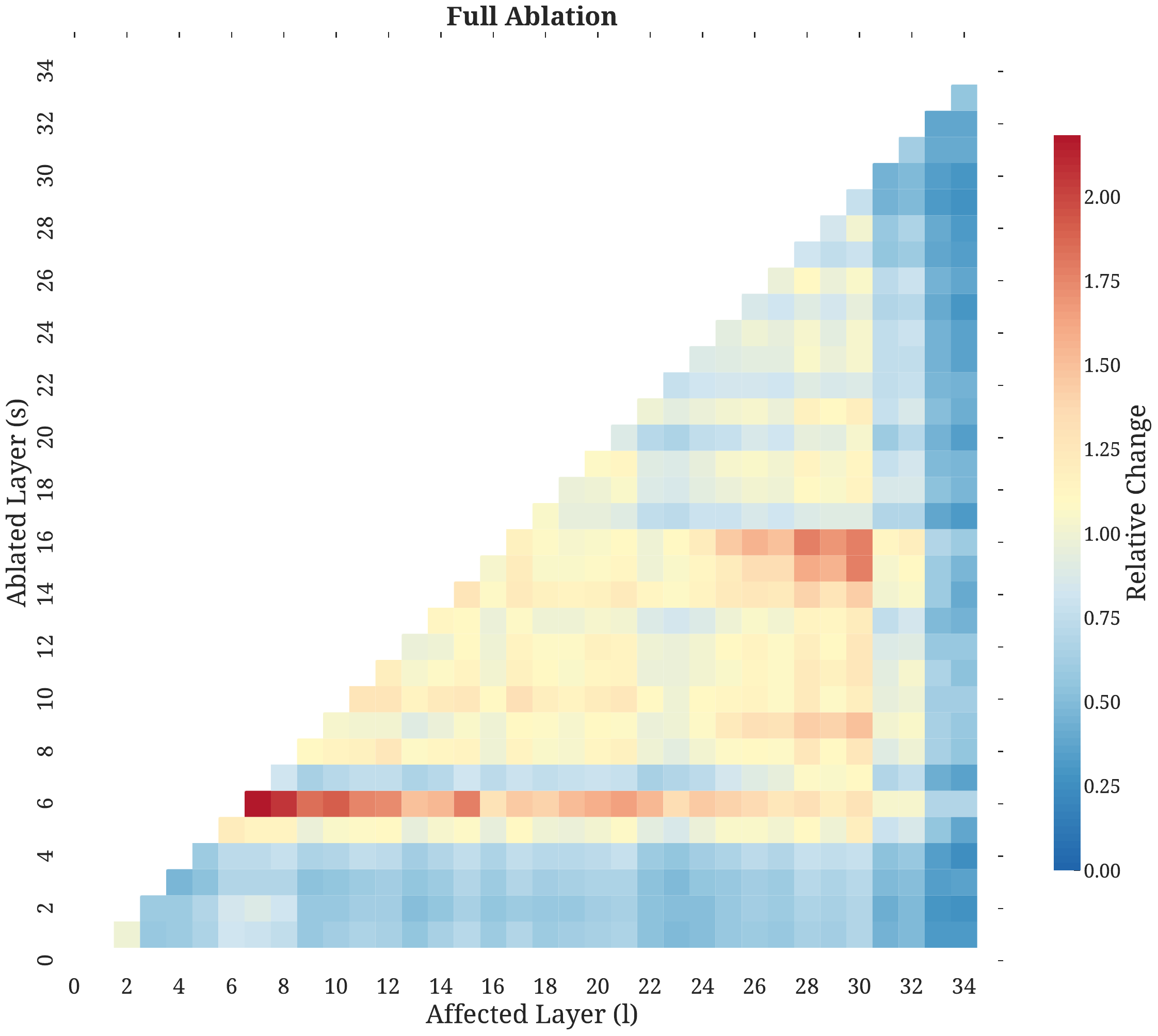}
        \caption{Full layer ablation}
        \label{fig:cross_layer_change_a}
    \end{subfigure}
    \hfill
    \begin{subfigure}[t]{0.48\columnwidth}
        \centering
        \includegraphics[width=\linewidth]{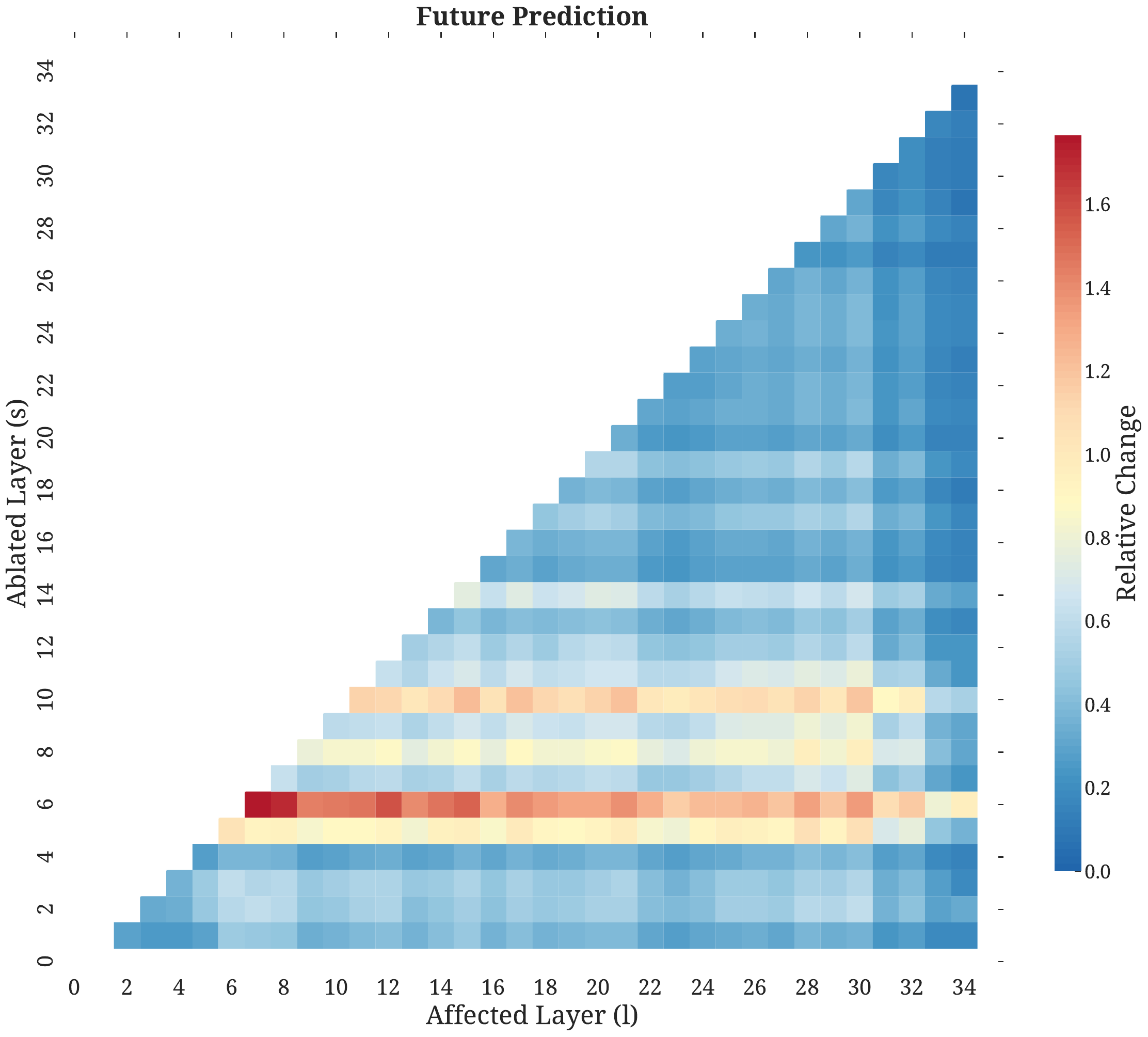}
        \caption{Prefill-only ablation}
        \label{fig:cross_layer_change_b}
    \end{subfigure}

    \vspace{0.5em}

    \begin{subfigure}[t]{0.48\columnwidth}
        \centering
        \includegraphics[width=\linewidth]{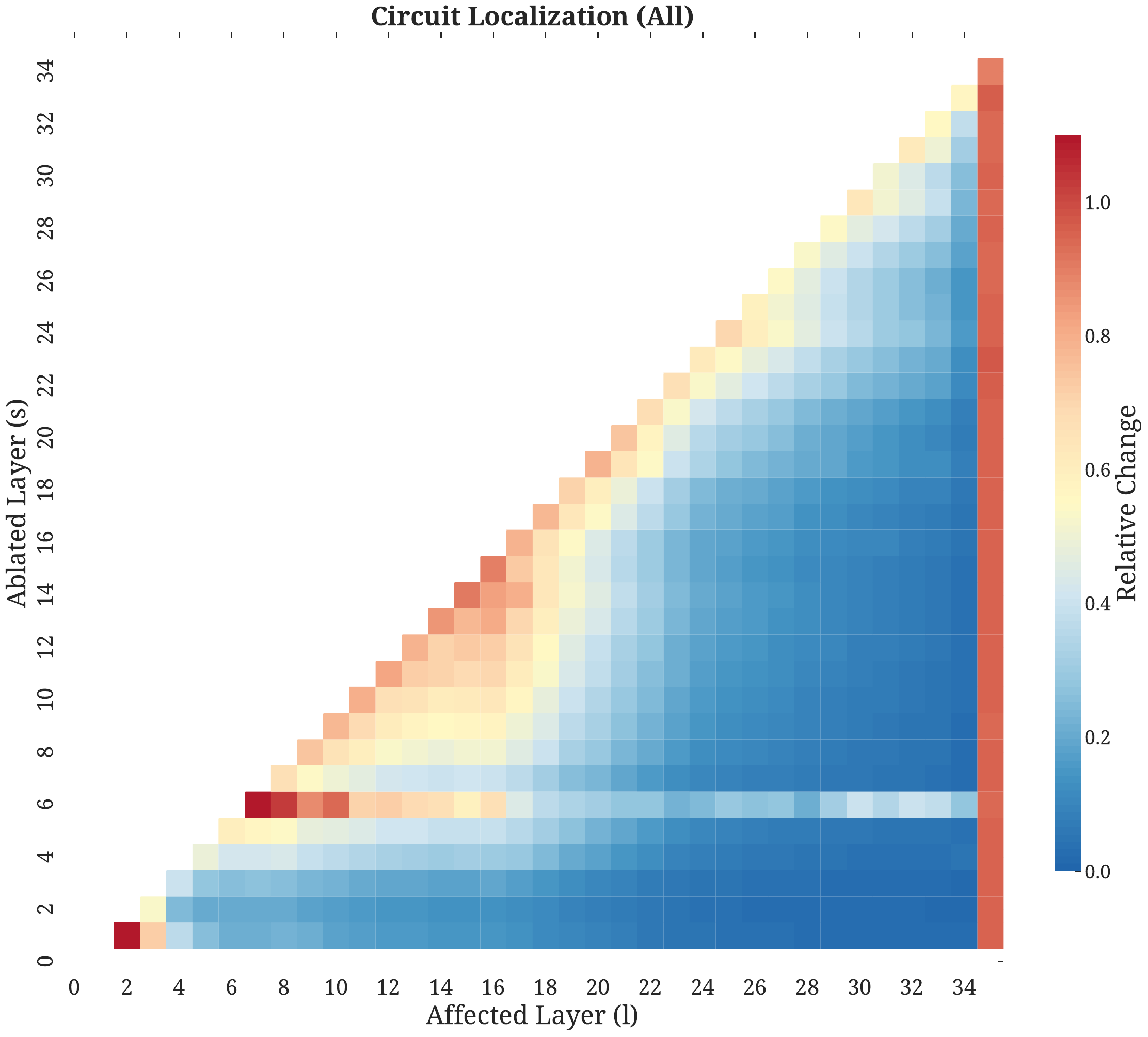}
        \caption{Circuit localization over all positions}
        \label{fig:cross_layer_change_c}
    \end{subfigure}
    \hfill
    \begin{subfigure}[t]{0.48\columnwidth}
        \centering
        \includegraphics[width=\linewidth]{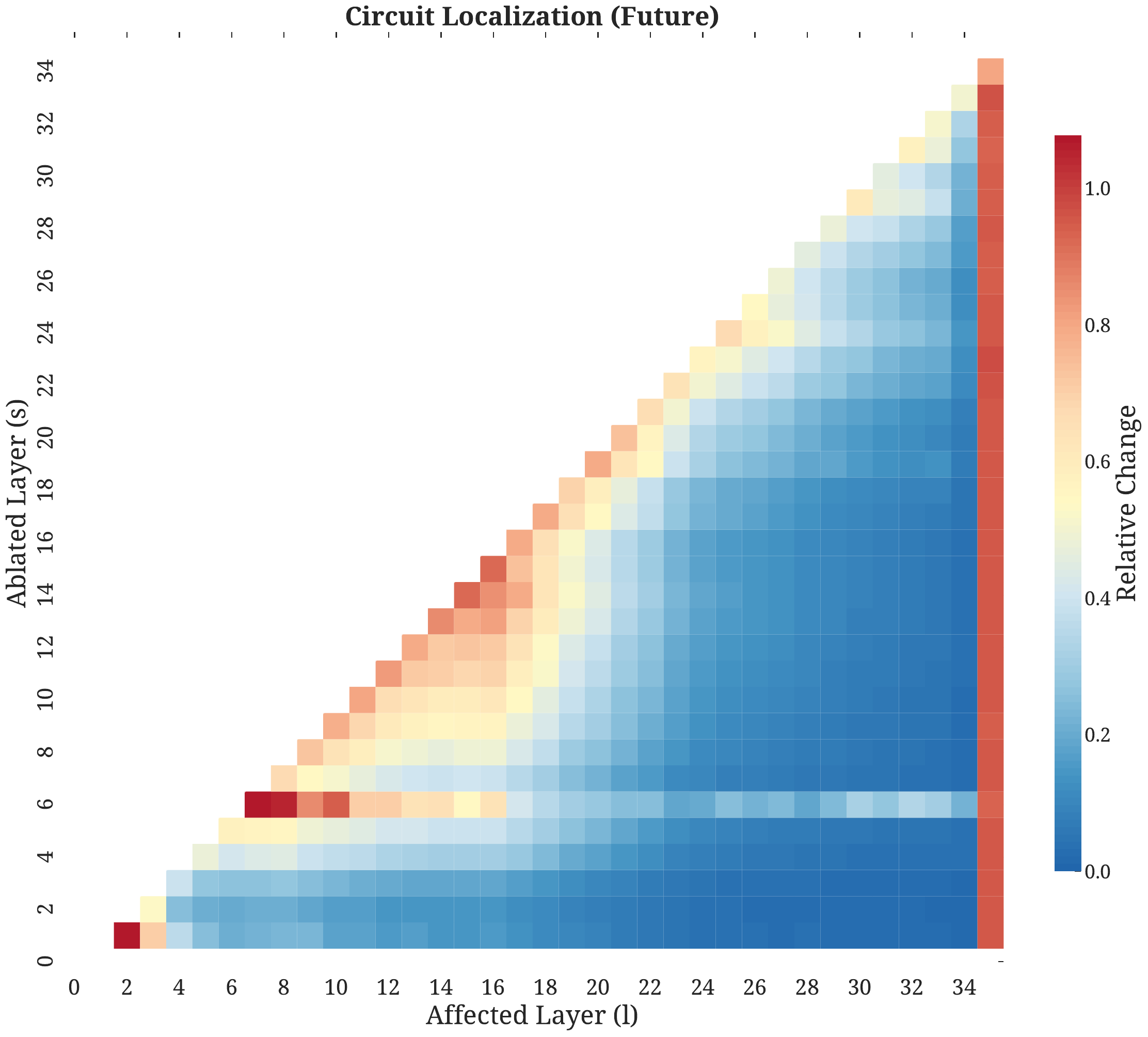}
        \caption{Circuit localization over future positions}
        \label{fig:cross_layer_change_d}
    \end{subfigure}

    \caption{
    Layer-level interventions for measuring downstream dependence.
    (a) Full layer ablation skips source layer $s$ during both prompt processing and decoding.
    (b) Prefill-only ablation skips source layer $s$ only during prompt processing and measures its delayed influence during later decoding.
    (c) Circuit localization removes the contribution of source layer $s$ from the input of a later layer $l$ over all positions.
    (d) Future-position circuit localization applies the same removal only to generated-token positions.
    Across these settings, stronger disturbances indicate that later computation depends more strongly on the information written by the intervened source layer.
    }
    
    \label{fig:cross_layer_change}
\end{figure}

\subsection{Low-Dimensional Rule-Level Representations}
\label{sec:low_dimensional_subspace}

The previous section localizes a restricted depth range where representation transformation is strongest. We next ask whether this region actually carries abstraction-related information, rather than merely showing larger changes in activation space. If middle layers support abstract reasoning, their representations should become less tied to surface tokens and more aligned with relational rules that generalize across vocabularies.

We test this hypothesis using two complementary analyses: representation geometry and cross-vocabulary probing. For geometry, we collect hidden states from autoregressive reasoning trajectories. For each layer $l$, we form a sample set
$\mathcal{H}^l=\{h_i^l\}_{i=1}^{N}$,
where each $h_i^l$ denotes the last-position hidden state produced at one decoding step from a GSM8K generation trajectory.

We estimate the local intrinsic dimension of $\mathcal{H}^l$ using the $k$-NN maximum likelihood estimator:
\begin{equation}
\widehat{d}_l =
\left[
\frac{1}{N(k-1)}
\sum_{i=1}^{N}
\sum_{j=1}^{k-1}
\log \frac{r_k(h_i^l)}{r_j(h_i^l)}
\right]^{-1},
\end{equation}
where $r_j(h_i^l)$ is the Euclidean distance from $h_i^l$ to its $j$-th nearest neighbor among all other samples in $\mathcal{H}^l$. A lower value of $\widehat{d}_l$ indicates that the representations occupy a lower-dimensional local manifold.

To test whether this lower-dimensional structure reflects lexical identity or relational structure, we construct paired symbolic sequences
$(x_i^A,x_i^B,y_i)$, where $x_i^A$ and $x_i^B$ instantiate the same rule
$y_i\in\{AABA,AABB,ABAA,ABAB,ABBA,ABBB\}$
using two disjoint single-token vocabularies. For example, the same rule $ABBA$ may appear as ``cat dog dog cat'' in $\mathcal{V}_A$ and ``oil pig pig oil'' in $\mathcal{V}_B$. Since $\mathcal{V}_A\cap\mathcal{V}_B=\emptyset$, alignment across these two vocabularies cannot be explained by shared token identity.

For each layer, we collect final-position representations
$X_l=\{h_l(x_i^A)\}$ and $Y_l=\{h_l(x_i^B)\}$,
and compute cross-vocabulary CKA:
\begin{equation}
\small
\begin{aligned}
\mathrm{CKA}(X_l,Y_l)
&=
\frac{\mathrm{HSIC}(K_l,L_l)}
{\sqrt{\mathrm{HSIC}(K_l,K_l)\mathrm{HSIC}(L_l,L_l)}} ,\\
K_l&=K(X_l),\quad L_l=K(Y_l).
\end{aligned}
\end{equation}
Higher CKA indicates stronger alignment between representations induced by different vocabularies but the same underlying rule.

We further probe the information written by each layer by training classifiers on the residual update
$\Delta_l=h_{l+1}-h_l$.
A cross-vocabulary rule probe is trained as
\[
p(y_i\mid \Delta_l)=\mathrm{softmax}(W_l\Delta_l+b_l),
\]
with train/test directions $\mathcal{V}_A\!\rightarrow\!\mathcal{V}_B$ and $\mathcal{V}_B\!\rightarrow\!\mathcal{V}_A$. We also train an identical probe in which the label is changed from the abstract rule $y_i$ to the visible query token $t_i$. Thus, a rule-level representation should show a double dissociation: high transfer accuracy for rules but lower recoverability of surface token identity.

Figure~\ref{fig:low_dimensional_subspace} supports this interpretation. The intrinsic dimension decreases in the middle computation stage, while cross-vocabulary CKA and rule-probe accuracy increase in a similar depth range. At the same time, visible token-identity accuracy becomes less recoverable. These results suggest that middle layers do not merely change the representation magnitude; they transform token-level information into a more compact and vocabulary-invariant rule-level representation. Detailed experimental settings, including sequence construction, vocabulary splits, and probe training protocols, are provided in Appendix~\ref{app:abstraction_probe}.

\begin{figure}[t]
    \centering
    \includegraphics[width=\columnwidth]{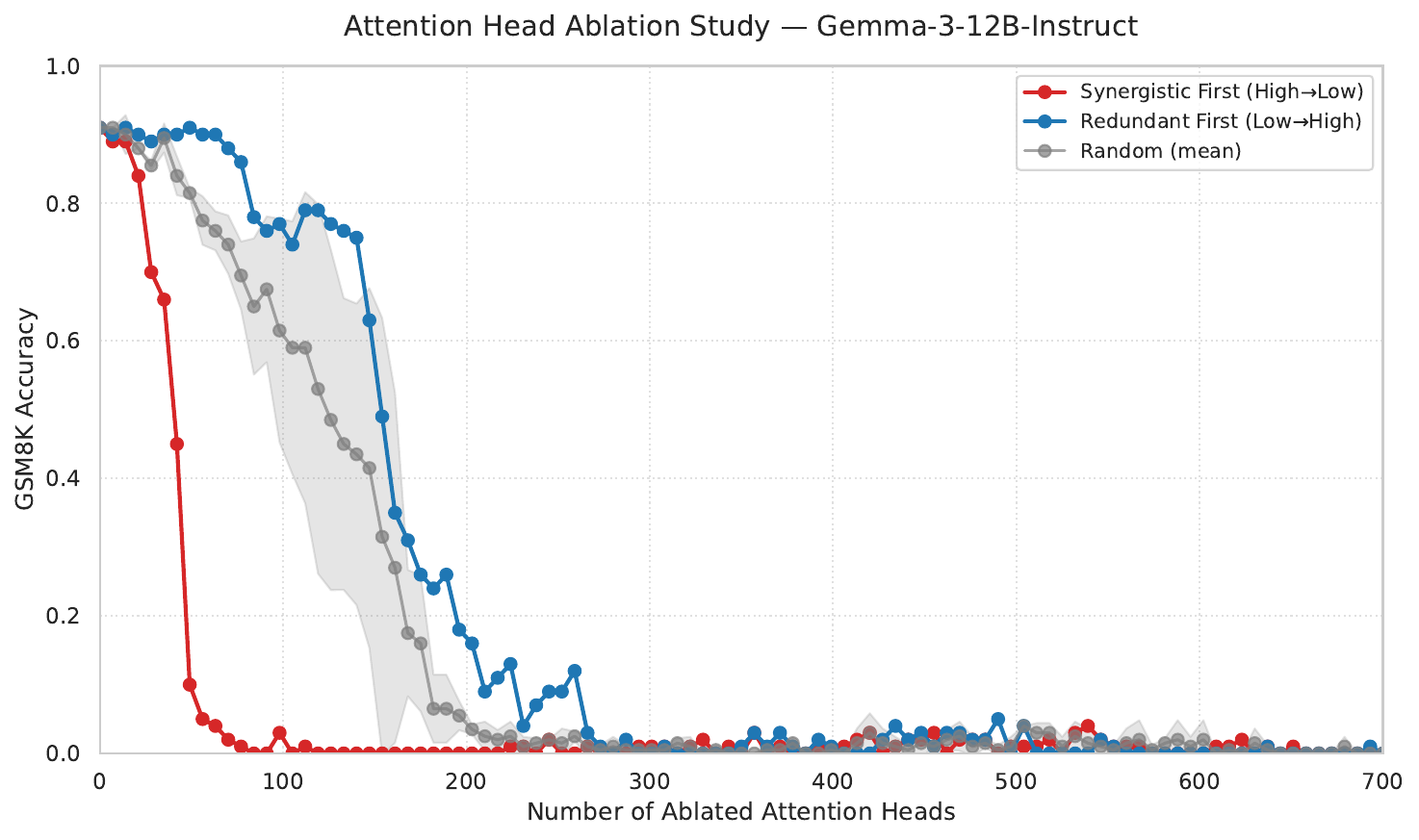}
    \caption{
    Cumulative attention-head ablation according to the composition--preservation orientation score in Gemma3-12B-IT.
    Removing composition-oriented heads produces the largest accuracy degradation, random ablation leads to a moderate drop, and preservation-oriented head ablation has the weakest effect.
    This ordered trend suggests that components identified by the interaction-based criterion have stronger functional influence on reasoning performance.
    }
    \label{fig:head_ablation}
\end{figure}

\subsection{Layer- and Head-Level Interventions}
\label{sec:layer_head_ablation}

The previous sections show that middle layers exhibit stronger representation transformation and more reusable rule-level information. We next test whether these components are functionally important for downstream computation. To this end, we perform interventions at both the layer and attention-head levels.

For layer-level analysis, we compare the residual update of a later layer $l$ in the original forward pass with the corresponding update after intervening on an earlier source layer $s$. The relative disturbance is defined as
\begin{equation}
D(s,l)=
\frac{
\left\|
(h_{l+1}-h_l)-(\bar{h}_{l+1}-\bar{h}_l)
\right\|_2
}{
\left\|h_{l+1}-h_l\right\|_2
},
\label{eq:relative_disturbance}
\end{equation}
where $h_l$ denotes the hidden state in the original forward pass and $\bar{h}_l$ denotes the hidden state under intervention. A larger value indicates that removing or modifying the source computation produces a stronger change in the later residual update.

We consider four complementary interventions, illustrated in Figure~\ref{fig:cross_layer_change}. First, in full layer ablation, we skip source layer $s$ during both prompt processing and decoding, and measure the resulting disturbance on all later layers $l>s$. This evaluates the overall downstream dependence on each source layer. Second, in prefill-only ablation, we skip layer $s$ only during prompt processing and then continue decoding normally. This tests whether a source layer can influence later generated-token computation through information written during the prompt stage.

Third, we perform circuit localization by removing the contribution written by a source layer from the input of a later layer. After obtaining the original forward pass, we compute the source-layer contribution as
\[
c_s = h_{s+1}-h_s .
\]
For each later layer $l>s$, we replace its input $h_l$ with $h_l-c_s$ and recompute only layer $l$. The resulting change measures how strongly layer $l$ depends on information supplied by layer $s$. Finally, we repeat the same intervention only on generated-token positions after the temporal split point, which isolates whether the source-layer contribution is reused during future token generation.

Figure~\ref{fig:cross_layer_change} shows that the largest downstream disturbances tend to arise from the middle computation stage across these intervention settings. Full layer ablation indicates that removing these layers has a larger effect on subsequent residual updates. Prefill-only ablation further suggests that their influence is not limited to the current token, but can propagate into later decoding steps. The circuit-localization results support the same interpretation: later layers are more sensitive to the removal of contributions written by middle layers than to the removal of contributions from outer layers. These results provide intervention-based evidence that the middle computation stage is functionally important for subsequent reasoning-related computation.

We further perform attention-head interventions using the head-level orientation score defined in Eq.~\ref{eq:head_orientation}. Heads are ranked by the difference between their composition and preservation scores. We then progressively ablate heads following three trajectories: from the most composition-oriented to the most preservation-oriented, from the most preservation-oriented to the most composition-oriented, and random ablation. This comparison tests whether components identified by the interaction-based criterion are more important than components selected at random or from the opposite end of the ranking.

As shown in Figure~\ref{fig:head_ablation}, ablating composition-oriented heads produces the largest accuracy degradation, random ablation leads to a moderate drop, and preservation-oriented head ablation has the weakest effect. This ordered pattern suggests that reasoning performance is not uniformly distributed across heads. Instead, it is more sensitive to the removal of heads identified as composition-oriented by our criterion.

Together, the layer- and head-level interventions support the same conclusion as the representation analyses: middle-stage components are not only correlated with rule-level representations, but also have a stronger functional influence on downstream computation. Outer layers appear to play a larger role in preserving and routing information, whereas middle layers contribute more strongly to transforming token-level features into representations used by later prediction steps.

\begin{table}[t]
\centering
\small
\setlength{\tabcolsep}{4pt}
\renewcommand{\arraystretch}{1.08}
\begin{tabular}{lccc}
\toprule
Model & 
\shortstack{Layer\\pattern} & 
\shortstack{Rule\\geometry} & 
\shortstack{Ablation\\trend} \\
\midrule
Qwen3-4B  & \cmark & \cmark & \cmark \\
Qwen3-8B  & \cmark & \cmark & \cmark \\
Qwen3-14B & \cmark & \cmark & \cmark \\
Llama-3.1-8B & \cmark & \cmark & \cmark \\
Gemma3-12B-IT & \cmark & \cmark & \cmark \\
\bottomrule
\end{tabular}
\caption{Cross-model summary. A check mark indicates that the same qualitative pattern is observed under the corresponding analysis. Full per-model results are reported in Appendix~\ref{app:cross_model}.}
\label{tab:cross_model_summary}
\end{table}

\section{Discussion and Limitations}

Our results suggest a layer-wise division of labor in LLM reasoning. Outer layers are more strongly associated with preserving and routing input-related features, whereas middle layers show stronger evidence of representation transformation. This pattern is supported by three complementary observations: interaction-based localization identifies a restricted middle-depth region, representation analyses show lower-dimensional and more vocabulary-invariant rule information in a similar region, and interventions indicate that later computation is more sensitive to removing middle-stage contributions.

These findings provide a mechanistic perspective on abstract reasoning in transformer language models. Rather than treating abstraction as a property of the model as a whole, our analysis suggests that abstraction-related computation can be partially localized to a middle computation stage. In this stage, token-level features appear to be reorganized into more reusable relational representations, which are then used by later layers for prediction. This interpretation is also consistent with the head-level ablation results, where components identified as more composition-oriented have a stronger effect on task performance when removed.

Several limitations remain. First, our experiments mainly focus on mathematical and symbolic reasoning tasks. Whether the same organization holds for open-ended language generation, code generation, long-context reasoning, or multimodal reasoning requires further validation. Second, our interaction analysis is based on pairwise attention-head dynamics. It may therefore miss higher-order interactions among larger groups of heads or between attention and MLP components. Third, our intervention results support functional importance, but they do not fully specify the algorithm implemented by the identified components. Future work could combine this analysis with circuit-level methods to recover more explicit computational mechanisms.

Finally, although some terminology in information-integration analysis is historically connected to neuroscience, our claims are purely computational. We do not argue for biological equivalence between LLMs and brains. Instead, we use interaction, preservation, and composition as operational tools for studying how information is transformed across transformer layers.

\section{Conclusion}

We studied where abstract reasoning is organized inside large language models by analyzing how transformer components transform information during autoregressive computation. Across localization, representation, and intervention analyses, we find that abstract reasoning is not uniformly distributed across layers. Instead, middle layers show stronger evidence of converting token-level information into compact and vocabulary-invariant rule-level representations, while outer layers more strongly preserve and route information. These results provide a layer-wise account of abstract reasoning in LLMs and suggest that middle-stage representation transformation is a key target for future mechanistic studies of reasoning.





\newpage

\bibliography{custom}

\appendix

\onecolumn

\newpage

\section{Experimental Setup}
\label{app:experimental_setup}

\subsection{Model Coverage}

We evaluate the proposed analysis on multiple open-source decoder-only language models. The proxy-collection pipeline supports the following models:
Qwen3-4B-Base, Qwen3-8B-Base, Qwen3-14B-Base, Llama-3.1-8B, Gemma3-12B-Base, Gemma3-12B-IT.
In the main paper, Qwen3-8B-Base is used as the primary case study, while additional models are used for cross-model validation.

\begin{table}[h]
\centering
\small
\begin{tabular}{lcc}
\toprule
Model & Role in paper & Dataset \\
\midrule
Qwen3-4B-Base & Cross-model validation & GSM8K \\
Qwen3-8B-Base & Main analysis & GSM8K \\
Qwen3-14B-Base & Cross-model validation & GSM8K \\
Llama-3.1-8B & Cross-family validation & GSM8K \\
Gemma3-12B-IT & Head-level ablation / validation & GSM8K \\
\bottomrule
\end{tabular}
\caption{Models used for the main and supplementary analyses.}
\label{tab:app_models}
\end{table}

\subsection{GSM8K Sampling and Prompting}

For the GSM8K-based proxy collection, we randomly sample 10 questions from the GSM8K test split using random seed 42. Each question is formatted with a zero-shot prompt:
\begin{quote}
\small
\texttt{Question: \{question\}\\
Answer:}
\end{quote}
The model then generates an answer autoregressively. We set the maximum number of newly generated tokens to 2048 and stop generation when an EOS token is produced. When available, the \texttt{<end\_of\_turn>} token is also added to the stop-token set. Generation uses greedy decoding with \texttt{do\_sample=False} and \texttt{use\_cache=True}.

\subsection{Activation Proxy Collection}

During generation, we collect scalar time-series proxies for intermediate computations rather than storing full activation vectors. This substantially reduces memory usage and makes pairwise information-decomposition computation tractable.

For each generated token step, we collect three types of proxies:

\begin{itemize}
    \item \textbf{Attention-head trajectory}: for each head $h_i$, we collect a scalar sequence $\{a(h_i,t)\}_{t=1}^{T}$ by taking the $L_2$ norm of its output at the active autoregressive position at each generation step.

    \item \textbf{MLP trajectory}: for each layer $l$, we collect a scalar sequence from the $L_2$ norm of the MLP update at the active autoregressive position across generation steps.

    \item \textbf{Combined layer trajectory}: for each layer $l$, we collect a scalar sequence from the $L_2$ norm of the combined attention and MLP update at the active autoregressive position across generation steps.
\end{itemize}

We also record the number of generated tokens for each sampled question. These effective lengths are later used to truncate each time series to the actual generated length before information-decomposition analysis.

\subsection{Forward Hooks}

Intermediate proxies are collected using forward hooks. For attention heads, hooks are registered on each layer's self-attention module. For MLP outputs, hooks are registered on each layer's MLP module. A layer-level hook is also registered to combine the attention output and MLP output for the \texttt{al\_plus\_ml} proxy. All hooks are removed after proxy collection.

\section{Information Decomposition Details}
\label{app:info_decomposition}

This appendix provides the full information-decomposition formulation used in the main text. We include these details for completeness, while the main paper only uses the resulting composition and preservation scores.

\subsection{Partial Information Decomposition}

Given two source variables $X_1$ and $X_2$ and a target variable $Y$, Partial Information Decomposition (PID) \citep{PID} decomposes the total mutual information $I(Y;\{X_1,X_2\})$ into redundant, unique, and synergistic components:
\begin{equation}
\begin{aligned}
I(Y;\{X_1,X_2\}) =
&\operatorname{Red}(Y;X_1,X_2) \\
&+ \operatorname{Syn}(Y;X_1,X_2) \\
&+ \operatorname{Unq}_1(Y;X_1,X_2) \\
&+ \operatorname{Unq}_2(Y;X_1,X_2).
\end{aligned}
\label{app:pid_total}
\end{equation}

The information that each individual source provides about the target can be written as
\begin{equation}
\begin{aligned}
I(Y;X_1) =
\operatorname{Red}(Y;X_1,X_2)
+
\operatorname{Unq}_1(Y;X_1,X_2),
\end{aligned}
\label{app:pid_x1}
\end{equation}
and
\begin{equation}
\begin{aligned}
I(Y;X_2) =
\operatorname{Red}(Y;X_1,X_2)
+
\operatorname{Unq}_2(Y;X_1,X_2).
\end{aligned}
\label{app:pid_x2}
\end{equation}

Following the minimum mutual information criterion, we define the redundant component as
\begin{equation}
\begin{aligned}
\operatorname{Red}(Y;X_1,X_2)
&= I_{\min}(Y;X_1,X_2) \\
&= \min \{ I(Y;X_1), I(Y;X_2) \}.
\end{aligned}
\label{app:red_min}
\end{equation}

Under this choice, the remaining PID atoms can be obtained from the equations above. In our experiments, this decomposition is used as the static basis for distinguishing information that is available from individual components from information that requires joint observation of multiple components.

\subsection{Integrated Information Decomposition}

Integrated Information Decomposition ($\Phi$ID) \citep{IID} extends PID to temporal information dynamics. For two components observed at consecutive time steps, it decomposes the time-delayed mutual information \citep{TDMI}
\begin{equation}
I(X_t^1,X_t^2;X_{t+1}^1,X_{t+1}^2)
\label{app:tdmi}
\end{equation}
into a set of temporal information atoms.

Let the source variables be
\[
X_t=\{X_t^1,X_t^2\},
\]
and the target variables be
\[
X_{t+1}=\{X_{t+1}^1,X_{t+1}^2\}.
\]
$\Phi$ID defines information states over antichains of source and target subsets. For example, information may be redundant across individual components, unique to one component, or synergistically available only from their joint state. The same categories are also defined for the target time step.

The double redundancy function $I_{\cap}(\alpha\to\beta)$ measures the information that is redundantly available from a source antichain $\alpha$ to a target antichain $\beta$. Using the minimum mutual information criterion, we compute it as
\begin{equation}
I_{\cap}(\alpha\to\beta)
=
\min_{A\in\alpha,\;B\in\beta}
I(X_t^A;X_{t+1}^B).
\label{app:double_red}
\end{equation}

This quantity is related to the partial temporal atoms $I_{\partial}$ by
\begin{equation}
I_{\cap}(\alpha\to\beta)
=
\sum_{\alpha'\preceq\alpha}
\sum_{\beta'\preceq\beta}
I_{\partial}(\alpha'\to\beta'),
\label{app:integral_eq}
\end{equation}
where $\preceq$ denotes the partial order over the antichain lattice.

The pure atoms are obtained by Möbius inversion over the source and target lattices:
\begin{equation}
\begin{aligned}
I_{\partial}(\alpha\to\beta)
=
\sum_{\alpha'\preceq\alpha}
\sum_{\beta'\preceq\beta}
&\mu(\alpha',\alpha)
\mu(\beta',\beta) \\
& I_{\cap}(\alpha'\to\beta'),
\end{aligned}
\label{app:mobius}
\end{equation}
where $\mu$ is the Möbius function induced by the lattice topology.

The temporal atoms conserve the total time-delayed mutual information:
\begin{equation}
\begin{aligned}
I(X_t^1,X_t^2;X_{t+1}^1,X_{t+1}^2)
=
\sum_{\alpha\in\mathcal{A}}
\sum_{\beta\in\mathcal{A}}
I_{\partial}(\alpha\to\beta),
\end{aligned}
\label{app:tdmi_sum}
\end{equation}
where $\mathcal{A}$ is the set of antichains.

\subsection{Operational Scores Used in This Work}

In the main experiments, each attention-head pair is treated as a two-component temporal system. We use two $\Phi$ID atoms as operational indicators.

First, the synergy-to-synergy atom measures information that is jointly available from a pair of components and remains jointly available at the next step:
\begin{equation}
S_{\mathrm{comp}}(h_i,h_j)
=
I_{\partial}(\alpha_{\mathrm{syn}}\to\beta_{\mathrm{syn}})_{(h_i,h_j)} .
\label{app:composition_score}
\end{equation}
We refer to this quantity in the main text as the interaction-based composition score.

Second, the redundancy-to-redundancy atom measures information that is shared across individual components and remains shared at the next step:
\begin{equation}
S_{\mathrm{pres}}(h_i,h_j)
=
I_{\partial}(\alpha_{\mathrm{red}}\to\beta_{\mathrm{red}})_{(h_i,h_j)} .
\label{app:preservation_score}
\end{equation}
We refer to this quantity as the preservation score.

For each head $h_i$, we aggregate its pairwise interactions with all other heads:
\begin{equation}
S_{\mathrm{comp}}(h_i)
=
\frac{1}{N-1}
\sum_{j\neq i}
S_{\mathrm{comp}}(h_i,h_j),
\label{app:head_comp}
\end{equation}
and
\begin{equation}
S_{\mathrm{pres}}(h_i)
=
\frac{1}{N-1}
\sum_{j\neq i}
S_{\mathrm{pres}}(h_i,h_j).
\label{app:head_pres}
\end{equation}

The head-level orientation score is then defined as
\begin{equation}
S_{\mathrm{orient}}(h_i)
=
S_{\mathrm{comp}}(h_i)
-
S_{\mathrm{pres}}(h_i).
\label{app:orientation_score}
\end{equation}

A positive value indicates that a head is more composition-oriented under this criterion, whereas a negative value indicates that it is more preservation-oriented. In the main text, this score is used to rank attention heads for layer-wise localization and causal ablation.

\subsection{Estimation and Aggregation}

We estimate the temporal information atoms using the public implementation of Integrated Information Decomposition. Unless otherwise specified, all experiments use a one-step temporal delay $\tau=1$, a Gaussian estimator, and the minimum-mutual-information redundancy function. This matches the formulation in Eq.~\ref{app:red_min} and provides a consistent estimator across all models and proxy types.

For each input trajectory, we treat either attention heads or layers as the components of the temporal system. For attention-head analysis, each component corresponds to one head $h_i$ in a specific layer. For layer-level analysis, each component corresponds to the scalar proxy of a transformer layer. Given a set of components $\mathcal{C}$, we compute the decomposition for all unordered component pairs:
\begin{equation}
\mathcal{P}
=
\{(c_i,c_j): c_i,c_j\in\mathcal{C},\; i<j\}.
\end{equation}
Each pair produces two temporal activation sequences. The sequences are truncated to the same effective generation length before the decomposition is computed. Pairs with fewer than five valid time steps are excluded from aggregation.

For each pair $(c_i,c_j)$, we extract the two atoms used in the main text:
\begin{equation}
S_{\mathrm{comp}}(c_i,c_j)
=
\mathbb{E}\left[
I_{\partial}(\alpha_{\mathrm{syn}}\to\beta_{\mathrm{syn}})
\right],
\end{equation}
and
\begin{equation}
S_{\mathrm{pres}}(c_i,c_j)
=
\mathbb{E}\left[
I_{\partial}(\alpha_{\mathrm{red}}\to\beta_{\mathrm{red}})
\right].
\end{equation}
The expectation is taken over the estimated atom values returned for the temporal trajectory. The first quantity is used as the interaction-based composition score, while the second is used as the preservation score.

We then aggregate pairwise scores into component-level scores by averaging over all interactions involving a component:
\begin{equation}
S_{\mathrm{comp}}(c_i)
=
\frac{1}{|\mathcal{C}|-1}
\sum_{j\neq i}
S_{\mathrm{comp}}(c_i,c_j),
\end{equation}
and analogously for $S_{\mathrm{pres}}(c_i)$. The orientation score is defined as
\begin{equation}
S_{\mathrm{orient}}(c_i)
=
S_{\mathrm{comp}}(c_i)
-
S_{\mathrm{pres}}(c_i).
\end{equation}
Positive values indicate stronger composition-oriented dynamics, whereas negative values indicate stronger preservation-oriented dynamics.

\subsection{Activation Proxies}

To make the decomposition tractable for large models, we do not apply $\Phi$ID directly to full hidden vectors. Instead, we summarize each component by a scalar temporal proxy collected along the autoregressive generation trajectory. At each generation step, the proxy is extracted from the active prediction position, i.e., the last position of the current forward pass used for next-token prediction. This should not be understood as using only the final token of the completed response; rather, repeating this extraction across decoding steps yields a temporal trajectory for each component.

We use three proxy families. First, for attention-head analysis, each head $h_i$ is represented by a scalar sequence $\{a(h_i,t)\}_{t=1}^{T}$, where $a(h_i,t)$ is the $L_2$ norm of the head output at the active prediction position at step $t$. Second, for MLP-level analysis, each layer is represented by the $L_2$ norm of its MLP update at the same active position across steps. Third, for combined layer-level analysis, each layer is represented by the $L_2$ norm of the sum of its attention and MLP updates across steps. These three proxy choices allow us to compare head-level attention dynamics, MLP-only layer dynamics, and combined residual-update dynamics under the same information-decomposition framework.

Before computing information-decomposition scores, each temporal sequence is standardized as in Eq.~\ref{eq:standardization}. This normalization removes absolute scale differences between components and makes the resulting scores depend primarily on temporal dependency structure.

\subsection{Trajectory Sampling}

For the GSM8K-based localization experiments, we sample a fixed set of questions from the test split and collect autoregressive trajectories using greedy decoding. Each prompt follows the zero-shot format:
\begin{quote}
\small
\texttt{Question: \{question\}\\
Answer:}
\end{quote}
Generation stops when an end-of-sequence token is produced or when the maximum generation length is reached. For each example, the resulting trajectory consists of the sequence of active prediction positions visited during generation. The same sampling and decoding protocol is used across model families so that differences in layer-wise profiles are not caused by different data-processing procedures.

The main purpose of this trajectory collection is not to estimate benchmark accuracy, but to obtain representative reasoning-time activation dynamics for information-decomposition analysis. Accuracy and performance degradation are evaluated separately in the intervention experiments.

\section{Abstraction Probe Details}
\label{app:abstraction_probe}

This appendix describes the representation-geometry and probing analyses used to test whether the middle computation stage carries rule-level information rather than merely reflecting surface-token identity. We consider three complementary measurements: intrinsic dimension, cross-vocabulary representational alignment, and linear probes for rule and token information.

\subsection{Intrinsic Dimension of Reasoning Trajectories}

For the intrinsic-dimension analysis, we collect hidden states from autoregressive reasoning trajectories on GSM8K. For each layer $l$, let
\[
\mathcal{H}_l=\{h_{l,i}\}_{i=1}^{N}
\]
denote the set of hidden states collected from the active prediction position across generated tokens and questions. In our implementation, we use 50 GSM8K examples, greedy decoding, and at most 30 generated tokens per question. Thus, $\mathcal{H}_l$ contains the layer-$l$ representations observed along the reasoning trajectories.

We estimate the local intrinsic dimension of $\mathcal{H}_l$ using the $k$-nearest-neighbor maximum-likelihood estimator with $k=20$. Let $r_j(h_{l,i})$ be the Euclidean distance from $h_{l,i}$ to its $j$-th nearest neighbor among the other samples in $\mathcal{H}_l$. The estimated intrinsic dimension is
\begin{equation}
\widehat{d}_l
=
\left[
\frac{1}{N(k-1)}
\sum_{i=1}^{N}
\sum_{j=1}^{k-1}
\log
\frac{r_k(h_{l,i})}{r_j(h_{l,i})}
\right]^{-1}.
\label{app:eq:intrinsic_dimension}
\end{equation}
A smaller value of $\widehat{d}_l$ indicates that the corresponding layer representations occupy a lower-dimensional local manifold. In the main text, we use this quantity to examine whether the middle computation stage corresponds to a more compact representation geometry.

\subsection{Synthetic Rule Sequences}

To separate rule-level abstraction from lexical identity, we construct symbolic sequences from two disjoint vocabularies $\mathcal{V}_A$ and $\mathcal{V}_B$, where
\[
\mathcal{V}_A\cap\mathcal{V}_B=\emptyset .
\]
Each sequence is generated from an abstract equality pattern over two latent symbols $A$ and $B$, but instantiated using concrete single-token words from one of the vocabularies. For example, the rule $ABBA$ can be instantiated as ``cat dog dog cat'' in $\mathcal{V}_A$ and as ``oil pig pig oil'' in $\mathcal{V}_B$.

For CKA analysis, we use a multi-rule template set
\[
\mathcal{R}_{\mathrm{CKA}}
=
\{ABA, ABB, AAB, ABAB, ABBA, ABBB\},
\]
with 50 sequences per rule in each vocabulary group. For probing, we use a length-controlled rule set
\[
\mathcal{R}_{\mathrm{probe}}
=
\{AABA, AABB, ABAA, ABAB, ABBA, ABBB\},
\]
with 20 sequences per rule in each vocabulary group. The length-controlled design removes trivial length cues and makes the probe rely more directly on relational structure.

Each probe sequence consists of two complete examples followed by an incomplete query prefix. Formally, for a rule $y\in\mathcal{R}_{\mathrm{probe}}$, a sequence has the form
\[
x = [e_1(y), e_2(y), q(y)],
\]
where $e_1(y)$ and $e_2(y)$ are complete instantiations of the rule and $q(y)$ is a three-token prefix of a new instantiation. The visible query-token label is defined as the last token in $q(y)$.

\subsection{Cross-Vocabulary CKA}

For each layer $l$, we collect the last-position hidden states induced by the two vocabulary groups:
\[
X_l=\{h_l(x_i^A)\}_{i=1}^{M},
\qquad
Y_l=\{h_l(x_i^B)\}_{i=1}^{M},
\]
where $x_i^A$ and $x_i^B$ instantiate the same abstract rule using disjoint vocabularies. Since the two vocabularies share no surface tokens, high alignment between $X_l$ and $Y_l$ indicates that the representations are organized by rule structure rather than lexical identity.

We compute centered kernel alignment as
\begin{equation}
\mathrm{CKA}(X_l,Y_l)
=
\frac{\mathrm{HSIC}(K_l,L_l)}
{
\sqrt{
\mathrm{HSIC}(K_l,K_l)
\mathrm{HSIC}(L_l,L_l)
}
},
\label{app:eq:cka}
\end{equation}
where $K_l$ and $L_l$ are kernel matrices computed from $X_l$ and $Y_l$. By default, we use an RBF kernel with the bandwidth selected by the median heuristic:
\begin{equation}
K_l(i,j)
=
\exp\left(
-\frac{\|x_{l,i}-x_{l,j}\|_2^2}{2\sigma_X^2}
\right),
\qquad
\sigma_X^2
=
\operatorname{median}_{i\neq j}
\|x_{l,i}-x_{l,j}\|_2^2 ,
\label{app:eq:rbf_kernel}
\end{equation}
and analogously for $L_l$. The kernel matrices are centered before computing HSIC. We also support a linear-kernel variant, but the main reported CKA curve uses the RBF kernel.

\subsection{Layer-Wise Rule Probes}

To isolate the information written by each layer, we probe residual updates rather than raw hidden states. For each layer $l$, we define
\begin{equation}
\Delta_l(x)=h_{l+1}(x)-h_l(x).
\label{app:eq:delta_probe}
\end{equation}
This removes much of the residual-stream information already present before layer $l$ and focuses the probe on what the layer contributes.

Let $y_i\in\mathcal{R}_{\mathrm{probe}}$ be the abstract rule label of example $x_i$. For each layer, we train a linear classifier
\begin{equation}
p(y_i\mid \Delta_l(x_i))
=
\operatorname{softmax}
(W_l\Delta_l(x_i)+b_l).
\label{app:eq:rule_probe}
\end{equation}
The probe is evaluated in both cross-vocabulary directions:
\[
\mathcal{V}_A\rightarrow\mathcal{V}_B,
\qquad
\mathcal{V}_B\rightarrow\mathcal{V}_A.
\]
The reported rule-probe accuracy is
\begin{equation}
\mathrm{Acc}^{\mathrm{rule}}_l
=
\frac{1}{2}
\left(
\mathrm{Acc}_{A\rightarrow B,l}^{\mathrm{rule}}
+
\mathrm{Acc}_{B\rightarrow A,l}^{\mathrm{rule}}
\right).
\label{app:eq:rule_probe_acc}
\end{equation}
In our implementation, the classifier is a multinomial logistic regression model with $C=1.0$, maximum 500 optimization iterations, and fixed random seed 42.

\subsection{Visible Token-Identity Probes}

As a control, we train an analogous probe to predict the visible query-token identity instead of the abstract rule. Let $t_i$ denote the last visible token in the incomplete query prefix. The token probe is defined as
\begin{equation}
p(t_i\mid \Delta_l(x_i))
=
\operatorname{softmax}
(U_l\Delta_l(x_i)+c_l).
\label{app:eq:token_probe}
\end{equation}
Unlike the rule probe, the token probe is evaluated within each vocabulary group, because token labels are vocabulary-specific. For each group, we train and evaluate the classifier using stratified cross-validation and then average the two group-level accuracies:
\begin{equation}
\mathrm{Acc}^{\mathrm{tok}}_l
=
\frac{1}{2}
\left(
\mathrm{Acc}_{A,l}^{\mathrm{tok}}
+
\mathrm{Acc}_{B,l}^{\mathrm{tok}}
\right).
\label{app:eq:token_probe_acc}
\end{equation}

Before training the token probe, we reduce the residual-update features with PCA to at most 200 dimensions:
\begin{equation}
z_l(x_i)
=
P_l^\top \Delta_l(x_i),
\qquad
P_l\in\mathbb{R}^{d\times d'},
\quad
d'=\min(200,N,d).
\label{app:eq:pca_token_probe}
\end{equation}
The token classifier is then trained on $z_l(x_i)$. We use three-fold stratified cross-validation within each vocabulary group and report the mean accuracy across folds and groups.

\subsection{Interpreting the Probe Results}

The rule and token probes are designed to test a double dissociation. A layer is considered more rule-oriented when $\mathrm{Acc}^{\mathrm{rule}}_l$ is high under cross-vocabulary transfer, while $\mathrm{Acc}^{\mathrm{tok}}_l$ is comparatively low. This pattern indicates that the layer update contains information that generalizes across surface vocabularies, rather than merely preserving the identity of visible tokens. Together with intrinsic dimension and CKA, these probes provide complementary evidence that middle layers form compact and vocabulary-invariant rule-level representations.

\section{Intervention Details}
\label{app:intervention_details}

This appendix describes the layer- and head-level interventions used to test whether the components identified by our interaction-based criterion have functional influence on downstream computation. We consider two complementary intervention families. Layer-level interventions measure how removing or subtracting the contribution of a source layer changes later residual updates. Head-level interventions measure how cumulative removal of attention heads affects GSM8K accuracy.

\subsection{Layer-Level Interventions}

Let $F_l$ denote the computation performed by transformer layer $l$, and let $H_l\in\mathbb{R}^{T\times d}$ be the residual-stream representation entering layer $l$ for a generated sequence of length $T$. The normal layer transition is
\begin{equation}
H_{l+1}=F_l(H_l).
\end{equation}
We define the contribution written by layer $l$ as
\begin{equation}
C_l = H_{l+1}-H_l .
\label{app:eq:layer_contribution}
\end{equation}
For scalar summaries over generated sequences, we use the mean residual contribution
\begin{equation}
\bar{C}_l
=
\frac{1}{T}
\sum_{t=1}^{T}
C_l(t),
\label{app:eq:mean_layer_contribution}
\end{equation}
where $C_l(t)$ denotes the contribution at position $t$.

Given a source layer $s$ and a later affected layer $l>s$, we measure downstream change by comparing the normal contribution of layer $l$ with its contribution under an intervention on layer $s$. Let $\bar{C}_l$ denote the normal contribution and $\widetilde{C}^{(s)}_l$ denote the corresponding contribution after intervention. The relative disturbance is
\begin{equation}
D(s,l)
=
\frac{
\left\|
\bar{C}_l-\widetilde{C}^{(s)}_l
\right\|_2
}{
\left\|
\bar{C}_l
\right\|_2+\epsilon
},
\label{app:eq:layer_disturbance}
\end{equation}
where $\epsilon$ is a small numerical constant. A larger value indicates that later computation is more sensitive to the source-layer intervention.

\paragraph{Full layer ablation.}
The first intervention removes source layer $s$ throughout the entire generation process. Operationally, we replace the output of layer $s$ with its input:
\begin{equation}
\widetilde{H}_{s+1}=H_s .
\label{app:eq:full_layer_ablation}
\end{equation}
All subsequent layers operate on the modified residual stream. We then compute $D(s,l)$ for all $l>s$. This intervention measures the overall downstream dependence on layer $s$.

\paragraph{Prefill-only ablation.}
The second intervention removes source layer $s$ only during prompt processing. Let $t_{\mathrm{prefill}}$ denote the prefill stage. We apply
\begin{equation}
\widetilde{H}_{s+1}^{\,t_{\mathrm{prefill}}}
=
H_s^{t_{\mathrm{prefill}}},
\end{equation}
but restore the normal computation of layer $s$ during subsequent decoding steps:
\begin{equation}
\widetilde{H}_{s+1}^{\,t}
=
F_s(H_s^t),
\qquad
t>t_{\mathrm{prefill}} .
\end{equation}
This intervention tests whether information written during prompt processing continues to influence later token generation.

\paragraph{Circuit localization over all positions.}
The third intervention asks whether information written by source layer $s$ is directly used by a later layer $l$. After a normal forward pass, we compute the source contribution
\begin{equation}
C_s = H_{s+1}-H_s .
\end{equation}
For each later layer $l>s$, we subtract this contribution from the input of layer $l$:
\begin{equation}
\widetilde{H}_l^{(s)}
=
H_l-C_s,
\label{app:eq:subtract_source_contribution}
\end{equation}
and recompute only layer $l$:
\begin{equation}
\widetilde{H}_{l+1}^{(s)}
=
F_l(\widetilde{H}_l^{(s)}).
\end{equation}
The modified contribution is therefore
\begin{equation}
\widetilde{C}_{l}^{(s)}
=
\widetilde{H}_{l+1}^{(s)}
-
\widetilde{H}_{l}^{(s)} .
\end{equation}
We measure the position-wise relative disturbance as
\begin{equation}
D_{\mathrm{all}}(s,l)
=
\frac{1}{|\mathcal{T}|}
\sum_{t\in\mathcal{T}}
\frac{
\left\|
C_l(t)-\widetilde{C}_{l}^{(s)}(t)
\right\|_2
}{
\left\|
C_l(t)
\right\|_2+\epsilon
},
\label{app:eq:circuit_all}
\end{equation}
where $\mathcal{T}=\{1,\dots,T\}$ is the set of all sequence positions. This measures how much layer $l$ depends on the contribution written by layer $s$.

\paragraph{Future-position circuit localization.}
The fourth intervention uses the same contribution-subtraction operation, but evaluates the disturbance only on later generated positions. Let $\mathcal{T}_{\mathrm{future}}\subseteq\mathcal{T}$ denote the subset of positions after a temporal split point. We compute
\begin{equation}
D_{\mathrm{future}}(s,l)
=
\frac{1}{|\mathcal{T}_{\mathrm{future}}|}
\sum_{t\in\mathcal{T}_{\mathrm{future}}}
\frac{
\left\|
C_l(t)-\widetilde{C}_{l}^{(s)}(t)
\right\|_2
}{
\left\|
C_l(t)
\right\|_2+\epsilon
}.
\label{app:eq:circuit_future}
\end{equation}
This setting isolates whether the contribution written by source layer $s$ is reused during later generation, rather than only affecting the current local computation.

\subsection{Head-Level Cumulative Ablation}

We next evaluate whether the attention heads ranked as composition-oriented by our information-decomposition criterion are more important for reasoning performance. For each head $h_i$, we use the orientation score
\begin{equation}
S_{\mathrm{orient}}(h_i)
=
S_{\mathrm{comp}}(h_i)
-
S_{\mathrm{pres}}(h_i),
\end{equation}
defined in Appendix~\ref{app:info_decomposition}. Heads with larger $S_{\mathrm{orient}}$ are more composition-oriented, while heads with smaller values are more preservation-oriented.

Let $\mathcal{H}$ be the set of attention heads considered for ablation. We exclude the first and last transformer layers from the cumulative head-ablation experiment, so that the comparison focuses on internal computation rather than input embedding or final tokenization effects. We compare three ablation orders:
\begin{equation}
\pi_{\mathrm{comp}}
=
\operatorname{argsort}_{h\in\mathcal{H}}
\left(
-S_{\mathrm{orient}}(h)
\right),
\end{equation}
\begin{equation}
\pi_{\mathrm{pres}}
=
\operatorname{argsort}_{h\in\mathcal{H}}
\left(
S_{\mathrm{orient}}(h)
\right),
\end{equation}
and a random permutation $\pi_{\mathrm{rand}}$ of $\mathcal{H}$.

For an ablation budget $k$, let
\begin{equation}
\mathcal{A}_k^{\pi}
=
\{\pi_1,\pi_2,\dots,\pi_k\}
\end{equation}
be the first $k$ heads selected by order $\pi$. We evaluate the model after ablating $\mathcal{A}_k^{\pi}$ for increasing values of $k$. This produces an accuracy curve
\begin{equation}
\mathrm{Acc}_{\pi}(k)
=
\frac{1}{|\mathcal{D}|}
\sum_{(x,y)\in\mathcal{D}}
\mathbf{1}
\left[
\operatorname{Ans}
\left(
M_{\setminus \mathcal{A}_k^{\pi}}(x)
\right)
=
y
\right],
\label{app:eq:ablation_accuracy}
\end{equation}
where $\mathcal{D}$ is the GSM8K evaluation subset, $M_{\setminus \mathcal{A}_k^{\pi}}$ denotes the model after ablating the selected heads, and $\operatorname{Ans}(\cdot)$ extracts the final numerical answer from the generated response.

\paragraph{Head removal.}
For a selected head $h=(l,j)$, we remove its contribution by zeroing the corresponding query projection slice and output projection slice. Let $d_h$ be the head dimension. The intervention is
\begin{equation}
W_Q^{(l)}
[jd_h:(j+1)d_h,:]
\leftarrow
0,
\end{equation}
and
\begin{equation}
W_O^{(l)}
[:,jd_h:(j+1)d_h]
\leftarrow
0 .
\end{equation}
This prevents the selected head from contributing to the attention output. Original weights are saved before the intervention and restored after each ablation trajectory.

\paragraph{Evaluation protocol.}
For head-level ablation, we evaluate GSM8K exact-match accuracy under three trajectories: composition-oriented heads first, preservation-oriented heads first, and random heads. The random trajectory is repeated with multiple seeds when available, and the reported random curve is the mean across random runs. Generation uses greedy decoding and stops when a final-answer pattern is produced or when the maximum generation length is reached. This design tests whether accuracy degrades faster when heads identified as composition-oriented are removed first.

\section{Cross-Model Results}
\label{app:cross_model}

This appendix reports cross-model results for the same analyses used in the main text. The main paper uses Qwen3-8B-Base as the primary case study for layer-wise localization, representation geometry, and layer-level intervention. The head-level ablation result in the main paper is shown on Gemma3-12B-IT. Here, we provide additional results on Qwen3-4B-Base, Qwen3-14B-Base, Llama-3.1-8B, and Gemma3-12B-IT to evaluate whether the observed layer-wise organization is model-specific.

Across these models, we observe the same qualitative pattern. First, composition-oriented signals are concentrated away from the outermost layers and tend to appear in a restricted middle-depth region. Second, representation-geometry analyses show stronger rule-level structure in similar regions, as reflected by lower-dimensional representations, higher cross-vocabulary alignment, and higher cross-vocabulary rule-probe accuracy. Third, intervention results indicate that removing or perturbing middle-stage components produces larger downstream changes than perturbing outer-layer components. These results support the conclusion that the layer-wise organization reported in the main text is not specific to a single model family.

\subsection{Cross-Model Layer-wise Localization}
\label{app:cross_model_localization}

The main text reports the layer-wise localization results on Qwen3-8B-Base. Here we repeat the same analysis on four additional models. For each model, we show the head-level composition--preservation orientation heatmap together with the cosine similarity of residual updates. The goal is not to require the strongest composition-oriented region to appear at the exact geometric center of every model, but to test whether such components are consistently concentrated away from the outermost layers.

\begin{figure}[H]
\centering
\captionsetup{font=small,skip=3pt}
\captionsetup[subfigure]{font=small,skip=2pt}

\begin{subfigure}[t]{0.92\textwidth}
    \centering
    \includegraphics[
        width=\linewidth,
        trim={0 10 0 8},
        clip
    ]{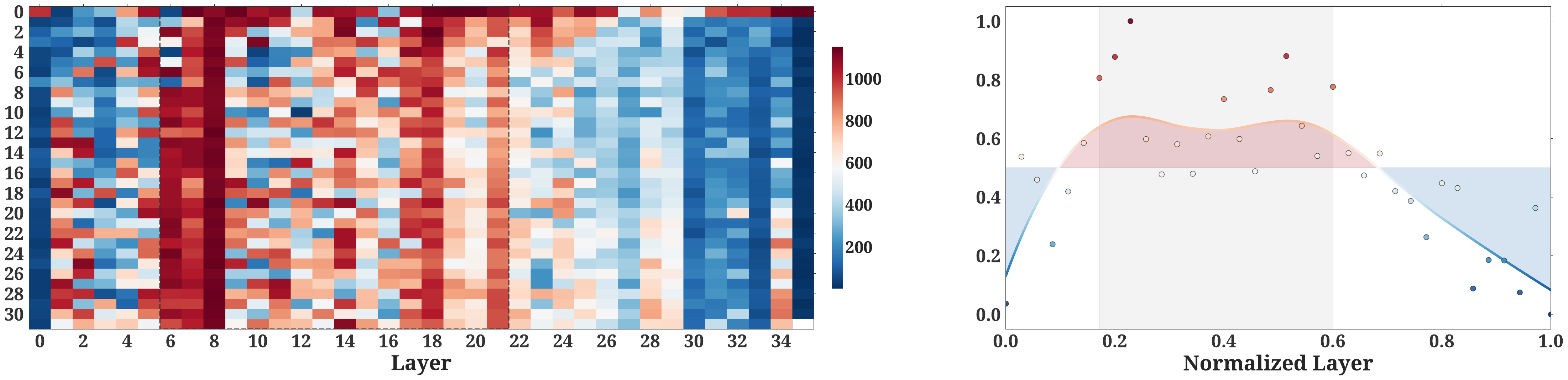}
    \caption{Head-level composition--preservation orientation}
\end{subfigure}

\begin{subfigure}[t]{0.92\textwidth}
    \centering
    \includegraphics[
        width=\linewidth,
        trim={0 16 0 8},
        clip
    ]{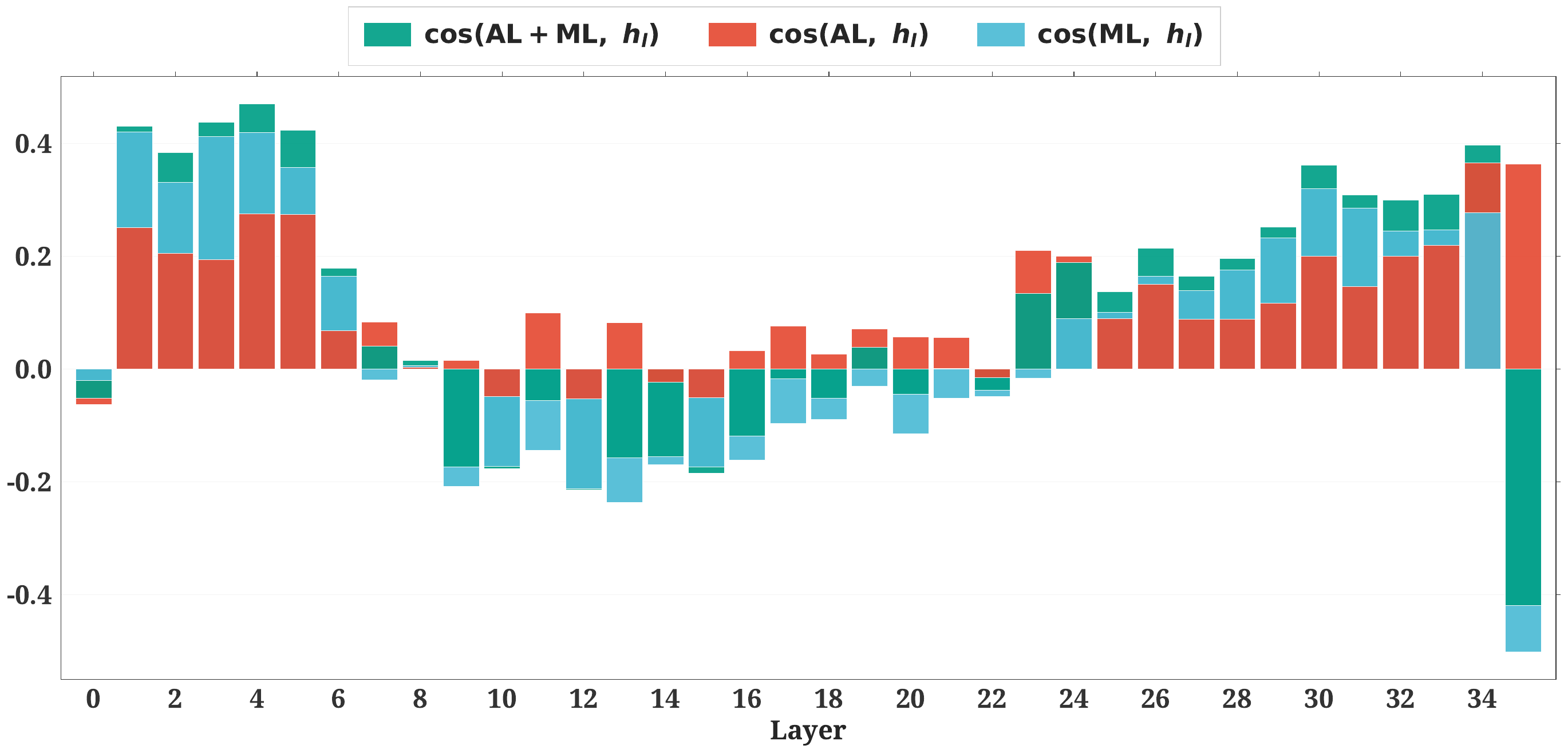}
    \caption{Cosine similarity of residual updates}
\end{subfigure}

\caption{
Layer-wise localization in Qwen3-4B-Base.
}
\label{fig:app_qwen3_4b_localization}
\end{figure}

\paragraph{Qwen3-4B-Base.}
Figure~\ref{fig:app_qwen3_4b_localization} shows the localization result for Qwen3-4B-Base.

\noindent\textbf{Analysis.}
Qwen3-4B-Base shows an early-to-middle composition-oriented region rather than a perfectly centered one. This is not unexpected for a smaller model. Since the model has fewer layers, representation transformation may need to begin earlier in normalized depth, leaving later layers to convert the transformed representation into prediction-relevant directions. In this sense, the location of the red region should not be interpreted by absolute layer index alone. What matters is that the composition-oriented dynamics are concentrated in an interior stage rather than being uniformly distributed over all layers.

The cosine profile supports this interpretation. Early layers show relatively positive alignment, suggesting that they mainly preserve or reinforce existing residual directions. In contrast, the middle region contains weaker and sometimes negative components, indicating that these layers are not simply adding information in the same direction but are changing the residual representation more substantially. This behavior is consistent with a representation-transformation stage, where the model reorganizes token-level features before they are routed toward prediction.

The late positive region suggests a return to prediction-oriented routing. After the intermediate transformation, later layers appear to write updates that are more aligned with the existing residual stream. This is consistent with the view that late layers are more involved in preparing the representation for next-token prediction. Thus, although the strongest red region is not exactly at the geometric midpoint, the result still supports the main claim: composition-oriented dynamics are concentrated in an interior computation stage rather than being dominated by the outermost layers.

\begin{figure}[H]
\centering
\captionsetup{font=small,skip=3pt}
\captionsetup[subfigure]{font=small,skip=2pt}

\begin{subfigure}[t]{0.92\textwidth}
    \centering
    \includegraphics[
        width=\linewidth,
        trim={0 10 0 8},
        clip
    ]{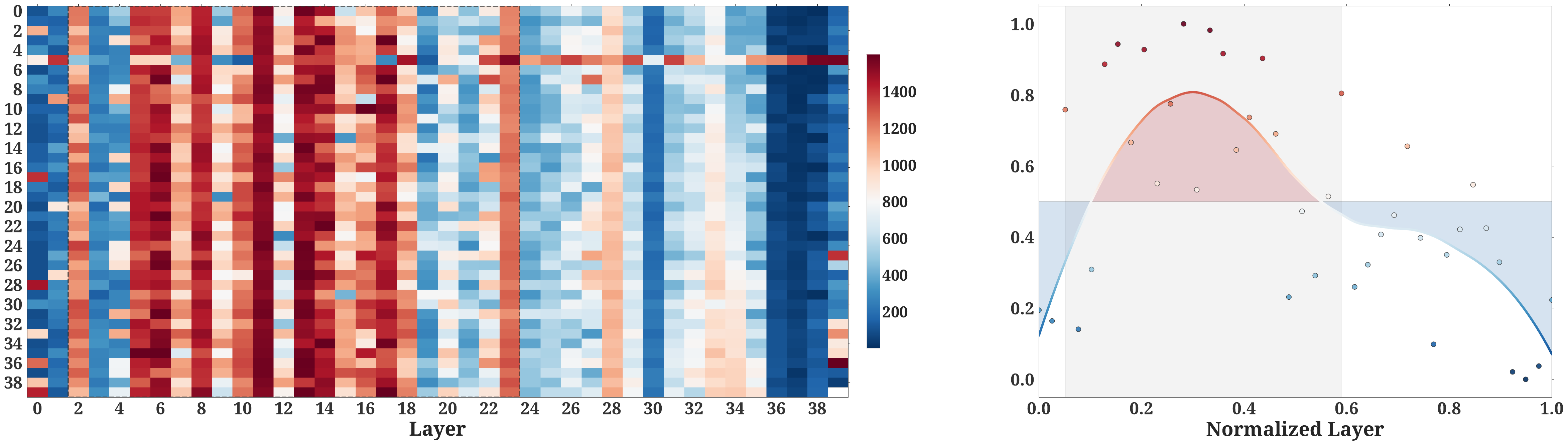}
    \caption{Head-level composition--preservation orientation}
\end{subfigure}

\begin{subfigure}[t]{0.92\textwidth}
    \centering
    \includegraphics[
        width=\linewidth,
        trim={0 16 0 8},
        clip
    ]{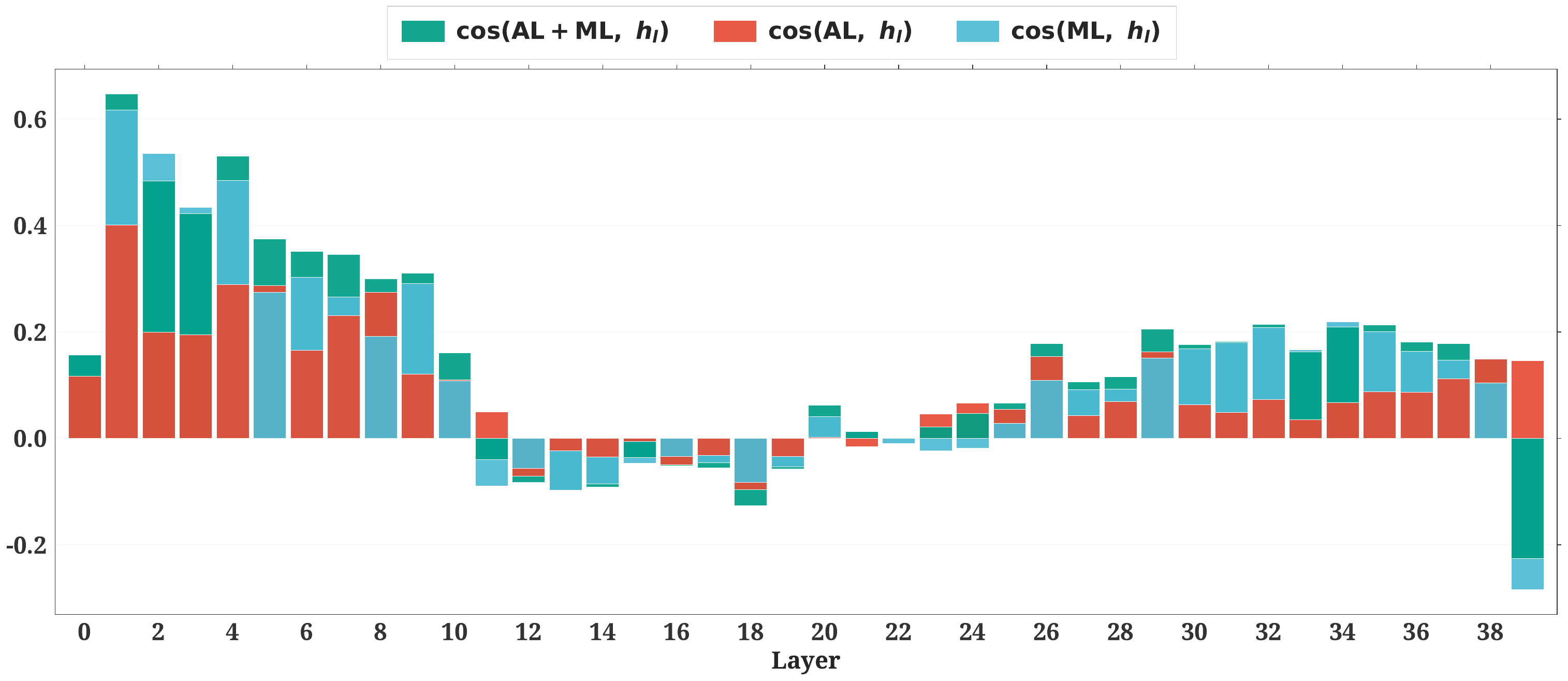}
    \caption{Cosine similarity of residual updates}
\end{subfigure}

\caption{
Layer-wise localization in Qwen3-14B-Base.
}
\label{fig:app_qwen3_14b_localization}
\end{figure}

\paragraph{Qwen3-14B-Base.}
Figure~\ref{fig:app_qwen3_14b_localization} reports the same analysis for Qwen3-14B-Base.

\noindent\textbf{Analysis.}
In Qwen3-14B-Base, the composition-oriented region is broader and appears over an early-to-middle depth range. This differs from Qwen3-4B-Base in absolute layer index, but the difference is natural because the 14B model has greater depth. A deeper model can distribute the same functional stage over more layers, so the relevant region need not be a narrow band at the exact center. The broader red region therefore suggests that the model may allocate a larger internal interval to representation transformation.

The cosine profile also differs from Qwen3-4B-Base. It shows strong positive alignment in early layers, followed by a middle interval with weaker alignment and then a later region with moderate positive values. This suggests a three-stage organization: early layers stabilize and route input-related information, intermediate layers transform the representation, and later layers re-align the residual stream toward prediction. The transition is smoother than in the smaller model, which is plausible given the increased depth and greater capacity of Qwen3-14B-Base.

Importantly, the localization result should be interpreted relative to model depth rather than by matching absolute layer numbers across models. A layer index that is early in Qwen3-14B-Base may correspond functionally to a more central region in a smaller model. Under this normalized-depth interpretation, Qwen3-14B-Base preserves the same qualitative organization as the main Qwen3-8B-Base result: representation transformation is concentrated away from the outermost layers, while the outer regions are more preservation-oriented.

\begin{figure}[H]
\centering
\captionsetup{font=small,skip=3pt}
\captionsetup[subfigure]{font=small,skip=2pt}

\begin{subfigure}[t]{0.92\textwidth}
    \centering
    \includegraphics[
        width=\linewidth,
        trim={0 10 0 8},
        clip
    ]{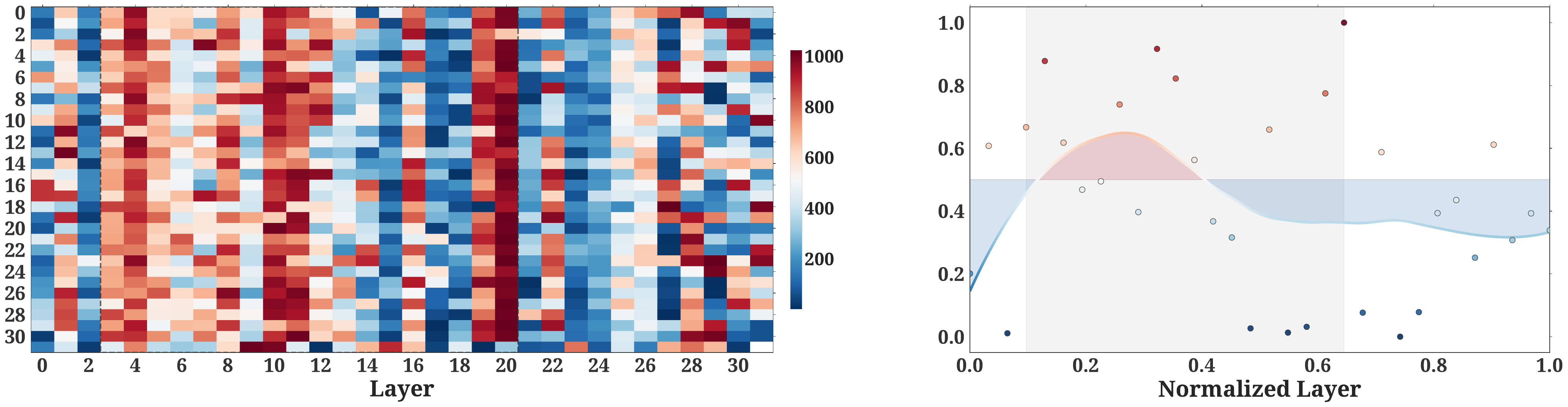}
    \caption{Head-level composition--preservation orientation}
\end{subfigure}

\begin{subfigure}[t]{0.92\textwidth}
    \centering
    \includegraphics[
        width=\linewidth,
        trim={0 18 0 10},
        clip
    ]{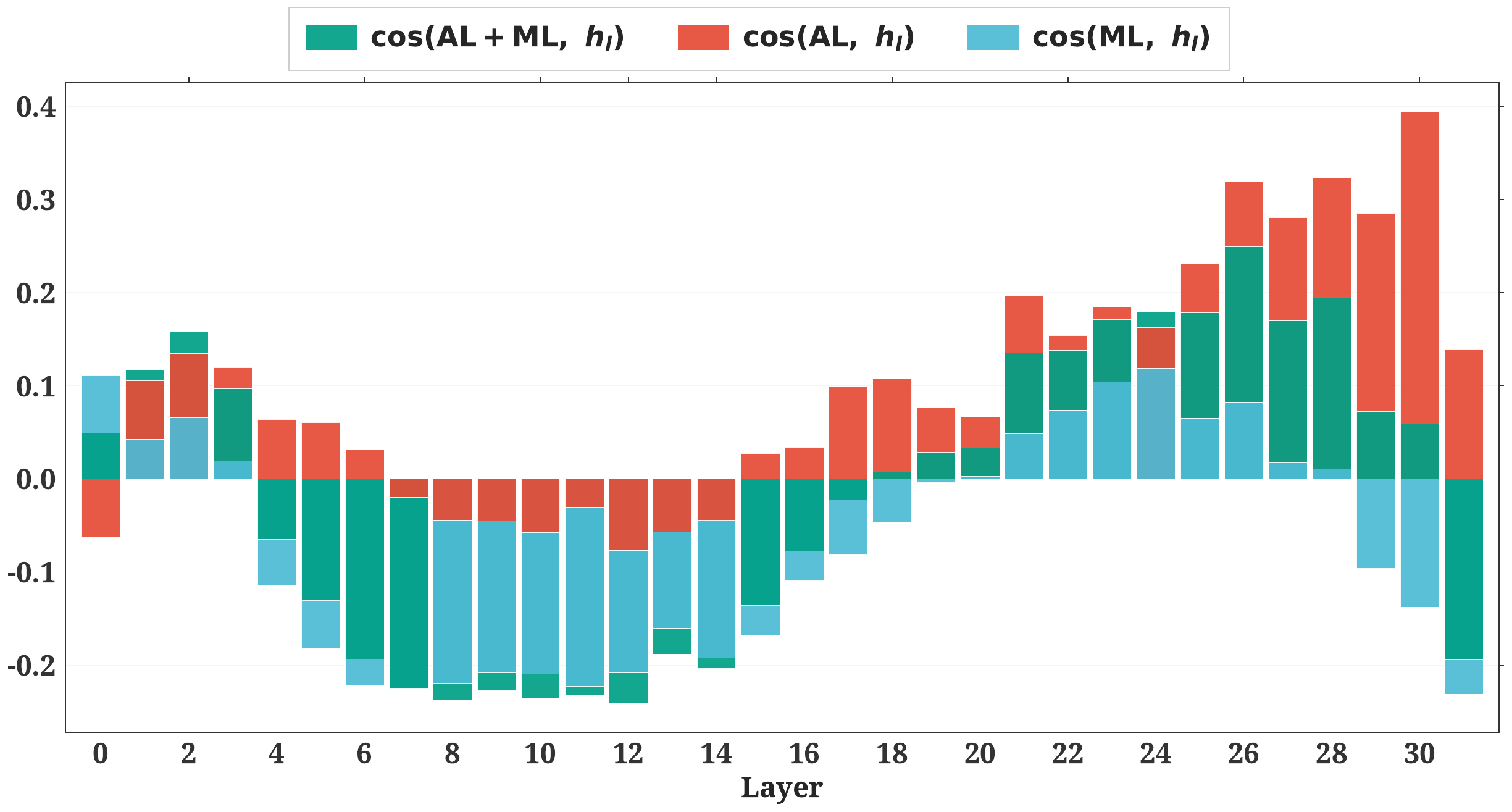}
    \caption{Cosine similarity of residual updates}
\end{subfigure}

\caption{
Layer-wise localization in Llama3.1-8B-Base.
}
\label{fig:app_llama31_8b_localization}
\end{figure}

\paragraph{Llama3.1-8B-Base.}
Figure~\ref{fig:app_llama31_8b_localization} shows the localization result for Llama3.1-8B-Base.

\noindent\textbf{Analysis.}
Llama3.1-8B-Base provides a useful cross-family comparison because its localization profile is not identical to the Qwen models. The heatmap shows a more interleaved structure: composition- and preservation-oriented heads alternate more strongly across intermediate layers. This suggests that the model may separate transformation and preservation roles at a finer layer scale, rather than assigning them to a single smooth contiguous block. Such behavior is reasonable for a different architecture and training recipe, and it highlights that the middle-stage organization should be understood as a functional regime rather than as a fixed layer interval.

The cosine profile is also distinctive. The middle layers contain a pronounced negative or suppressive region, followed by increasingly positive values in later layers. This indicates that intermediate layers are not merely accumulating residual information; they may rotate, suppress, or reorganize existing directions before the later layers recover a more prediction-aligned representation. This kind of directional reversal is particularly informative because it suggests that the intermediate layers actively reshape the residual stream, rather than only amplifying information already present in earlier layers.

Compared with Qwen3 models, the Llama profile therefore appears more ``transformational'' in the geometric sense. The negative cosine values imply that some layer updates oppose or substantially deviate from previous residual directions. This does not mean that these layers are harmful; rather, it suggests that they perform a nontrivial change of basis or reorganization of the internal representation. Such a process is compatible with abstract reasoning, where the model must move away from surface-token features and construct a representation that better supports relational computation.

This behavior strengthens rather than weakens the interpretation of a middle computation stage. The specific profile is architecture-dependent, but Llama3.1-8B-Base still shows that the strongest representation transformation is not located in the outermost layers. Instead, it appears in an interior depth range where the residual stream undergoes more substantial directional change. The later positive region then suggests that the transformed representation is gradually re-aligned with the output-facing computation required for next-token prediction.

\begin{figure}[H]
\centering
\captionsetup{font=small,skip=3pt}
\captionsetup[subfigure]{font=small,skip=2pt}

\begin{subfigure}[t]{0.92\textwidth}
    \centering
    \includegraphics[
        width=\linewidth,
        trim={0 10 0 8},
        clip
    ]{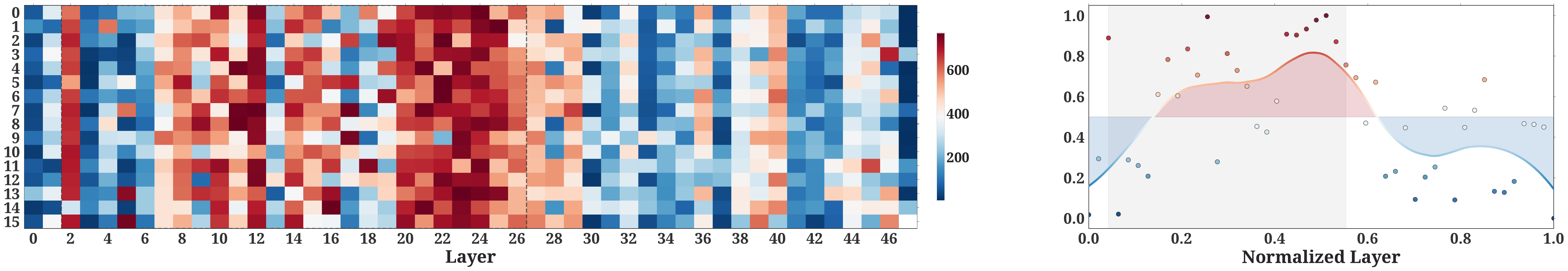}
    \caption{Head-level composition--preservation orientation}
\end{subfigure}

\begin{subfigure}[t]{0.92\textwidth}
    \centering
    \includegraphics[
        width=\linewidth,
        trim={0 18 0 10},
        clip
    ]{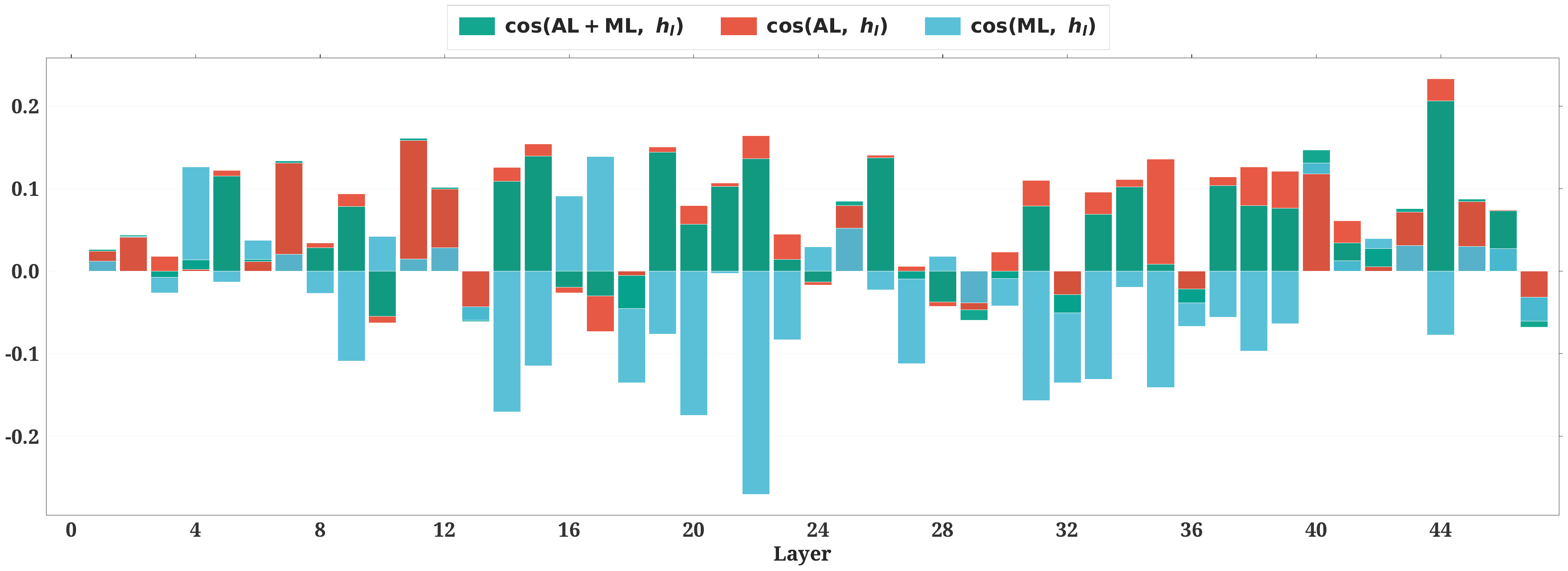}
    \caption{Cosine similarity of residual updates}
\end{subfigure}

\caption{
Layer-wise localization in Gemma3-12B-Base.
}
\label{fig:app_gemma3_12b_base_localization}
\end{figure}

\paragraph{Gemma3-12B-Instruct.}
Figure~\ref{fig:app_gemma3_12b_base_localization} provides the localization result for Gemma3-12B-Instruct.

\noindent\textbf{Analysis.}
Gemma3-12B-Instruct exhibits a broader and noisier localization profile than the Qwen models. The heatmap still shows an interior region with stronger composition-oriented dynamics, but the pattern is less sharply concentrated. This suggests that Gemma may distribute representation transformation across a wider set of internal layers. In other words, the abstraction-related computation may be less localized to a narrow band and instead spread over several neighboring internal regions.

The cosine profile is also more oscillatory. Instead of a smooth early-middle-late transition, the residual-update directions fluctuate across depth. This indicates that Gemma's residual stream may be shaped by more local transformations, with alternating preservation and transformation effects across neighboring layers. Such oscillations are not necessarily inconsistent with the main hypothesis. They suggest that different architectures may implement the same broad functional organization through different local update patterns.

The key point is that the outermost layers are still not the primary location of composition-oriented dynamics. Although the red region is not perfectly centered and the cosine curve is less smooth, the strongest transformation-related signals still appear inside the network. The model-dependent variability mainly affects where the interior region peaks and how sharply it is expressed, not whether such an interior region exists. This supports the broader conclusion that abstract-reasoning-related computation is organized in an internal stage rather than uniformly across all transformer layers.

\paragraph{Summary.}
Across these four additional models, the precise location and sharpness of the composition-oriented region vary. This variation is expected because the models differ in depth, architecture, tokenization, and training procedure. However, the qualitative organization is stable: composition-oriented dynamics are concentrated in an interior computation stage, whereas outer layers are more associated with preserving, routing, or re-aligning information for prediction. This cross-model pattern supports the claim that abstract-reasoning-related computation is not uniformly distributed across transformer depth.

\subsection{Cross-Model Layer-wise Interaction Profiles}
\label{app:cross_model_layer_profiles}

The main text reports layer-level interaction profiles under two sublayer settings: MLP-only ($ML$) and combined MLP-plus-attention ($ML+AL$). These profiles complement the head-level localization analysis by aggregating interaction dynamics at the layer level. The $ML$ setting isolates the contribution of MLP blocks, whereas the $ML+AL$ setting measures the combined effect of attention and MLP updates. Comparing the two settings allows us to examine whether the middle-stage organization is driven mainly by MLP transformations or by the interaction between attention and MLP computation.

\begin{figure}[H]
\centering
\captionsetup{font=small,skip=3pt}
\captionsetup[subfigure]{font=small,skip=2pt}

\begin{subfigure}[t]{0.92\textwidth}
    \centering
    \includegraphics[
        width=\linewidth,
        trim={0 10 0 8},
        clip
    ]{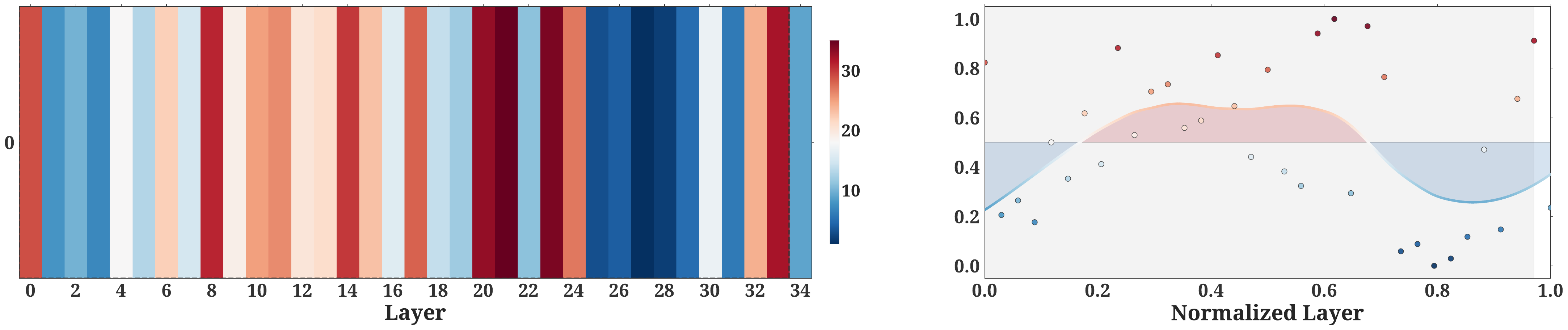}
    \caption{$ML$ interaction profile}
\end{subfigure}

\vspace{-0.1em}

\begin{subfigure}[t]{0.92\textwidth}
    \centering
    \includegraphics[
        width=\linewidth,
        trim={0 10 0 8},
        clip
    ]{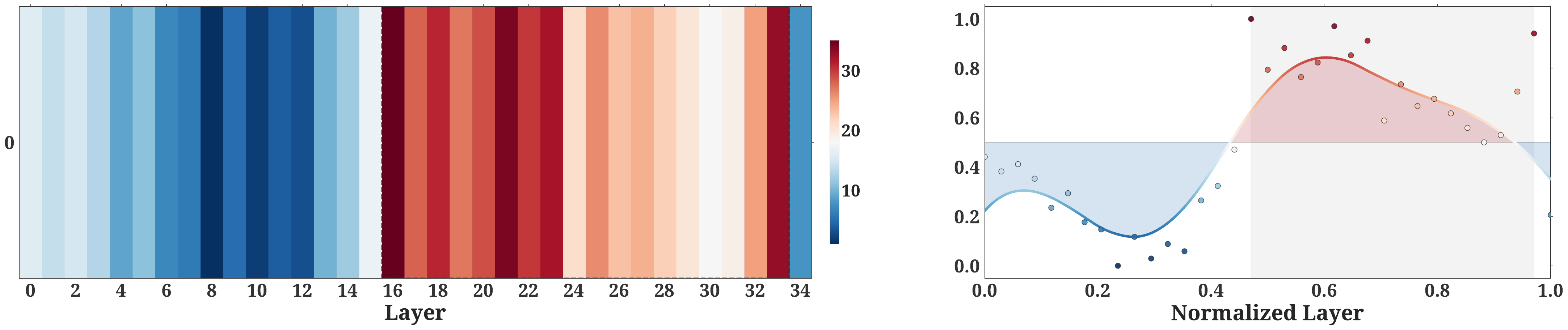}
    \caption{$ML+AL$ interaction profile}
\end{subfigure}

\vspace{-0.2em}

\caption{
Layer-wise interaction profiles in Qwen3-4B-Base.
}
\label{fig:app_qwen3_4b_layer_profiles}
\end{figure}

\paragraph{Qwen3-4B-Base.}
Figure~\ref{fig:app_qwen3_4b_layer_profiles} shows the layer-wise $ML$ and $ML+AL$ interaction profiles for Qwen3-4B-Base.

\noindent\textbf{Analysis.}
In Qwen3-4B-Base, the $ML$ profile shows a gradual transition across depth. The earlier layers contain stronger preservation-oriented signals, while the composition-oriented signal becomes more visible as the model enters the internal computation region. This indicates that MLP blocks alone already contribute to the depth-wise reorganization of information, but the transition is relatively smooth rather than sharply localized.

The $ML+AL$ profile is more structured. After attention updates are included, the composition-oriented region becomes more concentrated in the intermediate depth range. This suggests that attention and MLP updates are not simply independent contributors. Instead, attention may help route or select the relevant information that MLP blocks then transform, making the combined residual update more clearly reflect the middle-stage organization.

The difference between the two settings is important. If the middle-stage signal appeared only in the $ML$ profile, it would suggest a purely feed-forward transformation effect. If it appeared only in the attention-head heatmap, it could be interpreted as a head-level routing phenomenon. The fact that $ML+AL$ sharpens the pattern suggests that abstract-reasoning-related transformation is better understood as a coordinated layer-level process involving both attention and MLP computation.

\begin{figure}[H]
\centering
\captionsetup{font=small,skip=3pt}
\captionsetup[subfigure]{font=small,skip=2pt}

\begin{subfigure}[t]{0.92\textwidth}
    \centering
    \includegraphics[
        width=\linewidth,
        trim={0 10 0 8},
        clip
    ]{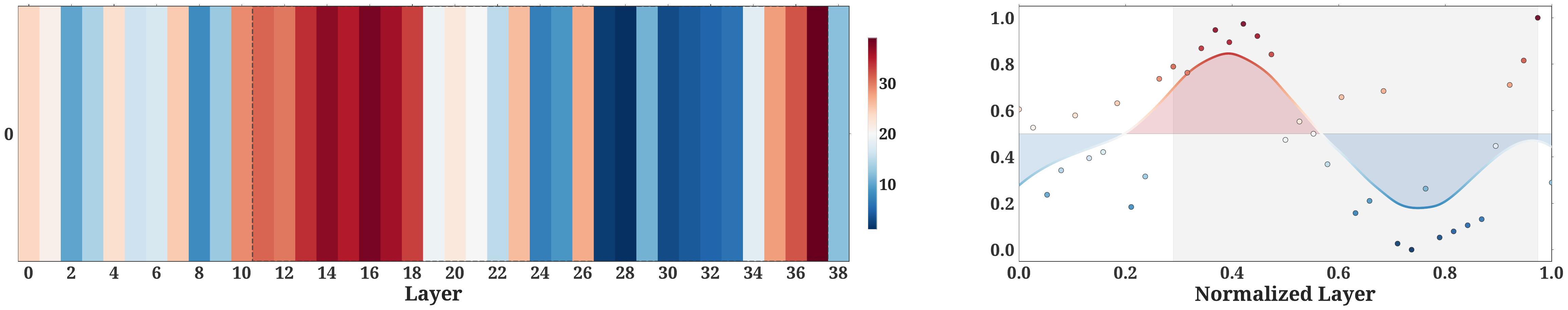}
    \caption{$ML$ interaction profile}
\end{subfigure}

\begin{subfigure}[t]{0.92\textwidth}
    \centering
    \includegraphics[
        width=\linewidth,
        trim={0 10 0 8},
        clip
    ]{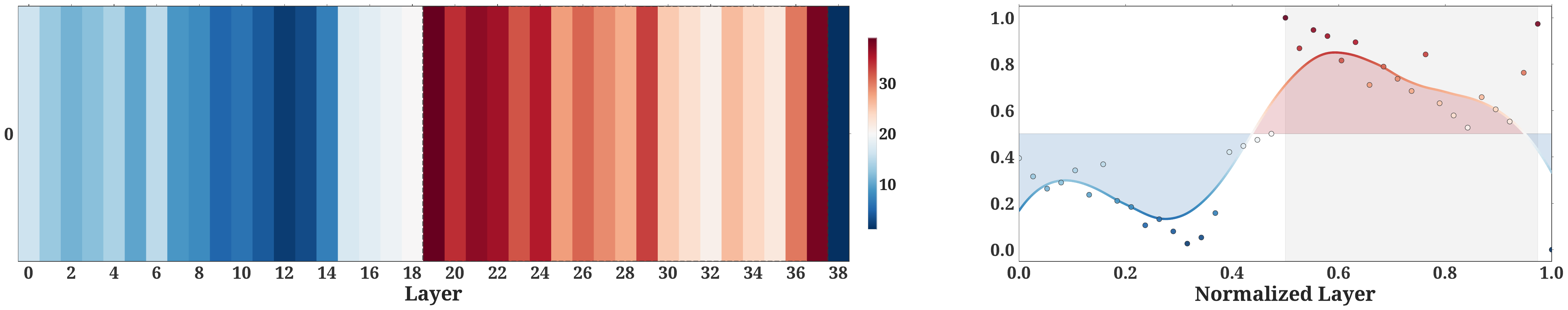}
    \caption{$ML+AL$ interaction profile}
\end{subfigure}

\caption{
Layer-wise interaction profiles in Qwen3-14B-Base.
}
\label{fig:app_qwen3_14b_layer_profiles}
\end{figure}

\paragraph{Qwen3-14B-Base.}
Figure~\ref{fig:app_qwen3_14b_layer_profiles} reports the same layer-level profiles for Qwen3-14B-Base.

\noindent\textbf{Analysis.}
Qwen3-14B-Base shows a broader interaction profile than Qwen3-4B-Base. This is expected because the model is deeper, and the internal computation stage can occupy a wider interval in absolute layer index. In the $ML$ setting, composition-oriented signals do not collapse into a single narrow region. Instead, they spread across a larger internal range, suggesting that feed-forward transformations are distributed across multiple layers.

The $ML+AL$ setting again makes the intermediate organization more pronounced. Compared with the $ML$ profile, the combined profile better separates the internal composition-oriented region from the outer preservation-oriented regions. This supports the view that attention and MLP blocks jointly define the effective transformation stage, rather than each sublayer independently producing the entire abstraction-related pattern.

This model is useful because it shows that the middle-stage hypothesis should not be interpreted as a fixed number of layers. In a larger model, the relevant computation can be stretched over a broader normalized-depth interval. The important observation is that the composition-oriented region remains away from the outermost layers, while the initial and final regions are more associated with preserving, routing, or re-aligning information.

\begin{figure}[H]
\centering
\captionsetup{font=small,skip=3pt}
\captionsetup[subfigure]{font=small,skip=2pt}

\begin{subfigure}[t]{0.92\textwidth}
    \centering
    \includegraphics[
        width=\linewidth,
        trim={0 10 0 8},
        clip
    ]{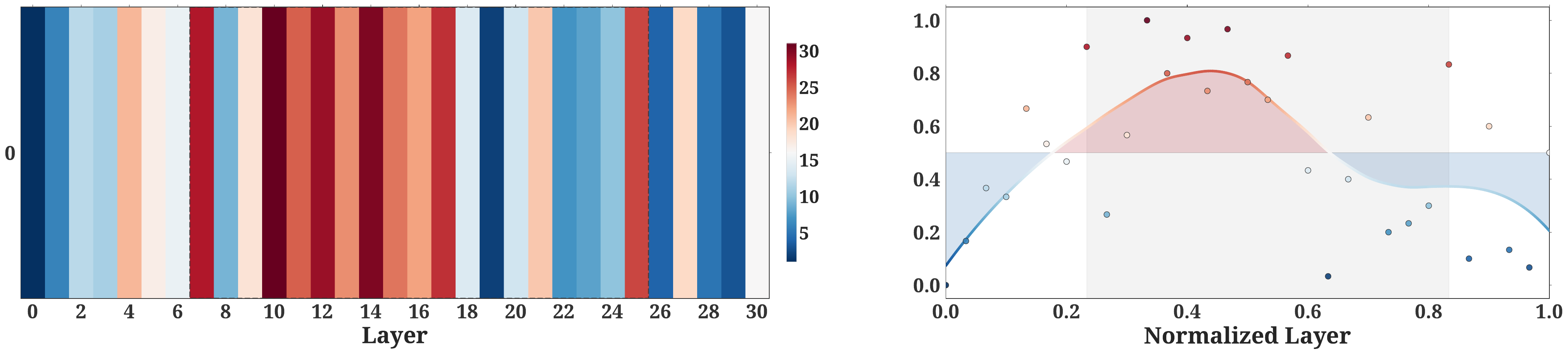}
    \caption{$ML$ interaction profile}
\end{subfigure}

\begin{subfigure}[t]{0.92\textwidth}
    \centering
    \includegraphics[
        width=\linewidth,
        trim={0 10 0 8},
        clip
    ]{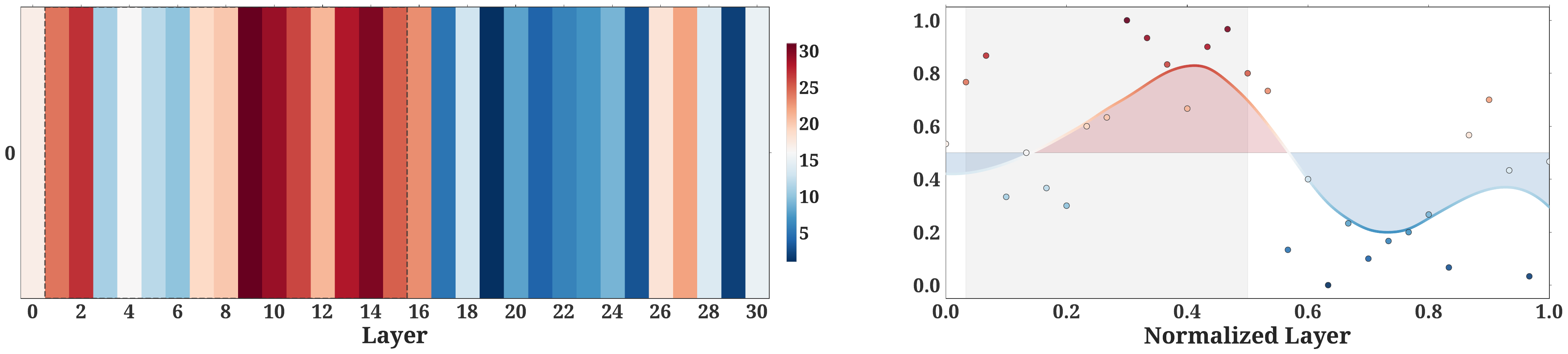}
    \caption{$ML+AL$ interaction profile}
\end{subfigure}

\caption{
Layer-wise interaction profiles in Llama3.1-8B-Base.
}
\label{fig:app_llama31_8b_layer_profiles}
\end{figure}

\paragraph{Llama3.1-8B-Base.}
Figure~\ref{fig:app_llama31_8b_layer_profiles} shows the $ML$ and $ML+AL$ profiles for Llama3.1-8B-Base.

\noindent\textbf{Analysis.}
Llama3.1-8B-Base exhibits a more architecture-specific profile than the Qwen models. In the $ML$ setting, the layer-wise transition is less smooth and contains stronger local variation. This suggests that MLP transformations in Llama are distributed in a more alternating manner across depth, rather than forming a single continuous band.

The $ML+AL$ profile provides an important complementary view. When attention updates are added, the interaction pattern becomes more interpretable as an internal transformation stage. The intermediate layers show stronger composition-oriented behavior than the outer regions, even though the profile is more oscillatory than in Qwen models. This matches the observation from the cosine-similarity analysis, where Llama exhibits stronger directional reorganization in the middle layers.

The difference between $ML$ and $ML+AL$ is especially informative for Llama3.1-8B-Base. It suggests that the representation-transformation stage may depend more strongly on the coupling between attention and MLP updates. Attention alone may route information across positions, while MLP blocks reshape local representations; their combination produces a clearer signature of abstract computation. Therefore, even though the exact layer-wise pattern differs from Qwen, the same broader organization remains visible.

\begin{figure}[H]
\centering
\captionsetup{font=small,skip=3pt}
\captionsetup[subfigure]{font=small,skip=2pt}

\begin{subfigure}[t]{0.92\textwidth}
    \centering
    \includegraphics[
        width=\linewidth,
        trim={0 10 0 8},
        clip
    ]{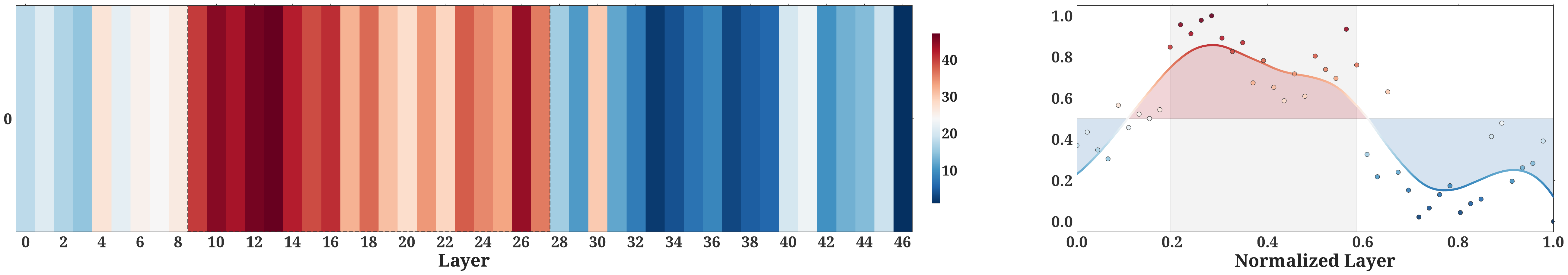}
    \caption{$ML$ interaction profile}
\end{subfigure}

\begin{subfigure}[t]{0.92\textwidth}
    \centering
    \includegraphics[
        width=\linewidth,
        trim={0 10 0 8},
        clip
    ]{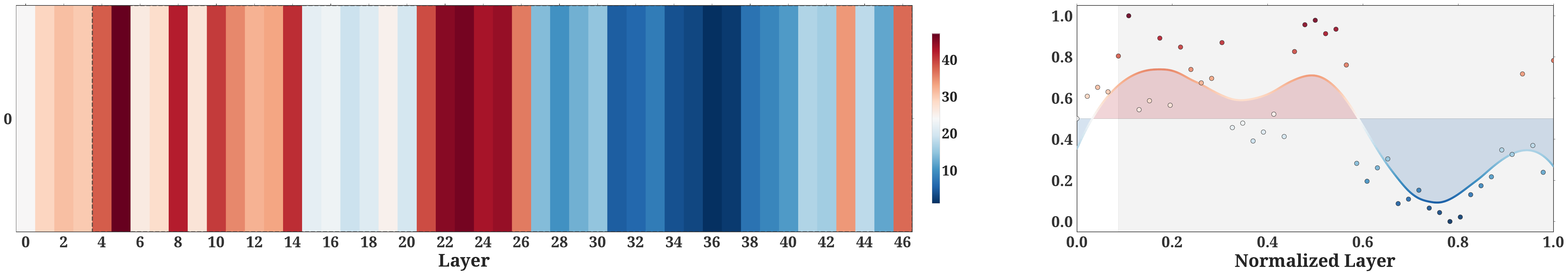}
    \caption{$ML+AL$ interaction profile}
\end{subfigure}

\caption{
Layer-wise interaction profiles in Gemma3-12B-Base.
}
\label{fig:app_gemma3_12b_base_layer_profiles}
\end{figure}

\paragraph{Gemma3-12B-Instruct.}
Figure~\ref{fig:app_gemma3_12b_base_layer_profiles} provides the corresponding profiles for Gemma3-12B-Instruct.

\noindent\textbf{Analysis.}
Gemma3-12B-Instruct shows a broader and less sharply localized profile than the Qwen models. In the $ML$ setting, composition- and preservation-oriented signals fluctuate across depth, indicating that MLP transformations are distributed more locally. This is consistent with the cosine-similarity results, where Gemma also exhibits a more oscillatory residual-update pattern.

The $ML+AL$ setting still reveals an internal computation region, but the pattern remains less smooth than in Qwen3 models. This suggests that Gemma may implement representation transformation through several local update stages rather than through one sharply concentrated middle band. Such behavior does not contradict the main hypothesis, because the hypothesis concerns the existence of an interior transformation stage, not the requirement that it be perfectly contiguous or centered.

The key observation is that the strongest transformation-related signals are still not concentrated in the outermost layers. Even though Gemma's profile is noisier, the combined attention-and-MLP view shows that internal layers participate more strongly in composition-oriented dynamics. This supports the same qualitative conclusion as the other models: layer-level abstraction-related computation is organized across depth rather than uniformly distributed.

\paragraph{Summary.}
Across the four additional models, the $ML$ and $ML+AL$ profiles provide complementary evidence for a non-uniform layer-wise organization. The $ML$ setting often shows a smoother or more distributed transformation pattern, reflecting feed-forward contributions to representation reshaping. The $ML+AL$ setting usually sharpens the intermediate region, suggesting that attention and MLP updates jointly produce the most visible abstraction-related signal.

The exact profile varies across architectures. Qwen models show a comparatively clearer middle-stage organization, Llama3.1-8B-Base shows stronger internal reorientation, and Gemma3-12B-Base shows a broader and more oscillatory profile. Despite these differences, all models support the same qualitative conclusion: representation transformation is concentrated in an interior computation stage, while outer layers more strongly preserve, route, or re-align information for prediction.

\subsection{Cross-Model Rule-Level Representations}
\label{app:cross_model_rule_representations}

The main text reports representation-geometry and probing results on Qwen3-8B-Base. Here we repeat the same analysis on additional models. For each model, we show four complementary measurements: intrinsic dimension, cross-vocabulary CKA, cross-vocabulary rule-probe accuracy, and visible token-identity probe accuracy. Together, these measurements test whether intermediate layers form compact and vocabulary-invariant rule-level representations.

\begin{figure}[H]
\centering
\captionsetup{font=small,skip=3pt}
\captionsetup[subfigure]{font=small,skip=2pt}

\begin{subfigure}[t]{0.48\textwidth}
    \centering
    \includegraphics[width=\linewidth]{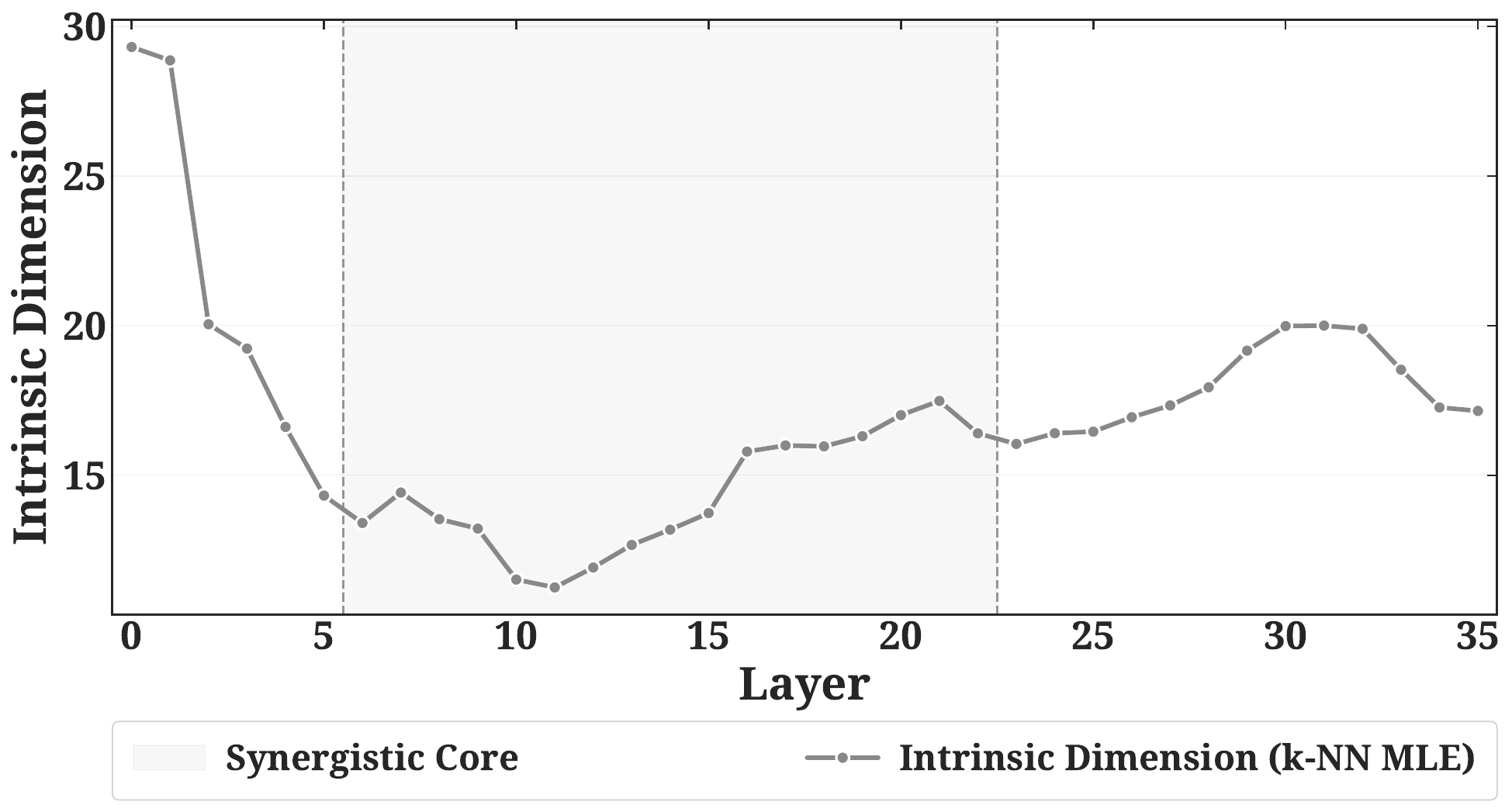}
    \caption{Intrinsic dimension}
\end{subfigure}
\hfill
\begin{subfigure}[t]{0.48\textwidth}
    \centering
    \includegraphics[width=\linewidth]{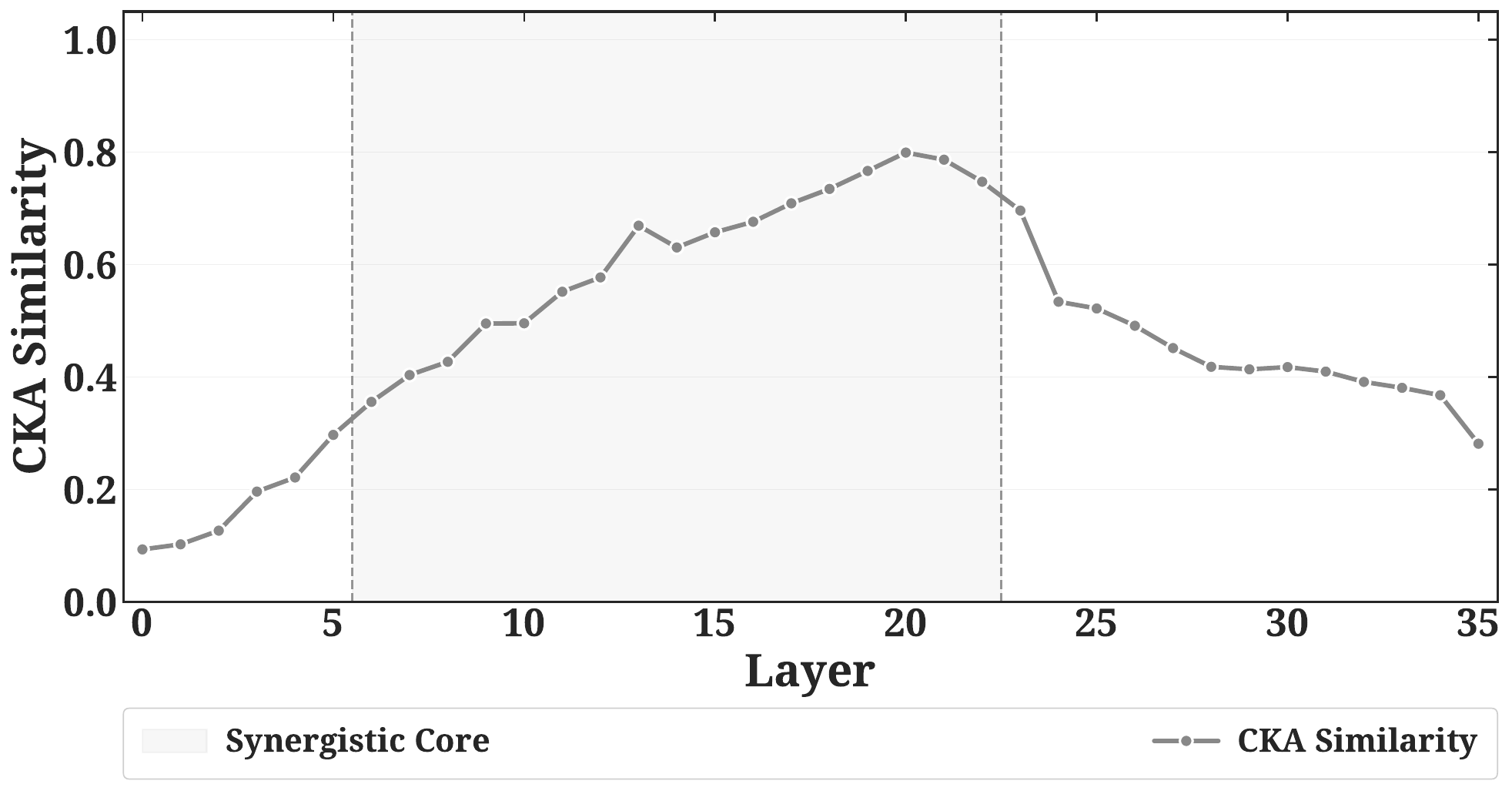}
    \caption{Cross-vocabulary CKA}
\end{subfigure}

\begin{subfigure}[t]{0.48\textwidth}
    \centering
    \includegraphics[width=\linewidth]{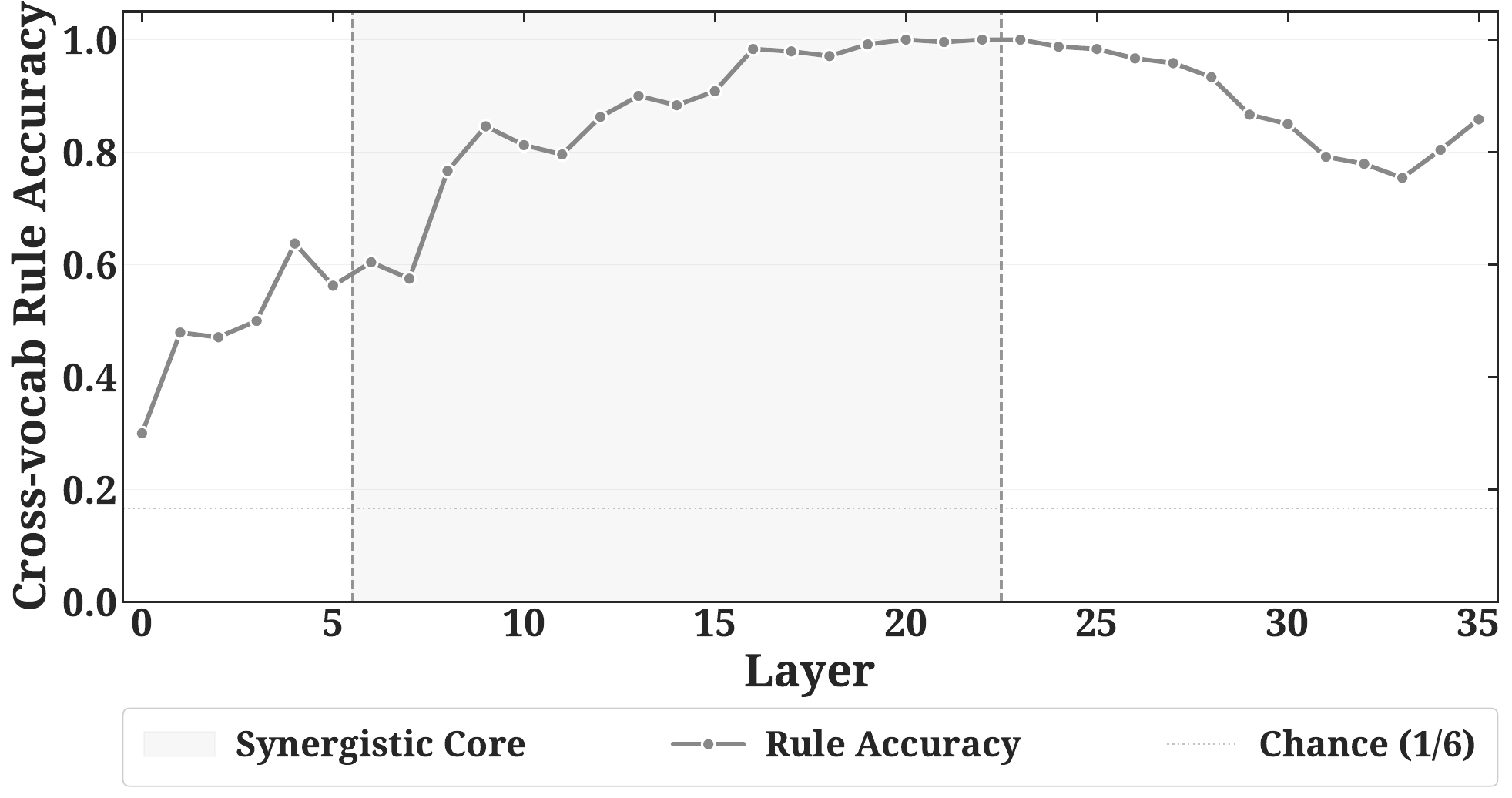}
    \caption{Rule-probe accuracy}
\end{subfigure}
\hfill
\begin{subfigure}[t]{0.48\textwidth}
    \centering
    \includegraphics[width=\linewidth]{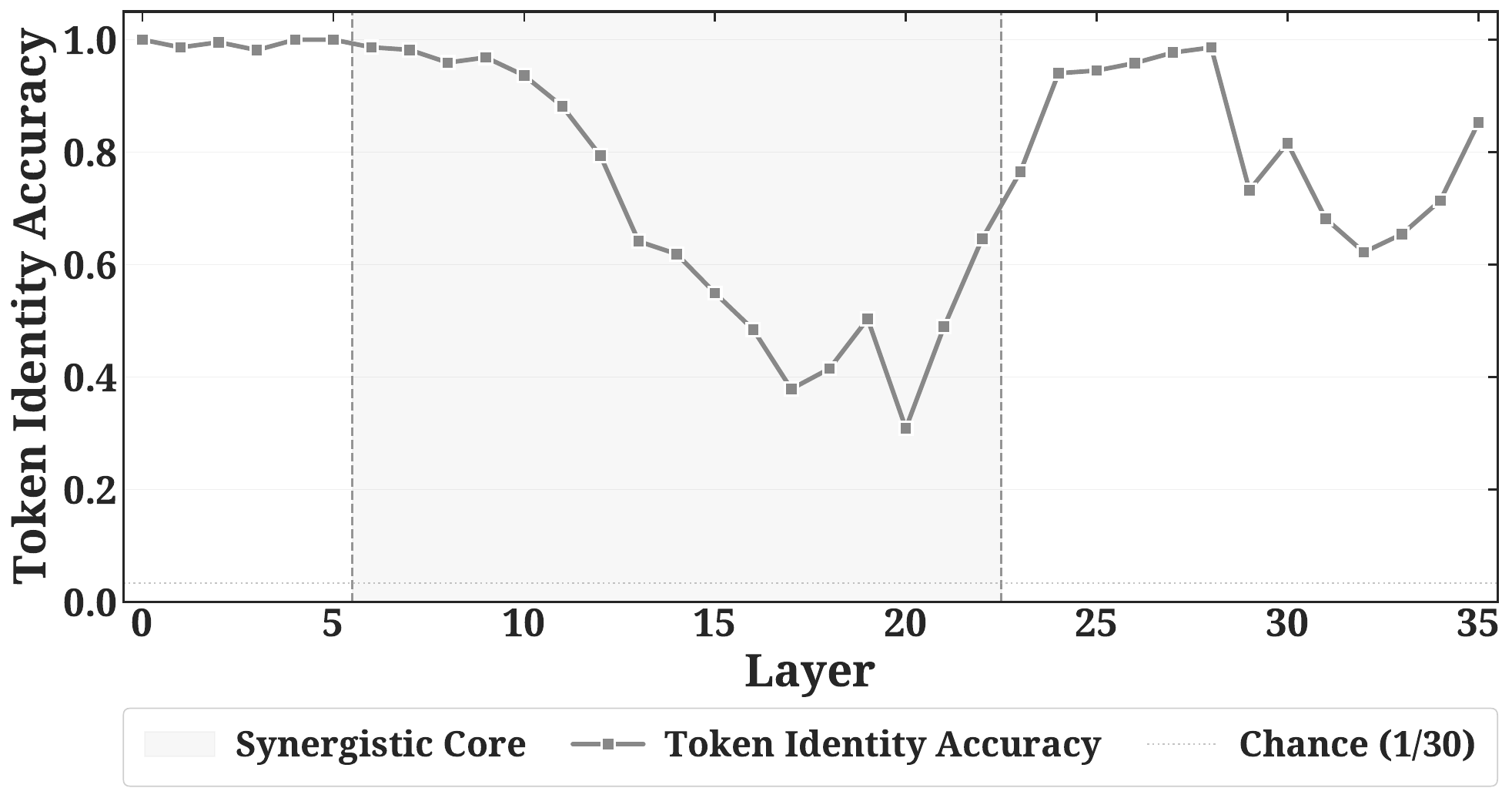}
    \caption{Token-probe accuracy}
\end{subfigure}

\caption{
Rule-level representation analysis in Qwen3-4B-Base.
}
\label{fig:app_qwen3_4b_rule_representations}
\end{figure}

\paragraph{Qwen3-4B-Base.}
Figure~\ref{fig:app_qwen3_4b_rule_representations} shows the rule-level representation results for Qwen3-4B-Base.

\noindent\textbf{Analysis.}
Qwen3-4B-Base shows the same qualitative representation pattern as the main Qwen3-8B-Base model, although the effect is expressed over a shallower depth range. The intrinsic-dimension curve decreases in an intermediate region, suggesting that hidden states become locally more compact during the internal computation stage. This supports the view that the model does not simply accumulate token-level features across layers, but compresses them into a lower-dimensional representation.

The CKA curve provides a complementary test of whether this compactness reflects lexical identity or abstract structure. Since the two vocabulary groups are disjoint, high cross-vocabulary CKA cannot be explained by shared surface tokens. The intermediate increase in CKA therefore suggests that representations induced by different vocabularies become more aligned when they instantiate the same relational rule.

The probe results sharpen this interpretation. Rule-probe accuracy increases in the same broad region where the intrinsic dimension is lower and CKA is higher, indicating that rule information is more linearly recoverable from layer updates. In contrast, token-probe accuracy does not show the same pattern, suggesting that the layer updates are less dominated by visible token identity. Thus, even in the smaller Qwen3 model, the middle computation stage appears to transform surface-form information into more reusable rule-level structure.

\begin{figure}[H]
\centering
\captionsetup{font=small,skip=3pt}
\captionsetup[subfigure]{font=small,skip=2pt}

\begin{subfigure}[t]{0.48\textwidth}
    \centering
    \includegraphics[width=\linewidth]{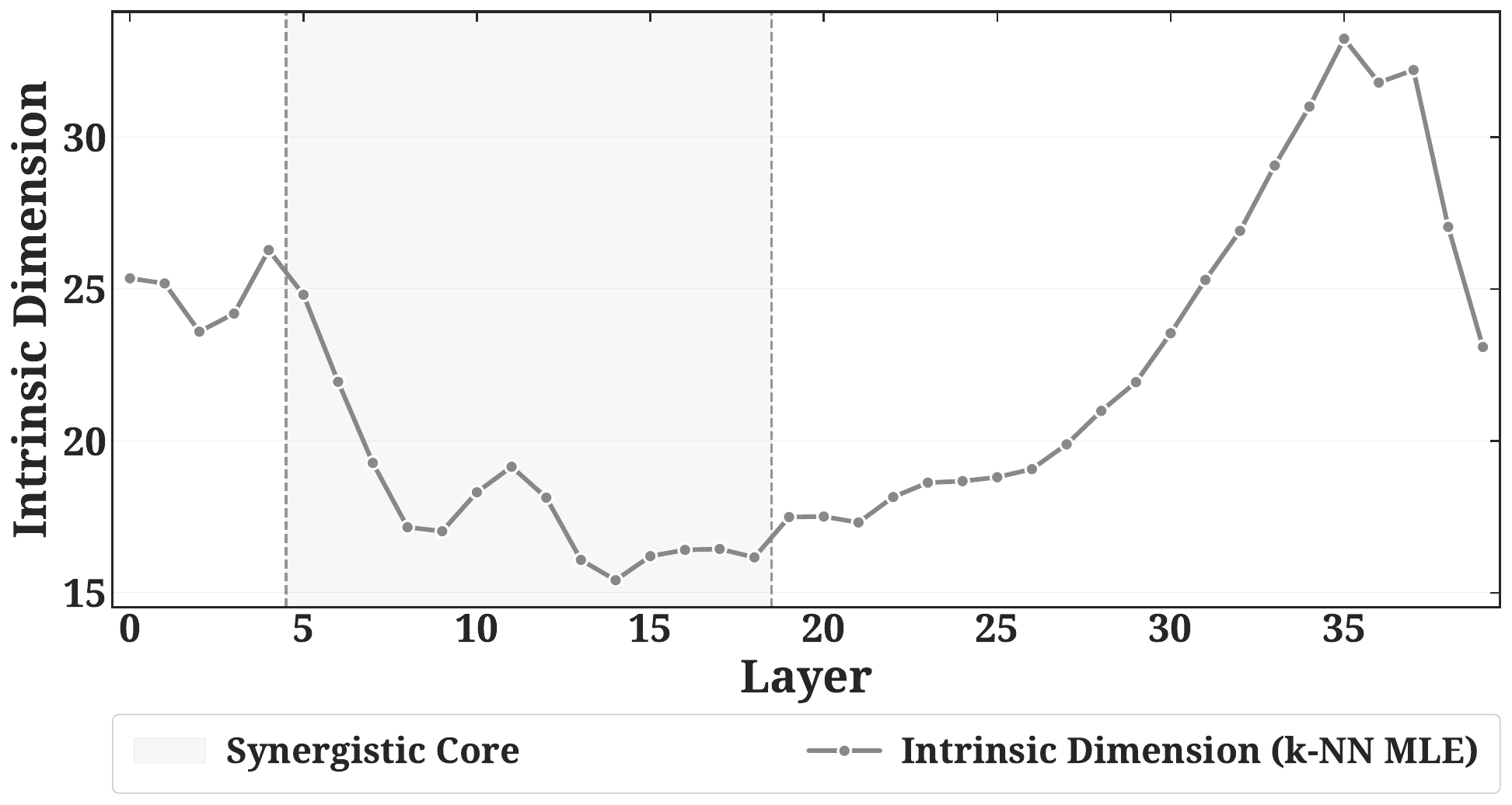}
    \caption{Intrinsic dimension}
\end{subfigure}
\hfill
\begin{subfigure}[t]{0.48\textwidth}
    \centering
    \includegraphics[width=\linewidth]{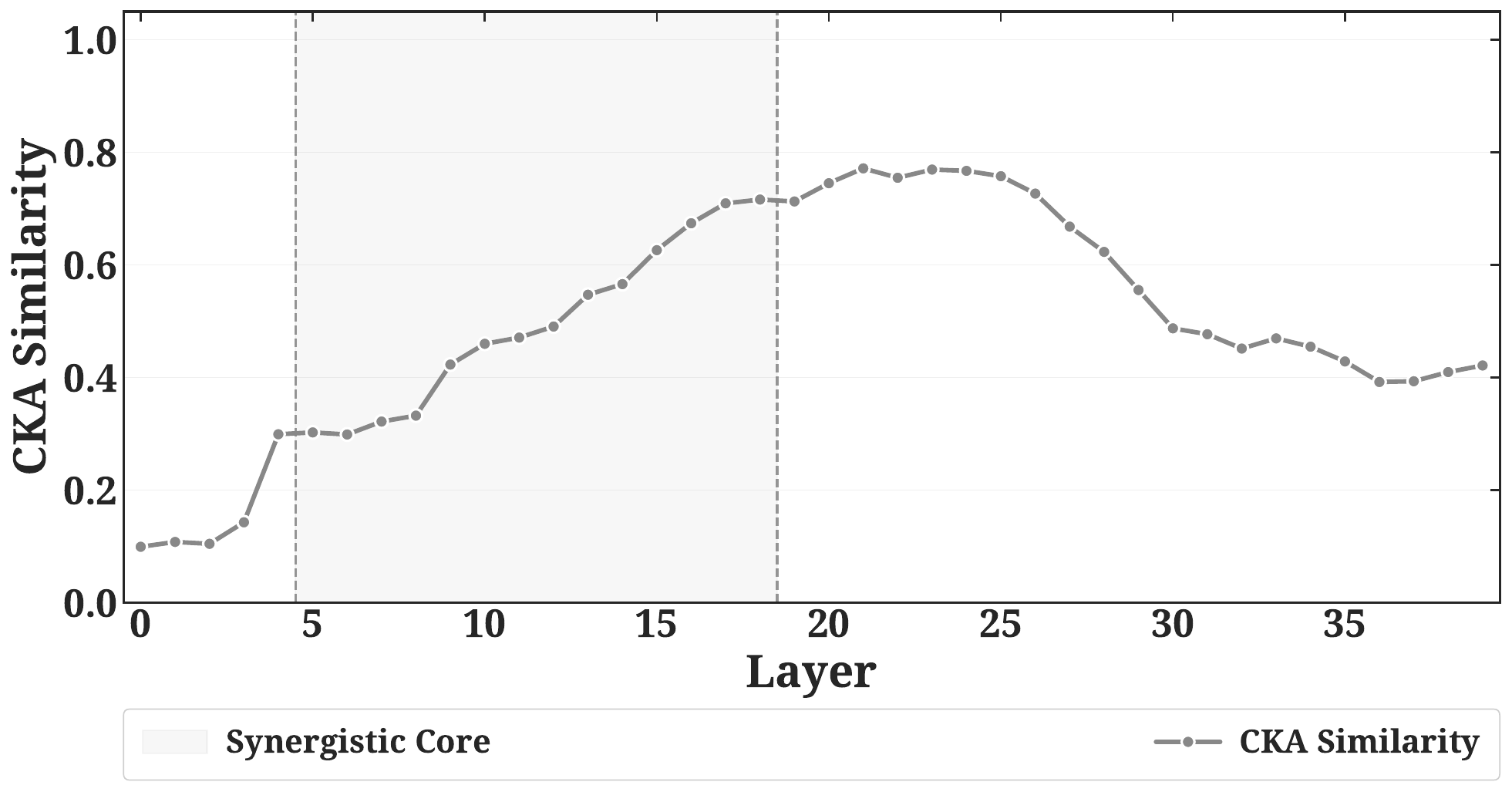}
    \caption{Cross-vocabulary CKA}
\end{subfigure}

\begin{subfigure}[t]{0.48\textwidth}
    \centering
    \includegraphics[width=\linewidth]{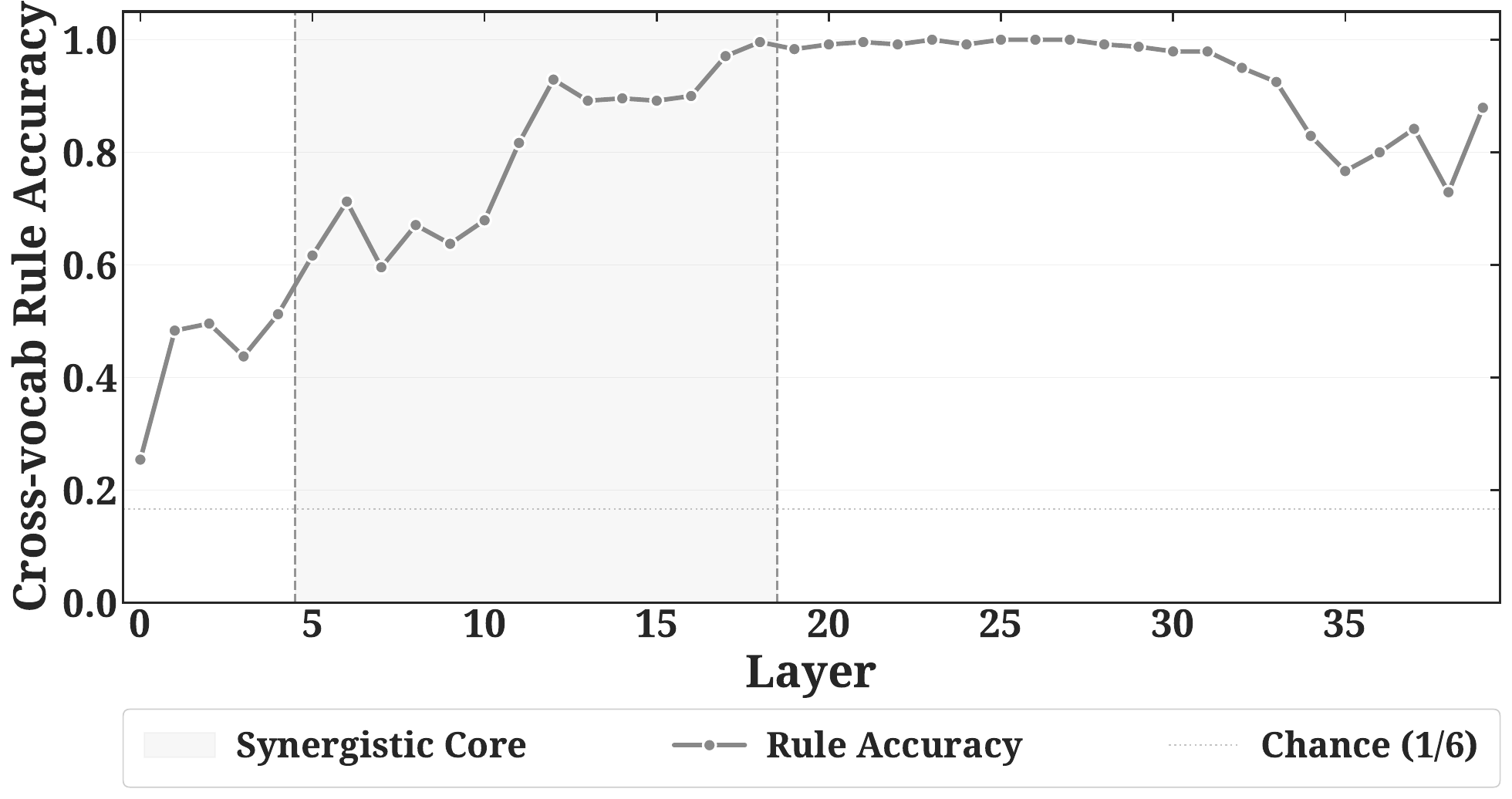}
    \caption{Rule-probe accuracy}
\end{subfigure}
\hfill
\begin{subfigure}[t]{0.48\textwidth}
    \centering
    \includegraphics[width=\linewidth]{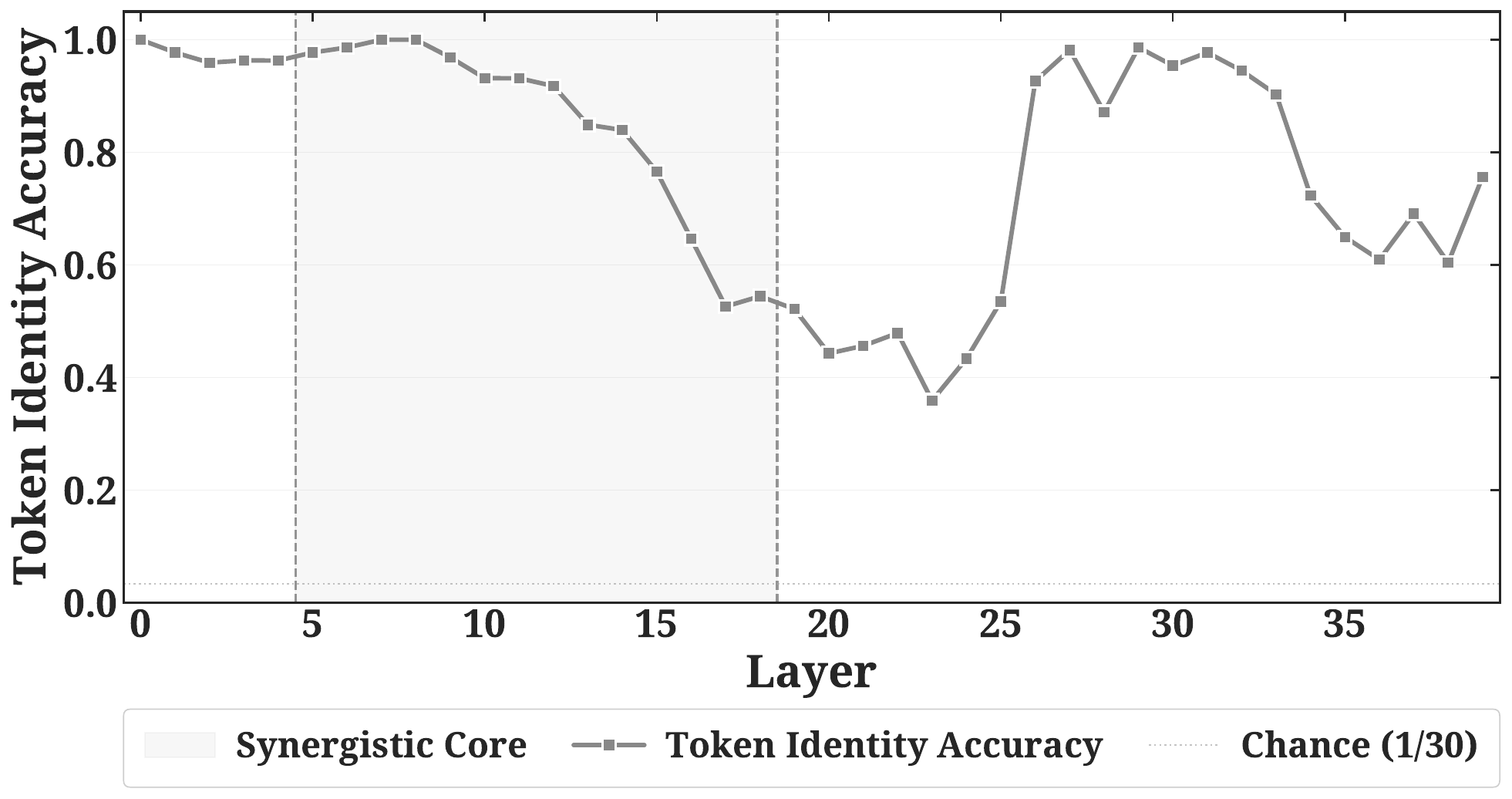}
    \caption{Token-probe accuracy}
\end{subfigure}

\caption{
Rule-level representation analysis in Qwen3-14B-Base.
}
\label{fig:app_qwen3_14b_rule_representations}
\end{figure}

\paragraph{Qwen3-14B-Base.}
Figure~\ref{fig:app_qwen3_14b_rule_representations} reports the same analysis for Qwen3-14B-Base.

\noindent\textbf{Analysis.}
Qwen3-14B-Base exhibits a broader representation-geometry pattern than the smaller Qwen3 models. This is consistent with its greater depth: rather than concentrating rule-level transformation into a narrow band, the model can distribute the relevant computation across a wider internal interval. The intrinsic-dimension curve still supports the main conclusion, because the most compact representation region appears away from the outermost layers.

The cross-vocabulary CKA results further indicate that this compact region is not merely a byproduct of reduced activation variance. When representations from disjoint vocabularies become more aligned, the alignment must be mediated by shared relational structure rather than token identity. Thus, high CKA in the internal region suggests that Qwen3-14B-Base develops a vocabulary-invariant representation of the underlying rule.

The rule and token probes again show a useful dissociation. Rule-probe accuracy is stronger in the internal region, while token-probe accuracy is comparatively less aligned with the same pattern. This suggests that layer updates in this depth range are more informative about abstract rule identity than about the visible surface token. Therefore, the larger Qwen3 model preserves the same qualitative mechanism as Qwen3-8B-Base, but stretches it over a broader depth range.

\begin{figure}[H]
\centering
\captionsetup{font=small,skip=3pt}
\captionsetup[subfigure]{font=small,skip=2pt}

\begin{subfigure}[t]{0.48\textwidth}
    \centering
    \includegraphics[width=\linewidth]{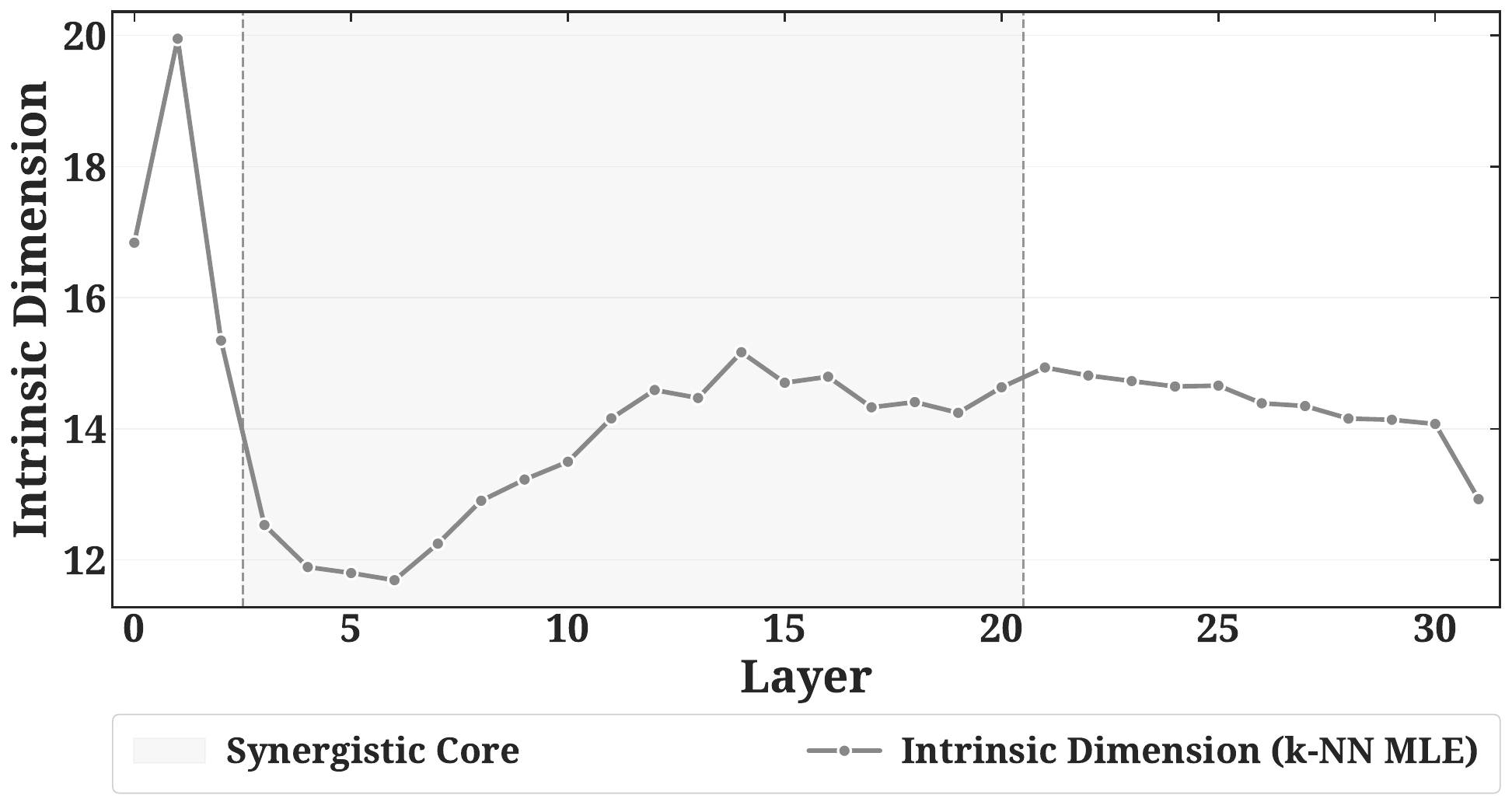}
    \caption{Intrinsic dimension}
\end{subfigure}
\hfill
\begin{subfigure}[t]{0.48\textwidth}
    \centering
    \includegraphics[width=\linewidth]{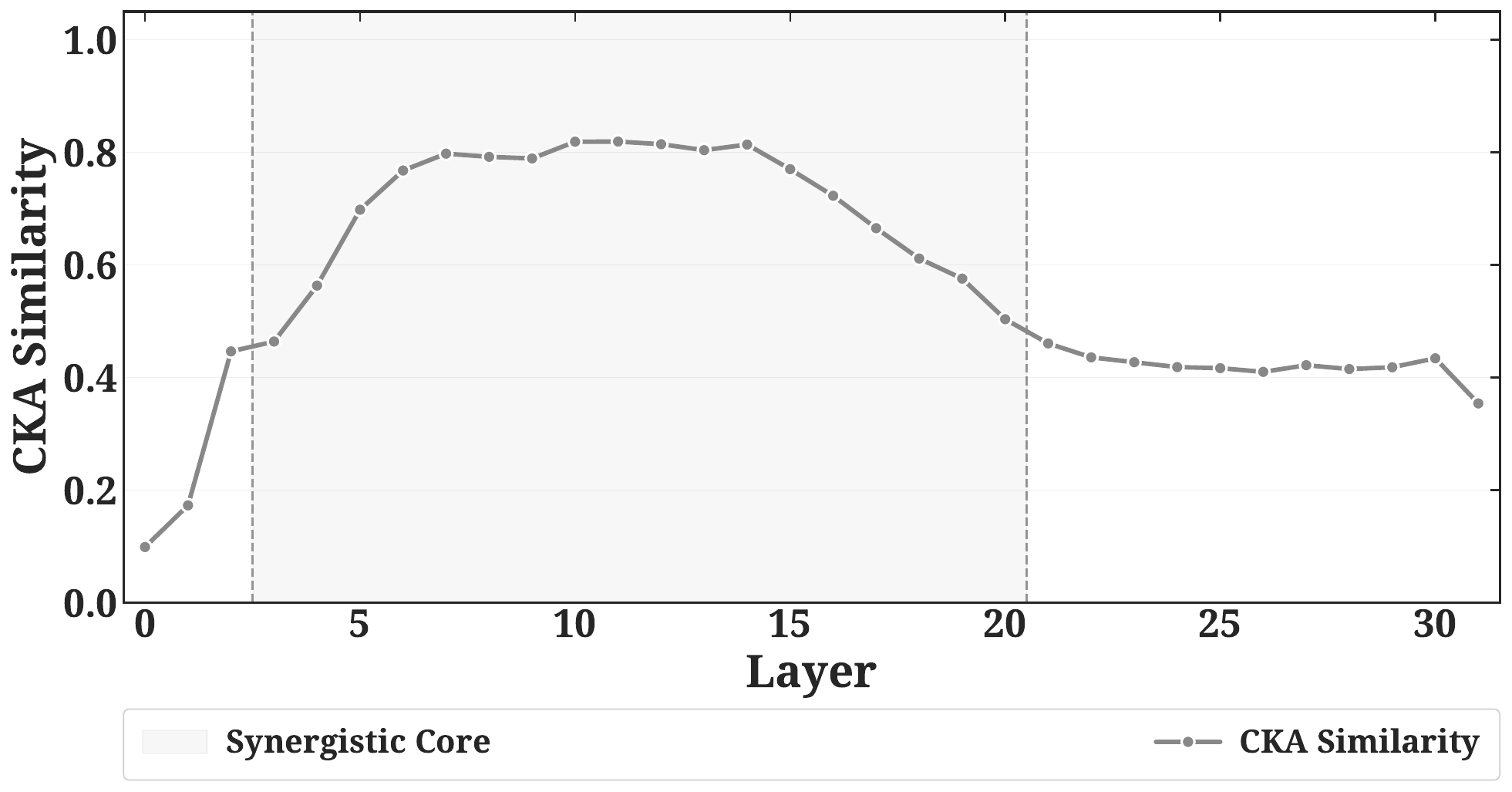}
    \caption{Cross-vocabulary CKA}
\end{subfigure}

\begin{subfigure}[t]{0.48\textwidth}
    \centering
    \includegraphics[width=\linewidth]{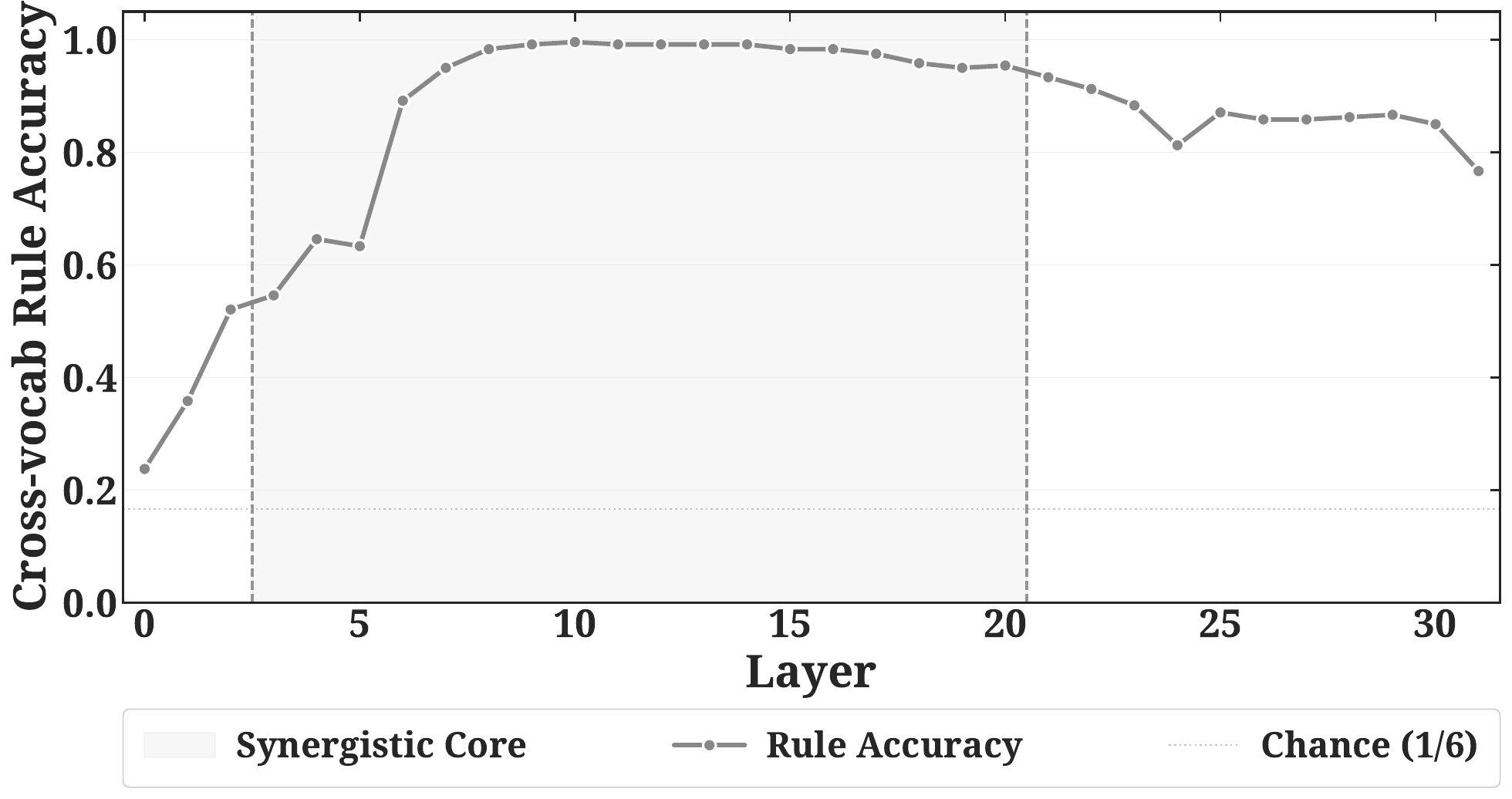}
    \caption{Rule-probe accuracy}
\end{subfigure}
\hfill
\begin{subfigure}[t]{0.48\textwidth}
    \centering
    \includegraphics[width=\linewidth]{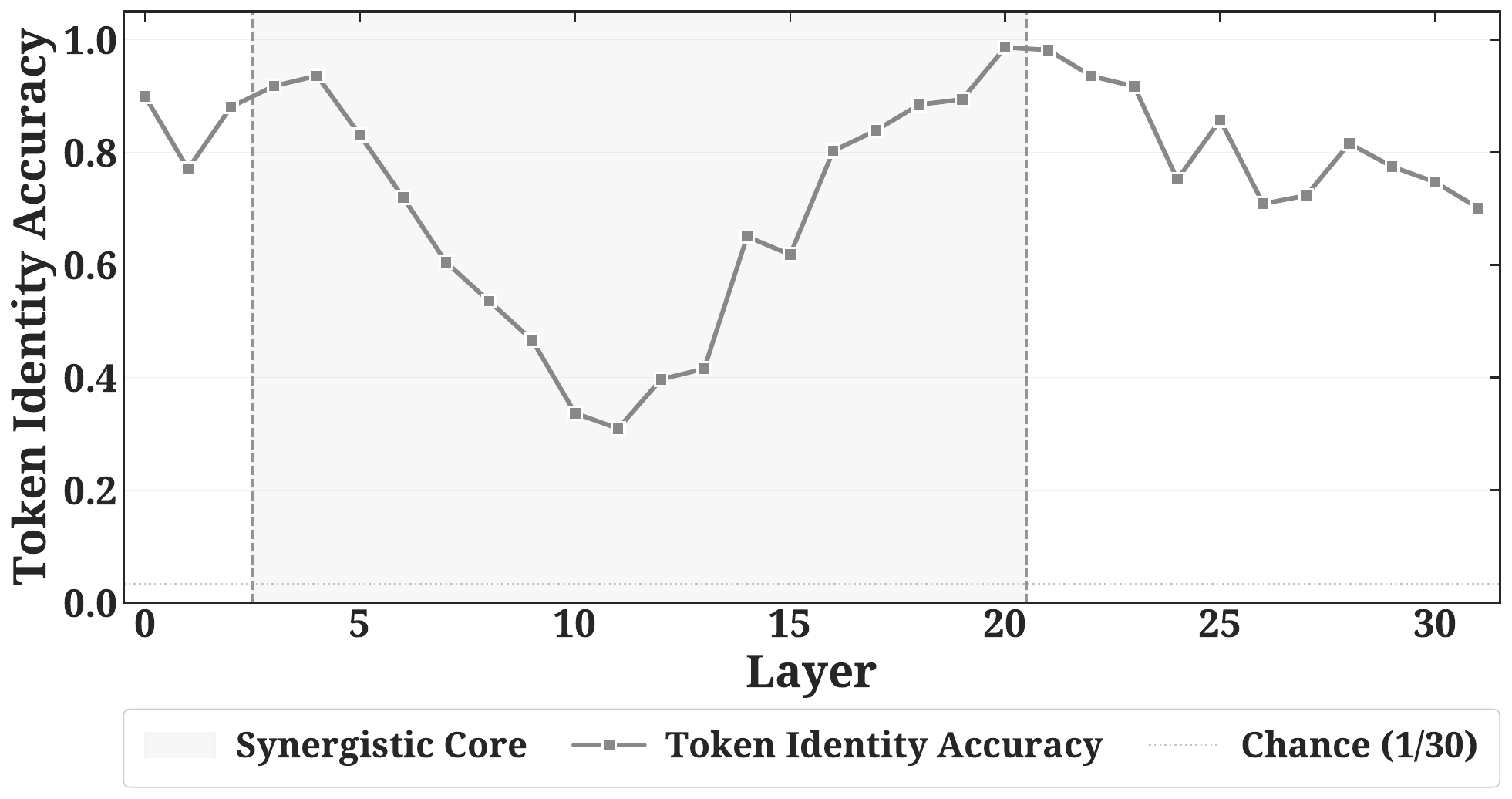}
    \caption{Token-probe accuracy}
\end{subfigure}

\caption{
Rule-level representation analysis in Llama3.1-8B-Base.
}
\label{fig:app_llama31_8b_rule_representations}
\end{figure}

\paragraph{Llama3.1-8B-Base.}
Figure~\ref{fig:app_llama31_8b_rule_representations} shows the rule-level representation results for Llama3.1-8B-Base.

\noindent\textbf{Analysis.}
Llama3.1-8B-Base provides a cross-family test of the representation hypothesis. Its geometry is not expected to match Qwen models layer by layer, but the same qualitative relationship between compactness, vocabulary-invariant alignment, and rule recoverability remains informative. The intrinsic-dimension curve indicates that the representation space becomes more constrained in an internal region, which is consistent with the idea that the model compresses token-level variation during reasoning.

The CKA result is particularly important for Llama3.1-8B-Base because its localization profile is more oscillatory than Qwen's. Even if composition- and preservation-oriented heads are more interleaved across layers, high cross-vocabulary alignment still indicates that the model can organize different surface vocabularies around the same abstract rule. This suggests that the rule-level representation is not tied to the exact same layer pattern as in Qwen, but emerges from a broader internal transformation process.

The probe results further support this interpretation. Higher rule-probe accuracy in the internal region shows that rule labels become more recoverable from layer updates, while the token-probe control helps distinguish this effect from surface-token memorization. Thus, the Llama results suggest that different model families may implement the abstraction stage with different local dynamics, but still produce a comparable vocabulary-invariant rule representation.

\begin{figure}[H]
\centering
\captionsetup{font=small,skip=3pt}
\captionsetup[subfigure]{font=small,skip=2pt}

\begin{subfigure}[t]{0.48\textwidth}
    \centering
    \includegraphics[width=\linewidth]{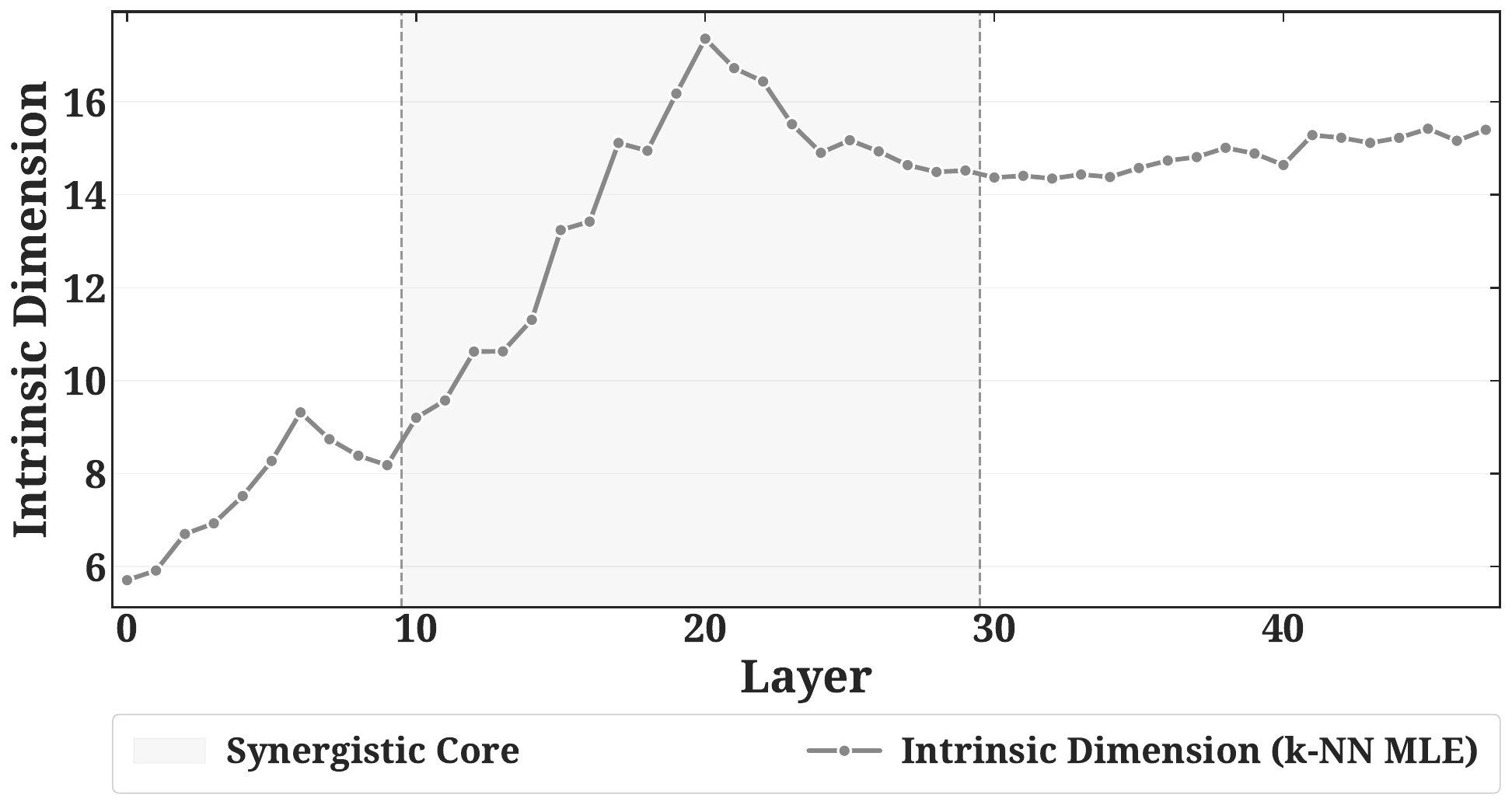}
    \caption{Intrinsic dimension}
\end{subfigure}
\hfill
\begin{subfigure}[t]{0.48\textwidth}
    \centering
    \includegraphics[width=\linewidth]{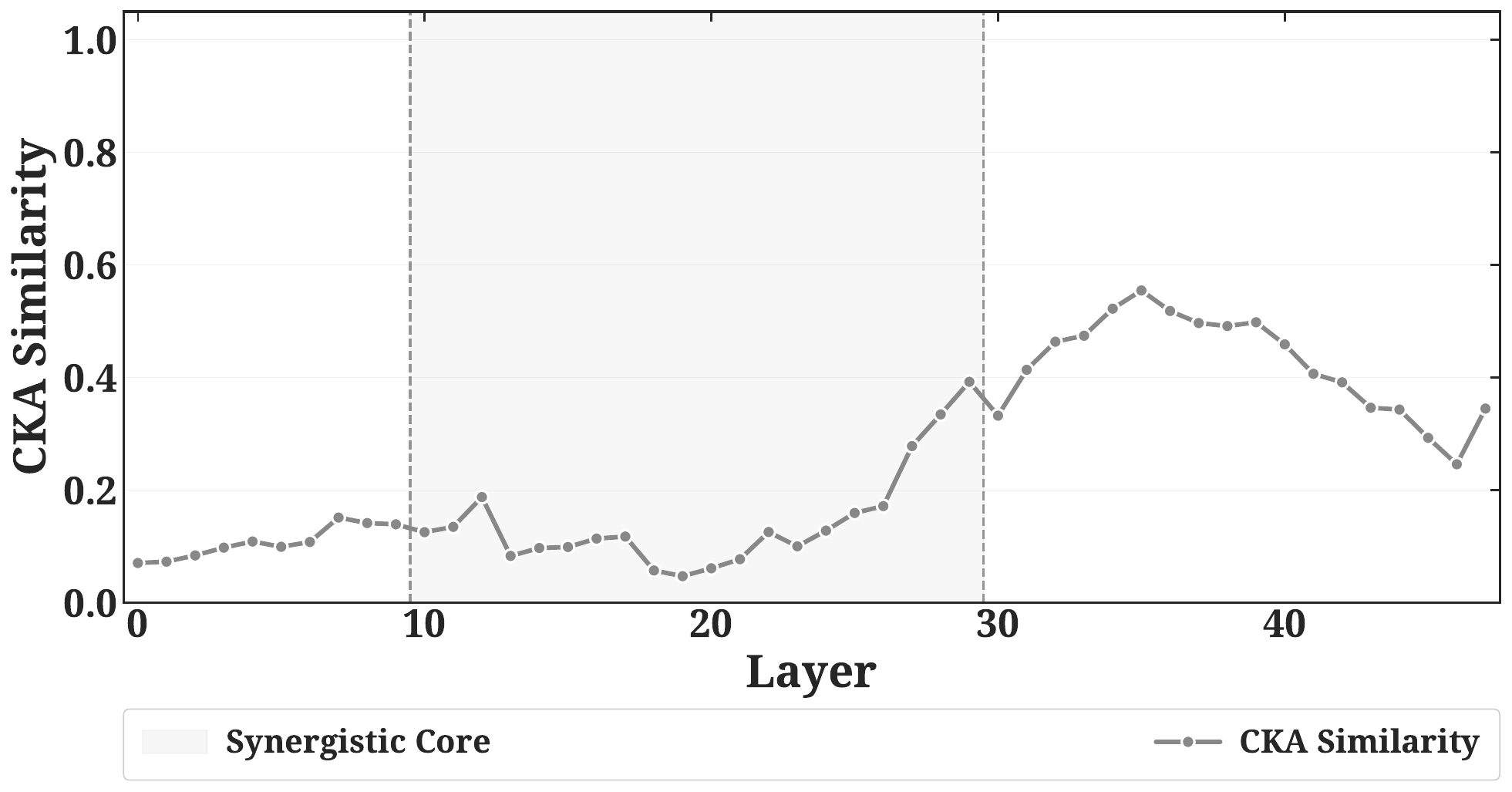}
    \caption{Cross-vocabulary CKA}
\end{subfigure}

\begin{subfigure}[t]{0.48\textwidth}
    \centering
    \includegraphics[width=\linewidth]{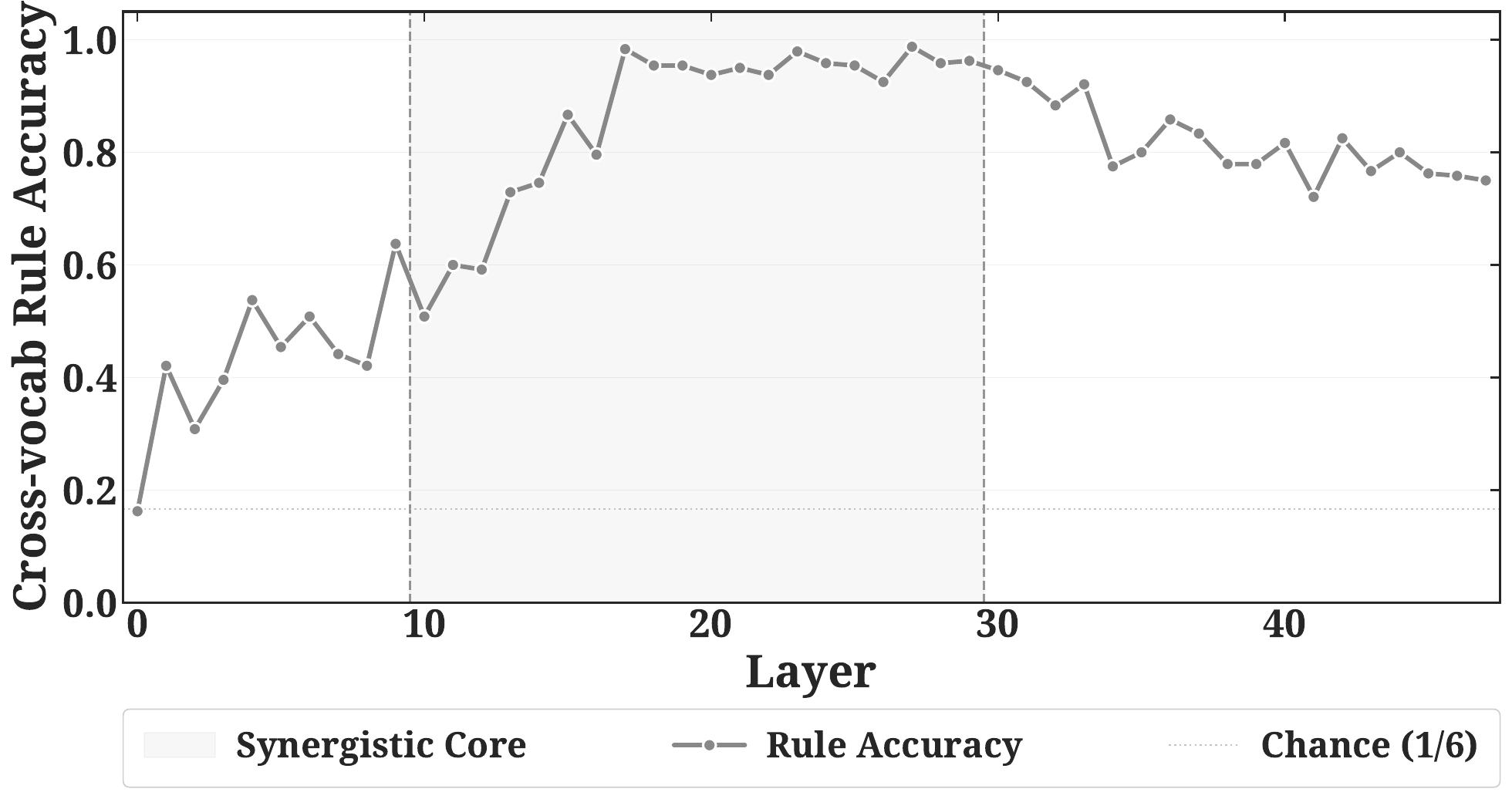}
    \caption{Rule-probe accuracy}
\end{subfigure}
\hfill
\begin{subfigure}[t]{0.48\textwidth}
    \centering
    \includegraphics[width=\linewidth]{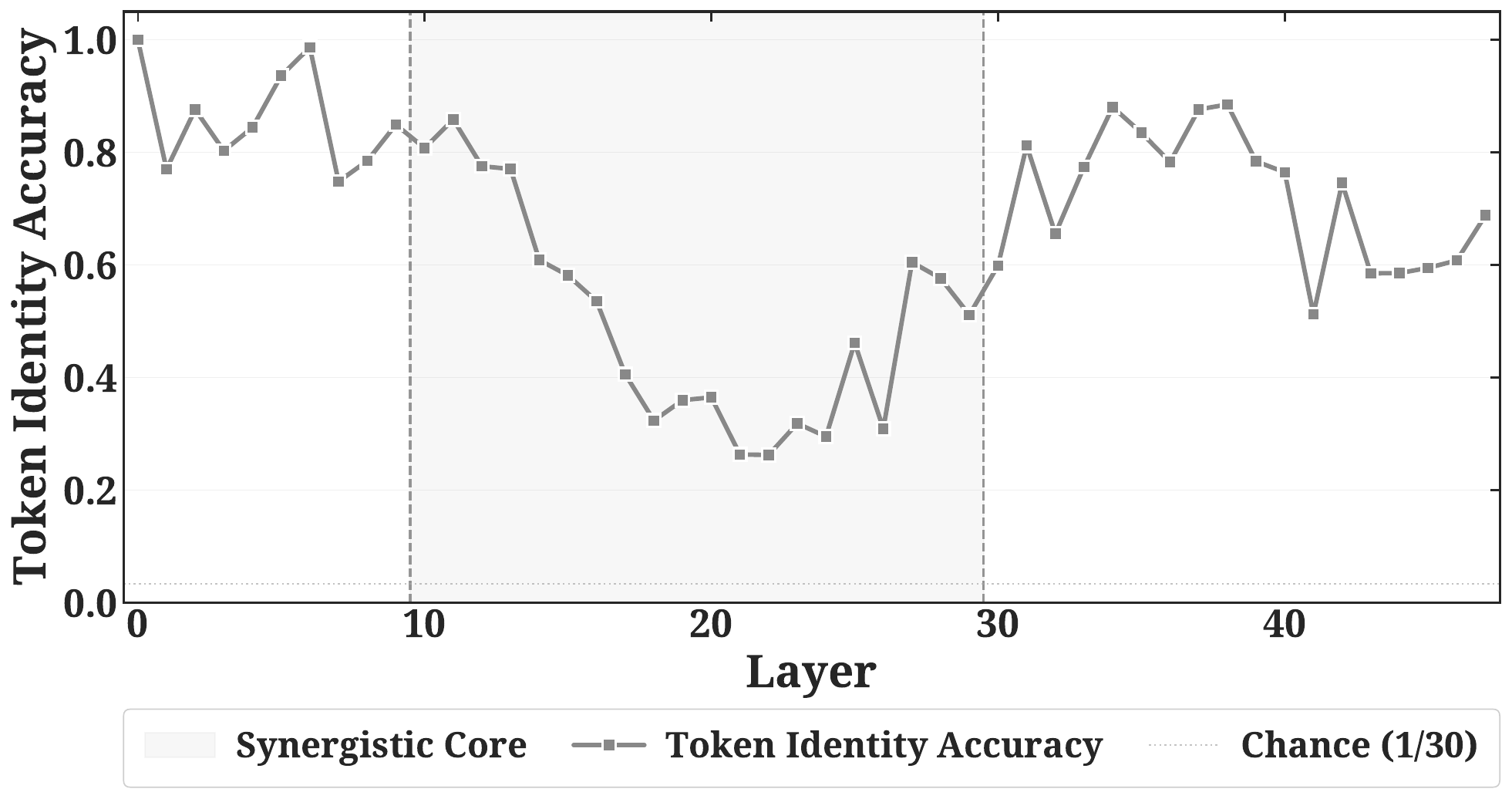}
    \caption{Token-probe accuracy}
\end{subfigure}

\caption{
Rule-level representation analysis in Gemma3-12B-Instruct.
}
\label{fig:app_gemma3_12b_instruct_rule_representations}
\end{figure}

\paragraph{Gemma3-12B-Instruct.}
Figure~\ref{fig:app_gemma3_12b_instruct_rule_representations} provides the representation analysis for Gemma3-12B-Instruct.

\noindent\textbf{Analysis.}
Gemma3-12B-Instruct shows a broader and more variable representation pattern than the Qwen models. This is consistent with the localization and interaction-profile analyses, where Gemma also exhibits more local variation across depth. The intrinsic-dimension curve should therefore be interpreted less as a single sharp minimum and more as evidence for an internal region where representations become more constrained than in the outermost layers.

The CKA and probe results are especially useful for interpreting this noisier profile. If Gemma's middle-stage organization were merely an artifact of activation magnitude or local layer fluctuations, we would not expect cross-vocabulary alignment and rule-probe accuracy to show compatible trends. The fact that these measurements still point to stronger rule-level structure in internal layers suggests that the model forms vocabulary-invariant relational representations, even if the exact layer-wise profile is less smooth.

The token-probe control also matters here. Gemma3-12B-Instruct is instruction-tuned, so its surface-form processing may differ from base models. A dissociation between rule recoverability and token-identity recoverability therefore provides stronger evidence that the internal representation is not simply preserving visible words. Overall, the Gemma result supports the same qualitative conclusion: abstraction-related information becomes more available in an internal computation stage rather than being uniformly encoded across all layers.

\paragraph{Summary.}
Across these additional models, the representation analyses support the same conclusion as the localization results. Intrinsic dimension, cross-vocabulary CKA, and rule probes do not produce identical layer curves in every architecture, but they consistently identify internal regions where representations become more compact, more vocabulary-invariant, and more informative about symbolic rules. The token-probe control further suggests that this effect is not reducible to surface-token identity. Together, these results indicate that the middle-stage computation identified by information-decomposition analysis corresponds to a genuine rule-level representation shift.

\subsection{Cross-Model Layer-Level Interventions}
\label{app:cross_model_layer_interventions}

The main text reports layer-level interventions on Qwen3-8B-Base. Here we repeat the same four intervention analyses on additional models. These interventions test whether the middle-stage components identified by the localization and representation analyses are functionally important for later computation.

For each model, we show four intervention maps. Full layer ablation skips a source layer throughout prompt processing and decoding. Prefill-only ablation skips a source layer only during the prompt stage and measures whether the effect propagates into later decoding. Circuit localization removes the contribution written by a source layer from the input of a later layer and recomputes the later layer. Future-position circuit localization applies the same analysis only to later generated positions. In all cases, stronger values indicate that later computation depends more strongly on the intervened source layer.

\begin{figure}[H]
\centering
\captionsetup{font=small,skip=3pt}
\captionsetup[subfigure]{font=small,skip=2pt}

\begin{subfigure}[t]{0.48\textwidth}
    \centering
    \includegraphics[width=\linewidth]{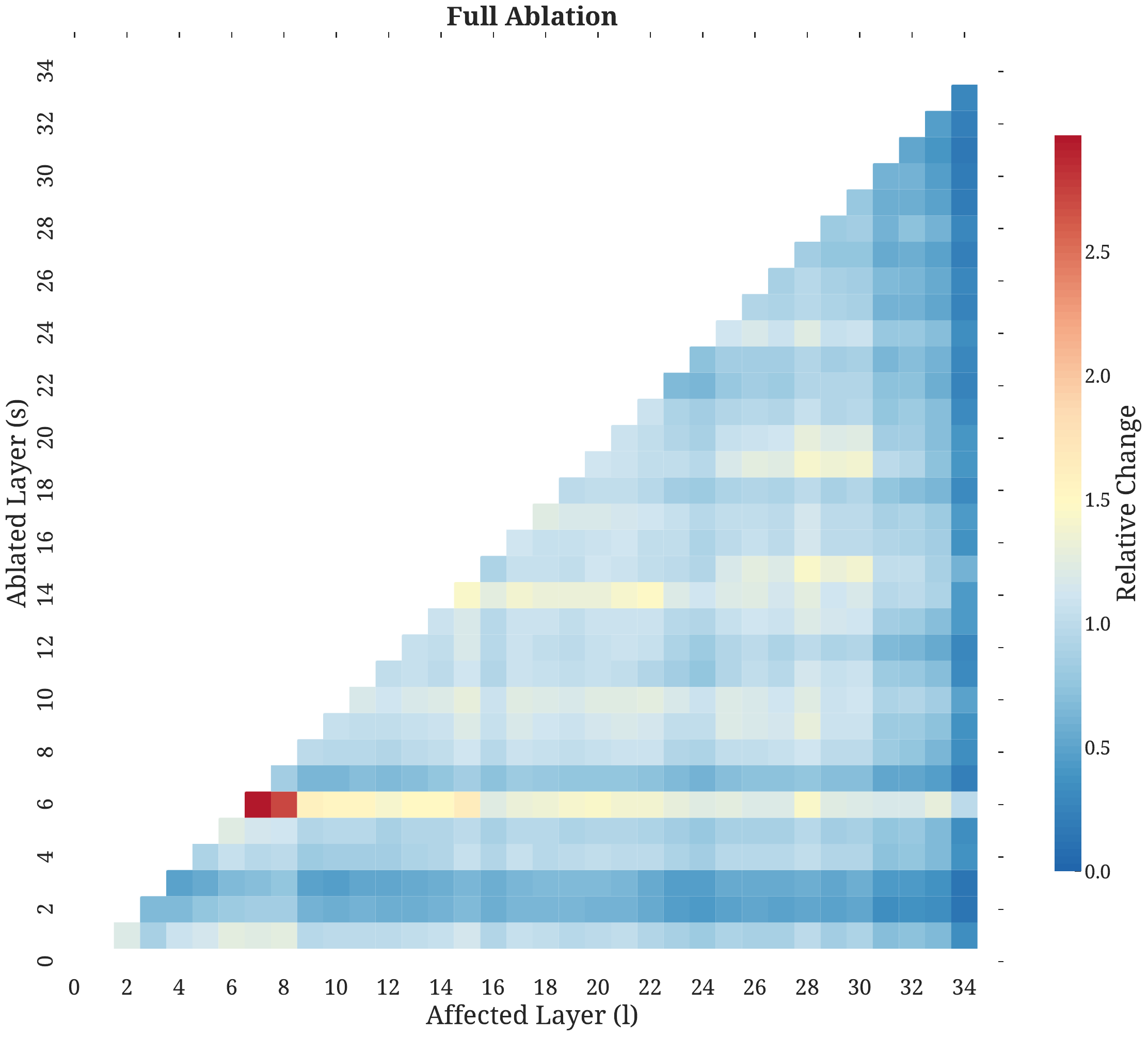}
    \caption{Full layer ablation}
\end{subfigure}
\hfill
\begin{subfigure}[t]{0.48\textwidth}
    \centering
    \includegraphics[width=\linewidth]{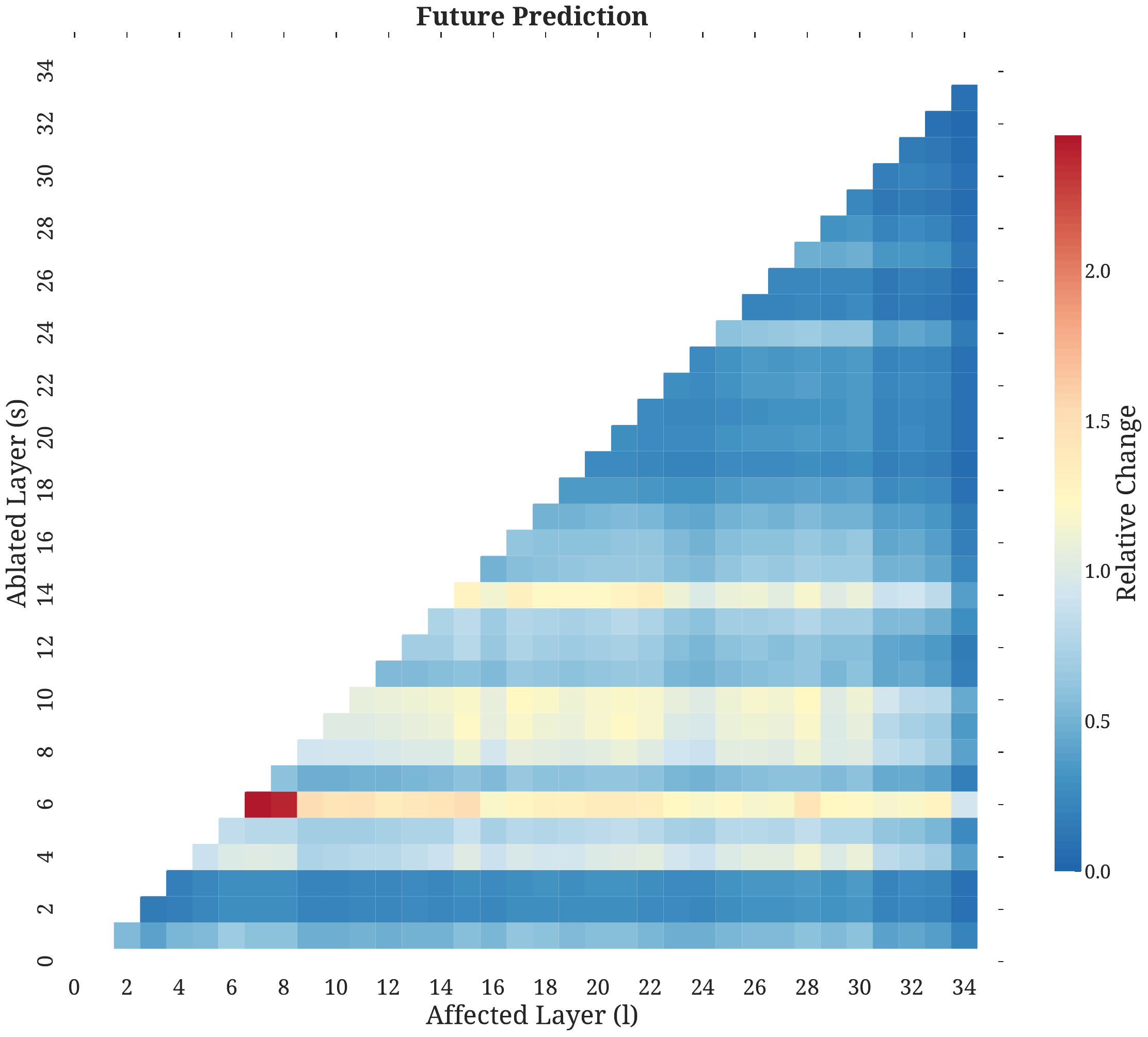}
    \caption{Prefill-only ablation}
\end{subfigure}

\begin{subfigure}[t]{0.48\textwidth}
    \centering
    \includegraphics[width=\linewidth]{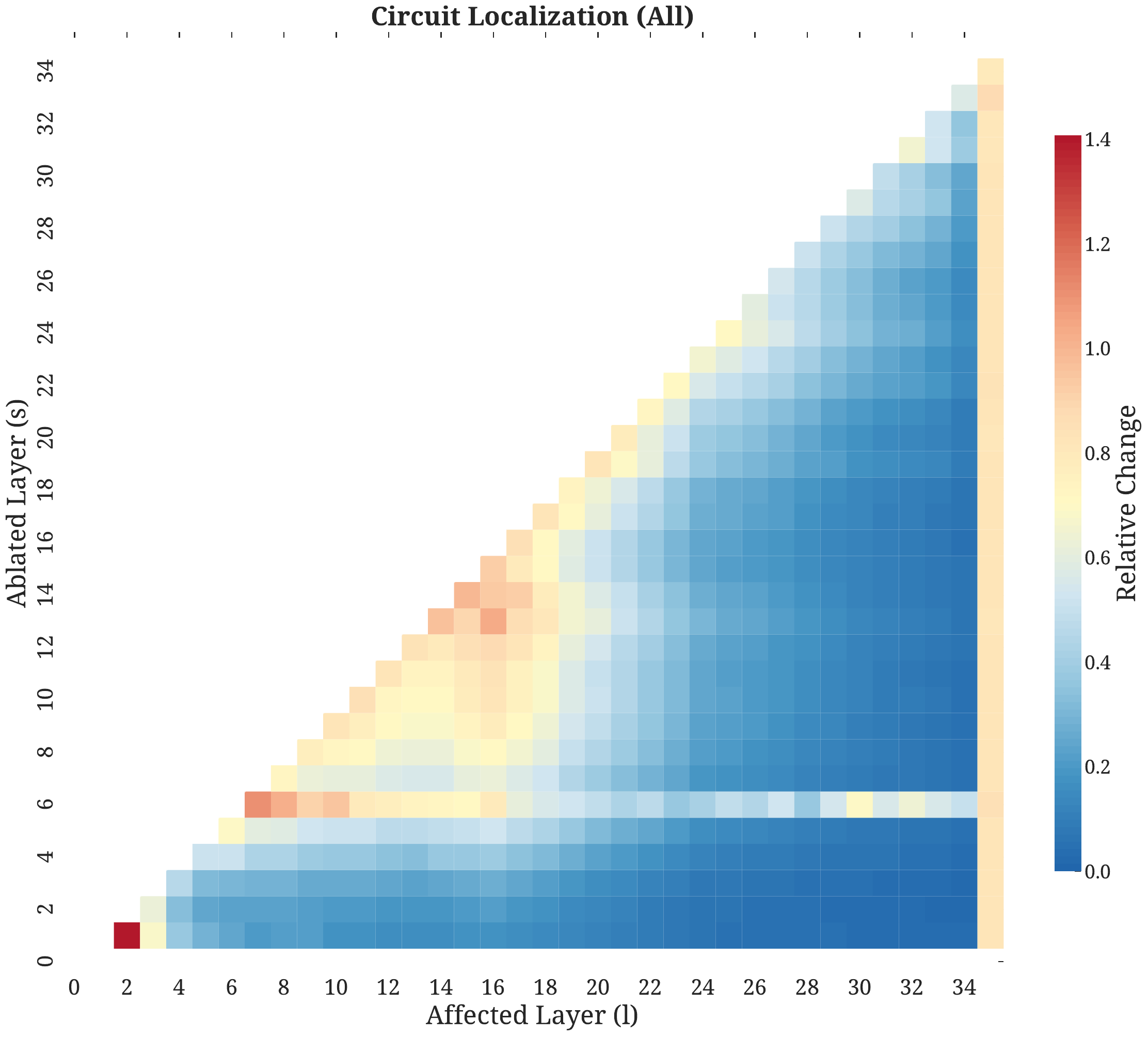}
    \caption{Circuit localization over all positions}
\end{subfigure}
\hfill
\begin{subfigure}[t]{0.48\textwidth}
    \centering
    \includegraphics[width=\linewidth]{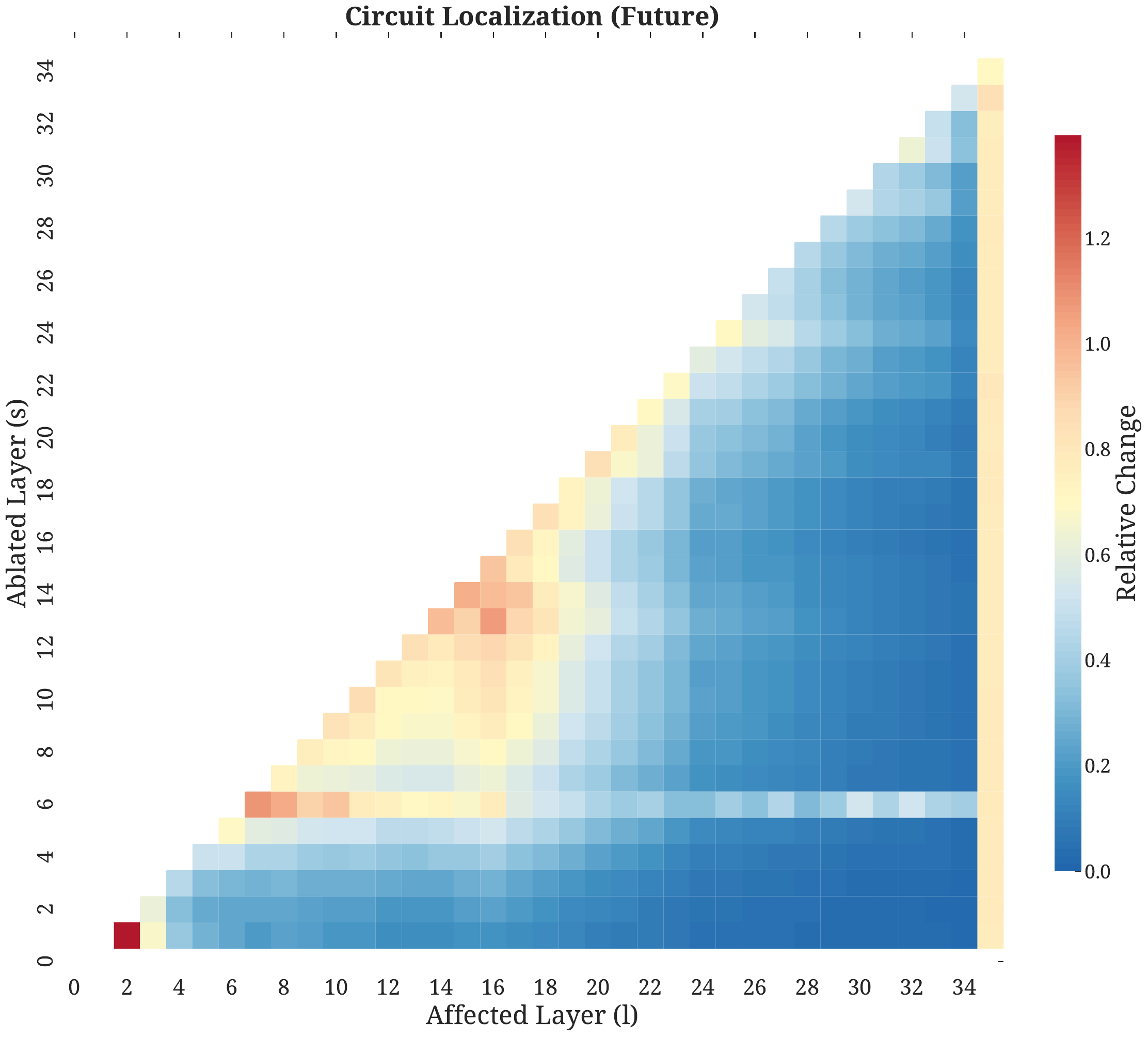}
    \caption{Circuit localization over future positions}
\end{subfigure}

\caption{
Layer-level intervention results in Qwen3-4B-Base.
}
\label{fig:app_qwen3_4b_layer_interventions}
\end{figure}

\paragraph{Qwen3-4B-Base.}
Figure~\ref{fig:app_qwen3_4b_layer_interventions} shows the layer-level intervention results for Qwen3-4B-Base.

\noindent\textbf{Analysis.}
Qwen3-4B-Base shows a compact but visible intervention-sensitive region. Because this model has fewer layers, the source layers that produce large downstream disturbance appear over a relatively narrow depth interval. This is consistent with the localization results: the model must allocate representation transformation to a smaller number of internal layers, so perturbing these layers produces a concentrated downstream effect.

The full ablation map measures the direct effect of removing each source layer during the entire generation process. Larger disturbances in the intermediate region suggest that later layers depend more strongly on computations written by these layers than on computations written by the outermost layers. The prefill-only setting is particularly informative: if skipping a layer only during prompt processing still changes later decoding computations, then the layer is not merely affecting the current token locally; it writes information that persists and is reused during generation.

The two circuit-localization maps provide a more targeted view. Instead of removing a layer entirely, they subtract the source-layer contribution from the input of a later layer. This tests whether later layers actually use the information written by the source layer. In Qwen3-4B-Base, the stronger intermediate effects indicate that the internal computation stage is not only correlated with representation transformation but also functionally upstream of later prediction-related computation.

\begin{figure}[H]
\centering
\captionsetup{font=small,skip=3pt}
\captionsetup[subfigure]{font=small,skip=2pt}

\begin{subfigure}[t]{0.48\textwidth}
    \centering
    \includegraphics[width=\linewidth]{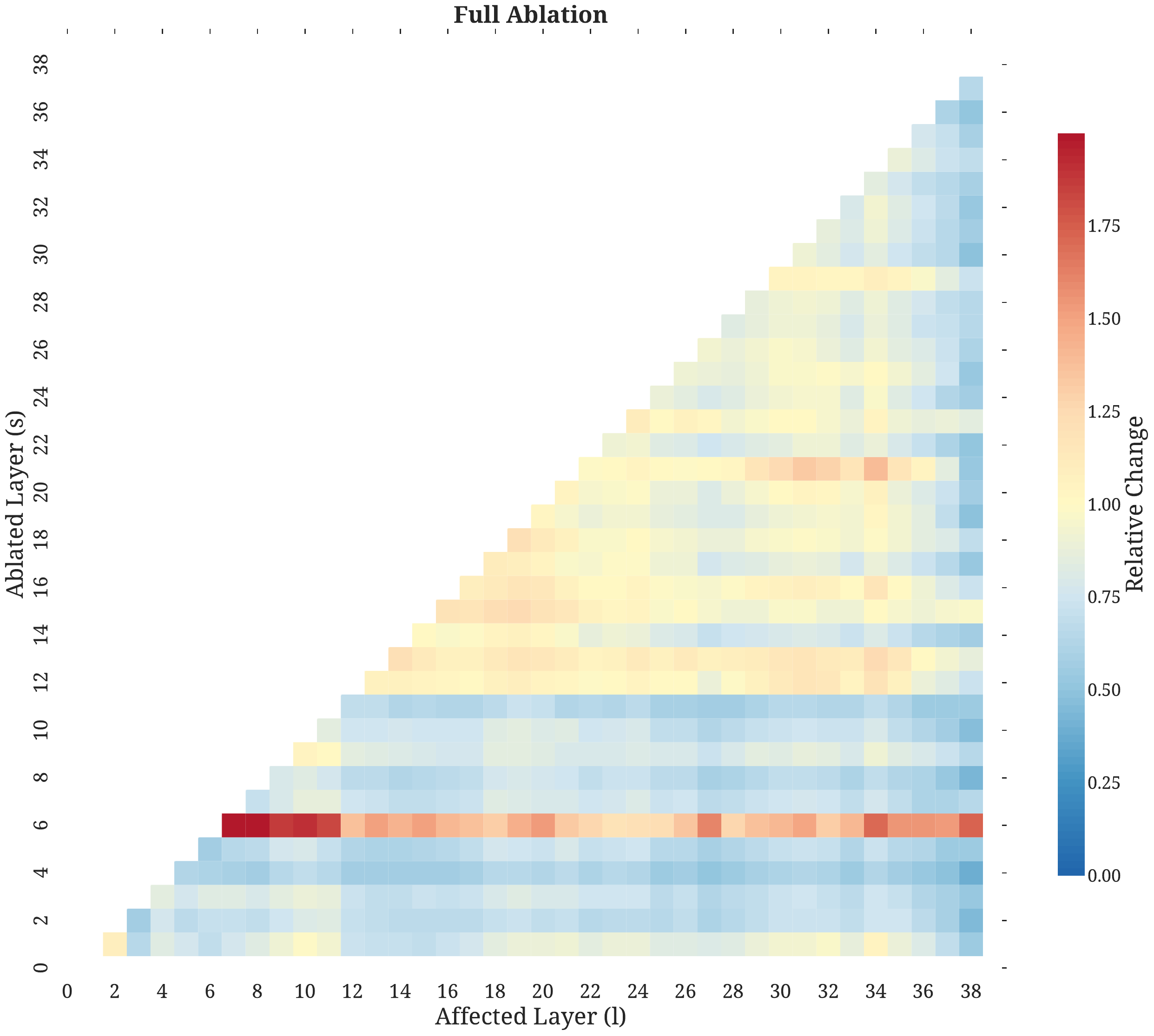}
    \caption{Full layer ablation}
\end{subfigure}
\hfill
\begin{subfigure}[t]{0.48\textwidth}
    \centering
    \includegraphics[width=\linewidth]{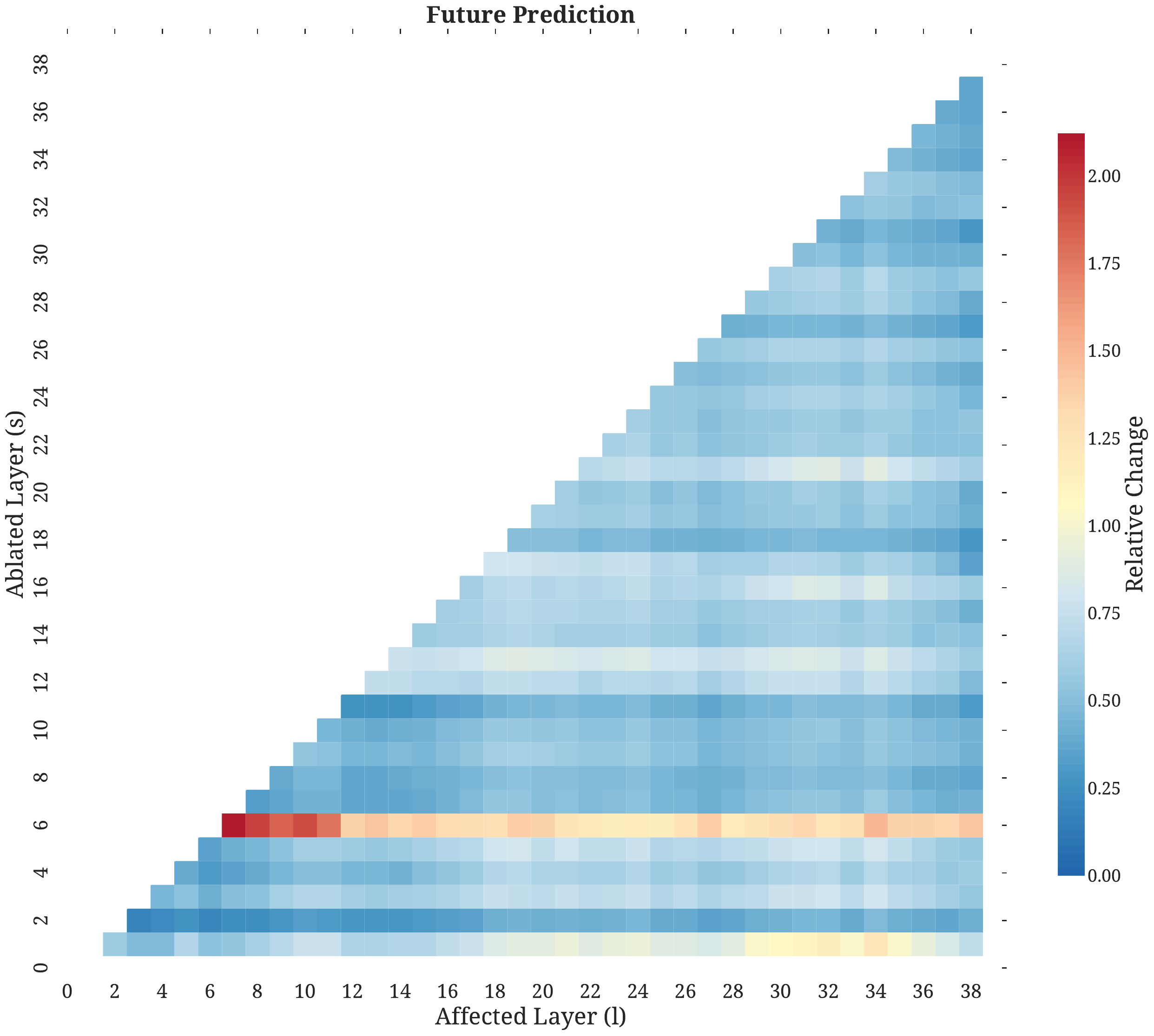}
    \caption{Prefill-only ablation}
\end{subfigure}

\begin{subfigure}[t]{0.48\textwidth}
    \centering
    \includegraphics[width=\linewidth]{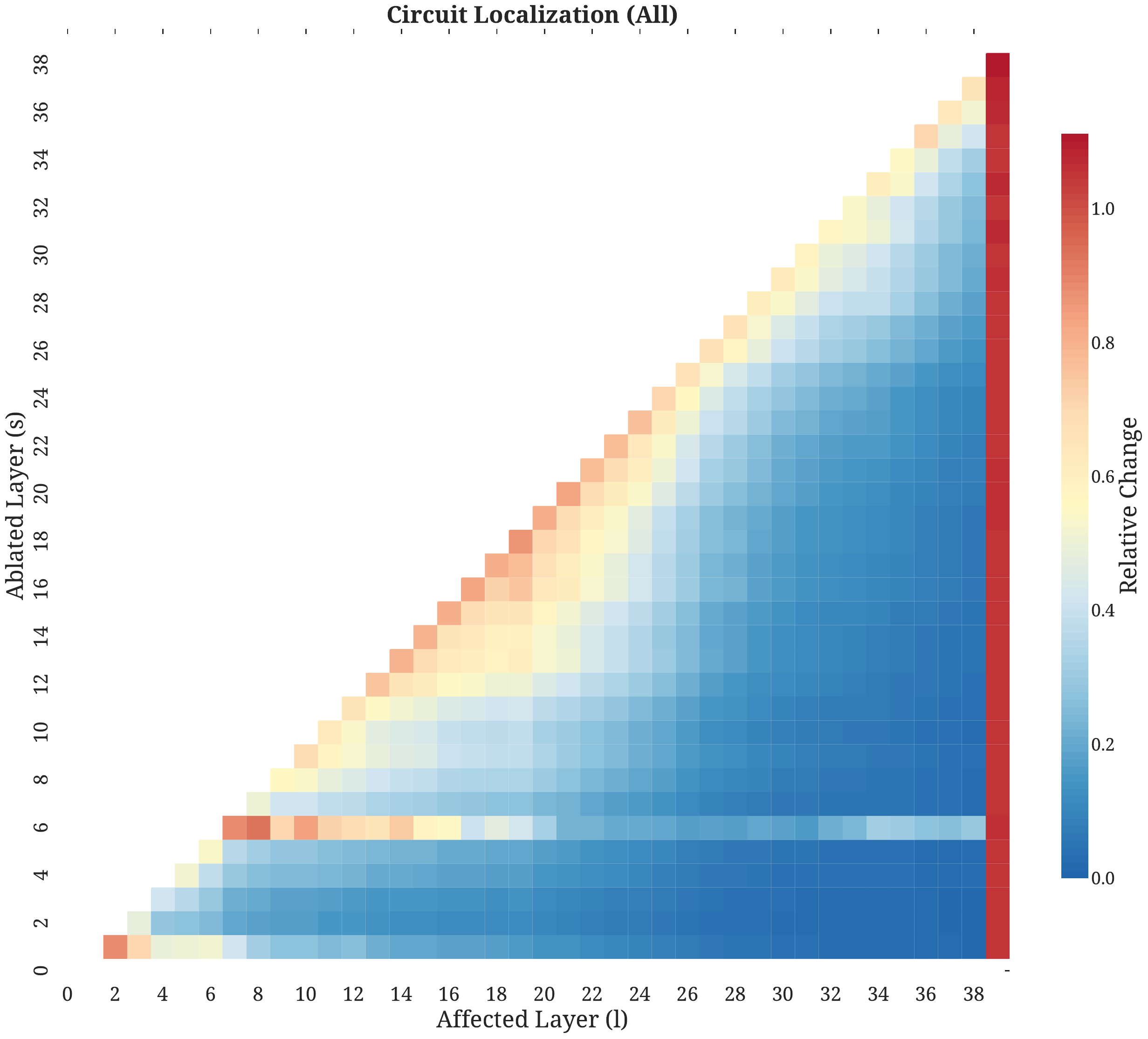}
    \caption{Circuit localization over all positions}
\end{subfigure}
\hfill
\begin{subfigure}[t]{0.48\textwidth}
    \centering
    \includegraphics[width=\linewidth]{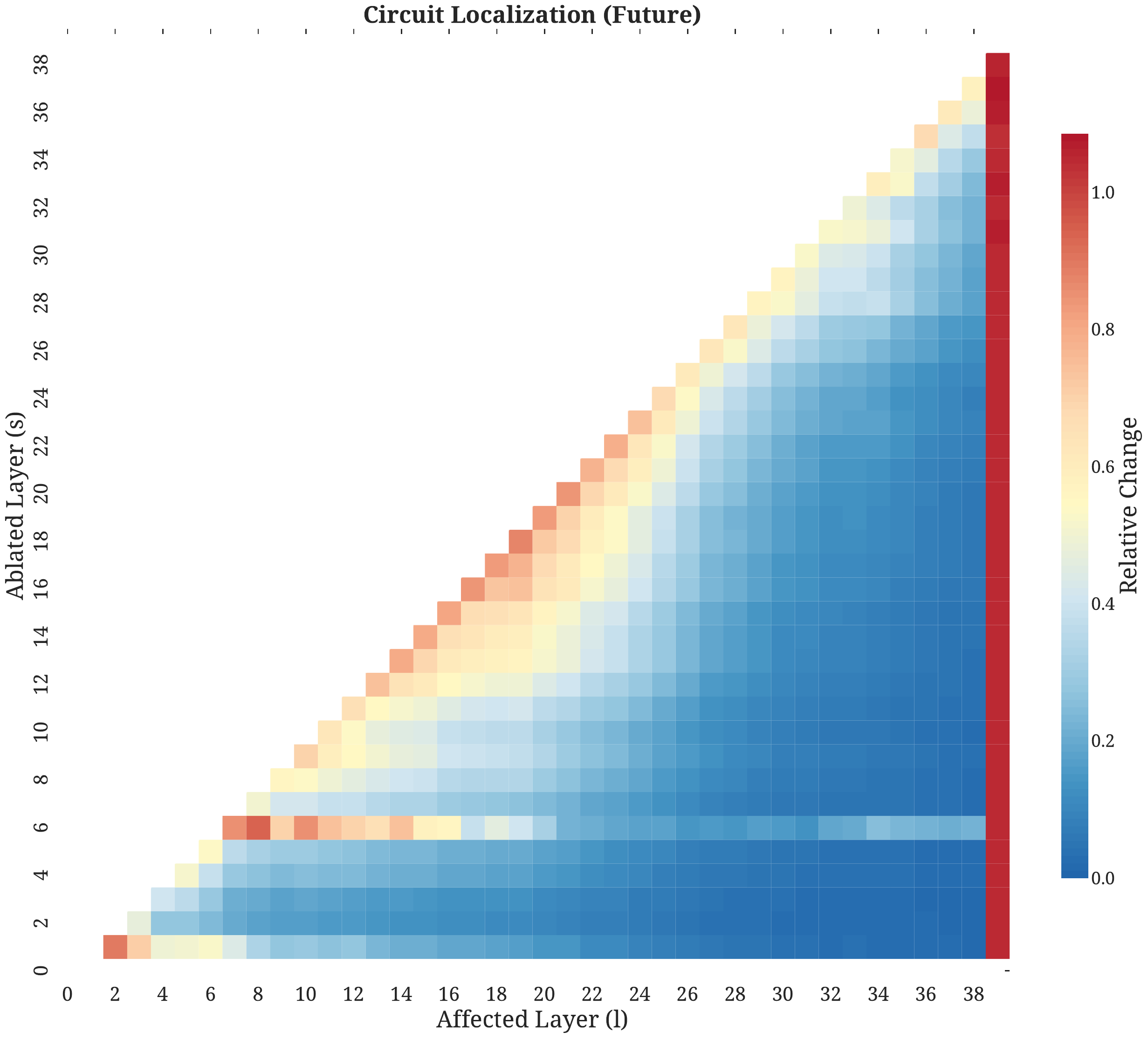}
    \caption{Circuit localization over future positions}
\end{subfigure}

\caption{
Layer-level intervention results in Qwen3-14B-Base.
}
\label{fig:app_qwen3_14b_layer_interventions}
\end{figure}

\paragraph{Qwen3-14B-Base.}
Figure~\ref{fig:app_qwen3_14b_layer_interventions} reports the same intervention analyses for Qwen3-14B-Base.

\noindent\textbf{Analysis.}
Qwen3-14B-Base shows a broader intervention-sensitive region than Qwen3-4B-Base. This is expected because the deeper model can distribute its internal transformation stage across more layers. The relevant phenomenon is therefore not a single maximally sensitive layer, but a contiguous or semi-contiguous internal band whose removal or perturbation affects later computation more strongly than perturbing outer layers.

The full ablation and prefill-only maps together suggest that the intermediate layers influence both immediate residual updates and later decoding behavior. Full ablation captures direct downstream dependence, while prefill-only ablation tests whether information written during prompt processing continues to matter during generation. The fact that both settings highlight internal layers supports the interpretation that middle-stage computation is not merely transient.

The circuit-localization maps further refine this claim. By subtracting the source-layer contribution from later-layer inputs, these experiments test a more local dependency relation between source and affected layers. In Qwen3-14B-Base, the affected region is broader than in smaller models, suggesting that larger models may implement the same reasoning-related transformation through a wider distributed pathway. Nevertheless, the strongest dependence remains away from the outermost layers, consistent with the main layer-wise organization.

\begin{figure}[H]
\centering
\captionsetup{font=small,skip=3pt}
\captionsetup[subfigure]{font=small,skip=2pt}

\begin{subfigure}[t]{0.48\textwidth}
    \centering
    \includegraphics[width=\linewidth]{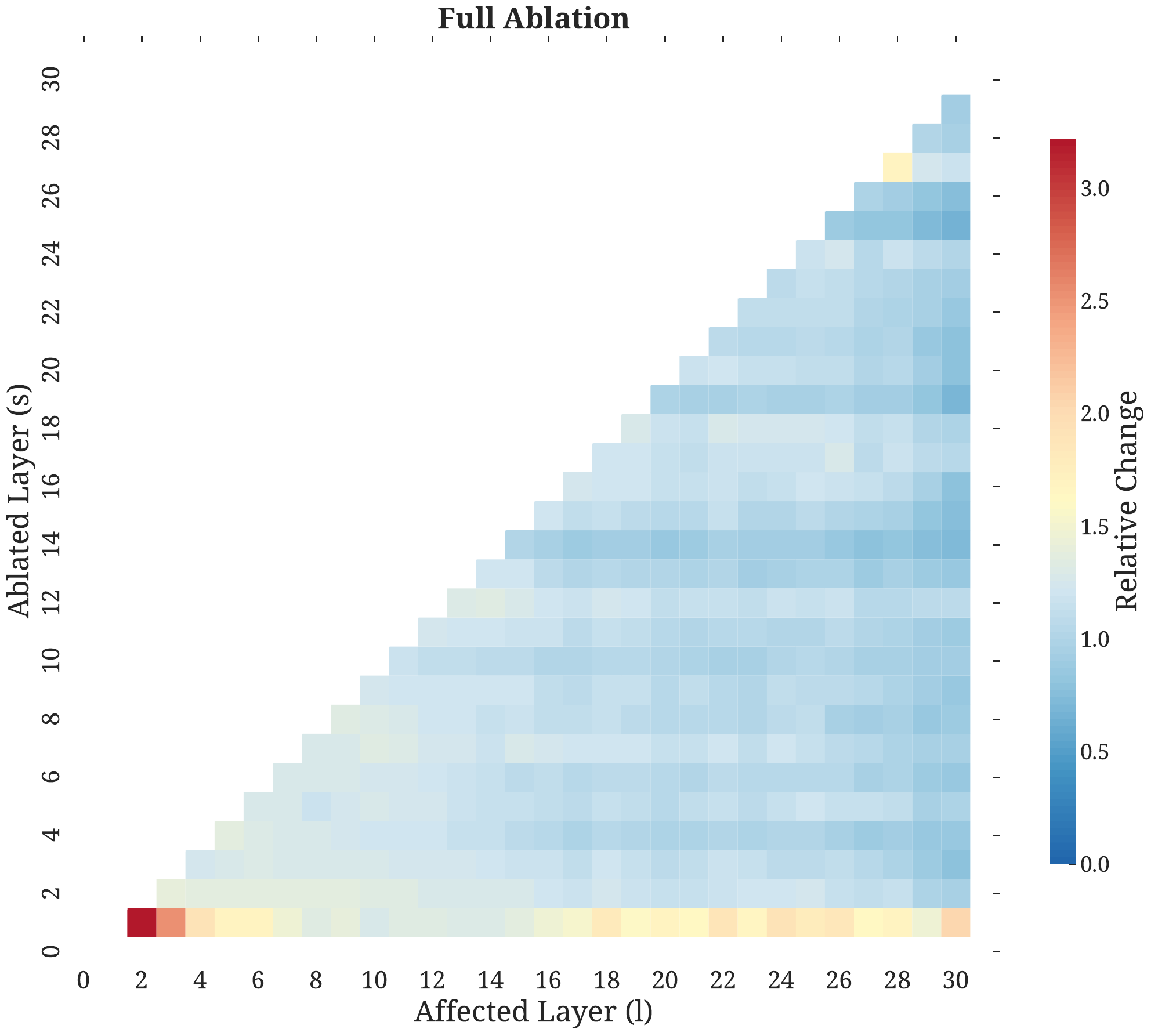}
    \caption{Full layer ablation}
\end{subfigure}
\hfill
\begin{subfigure}[t]{0.48\textwidth}
    \centering
    \includegraphics[width=\linewidth]{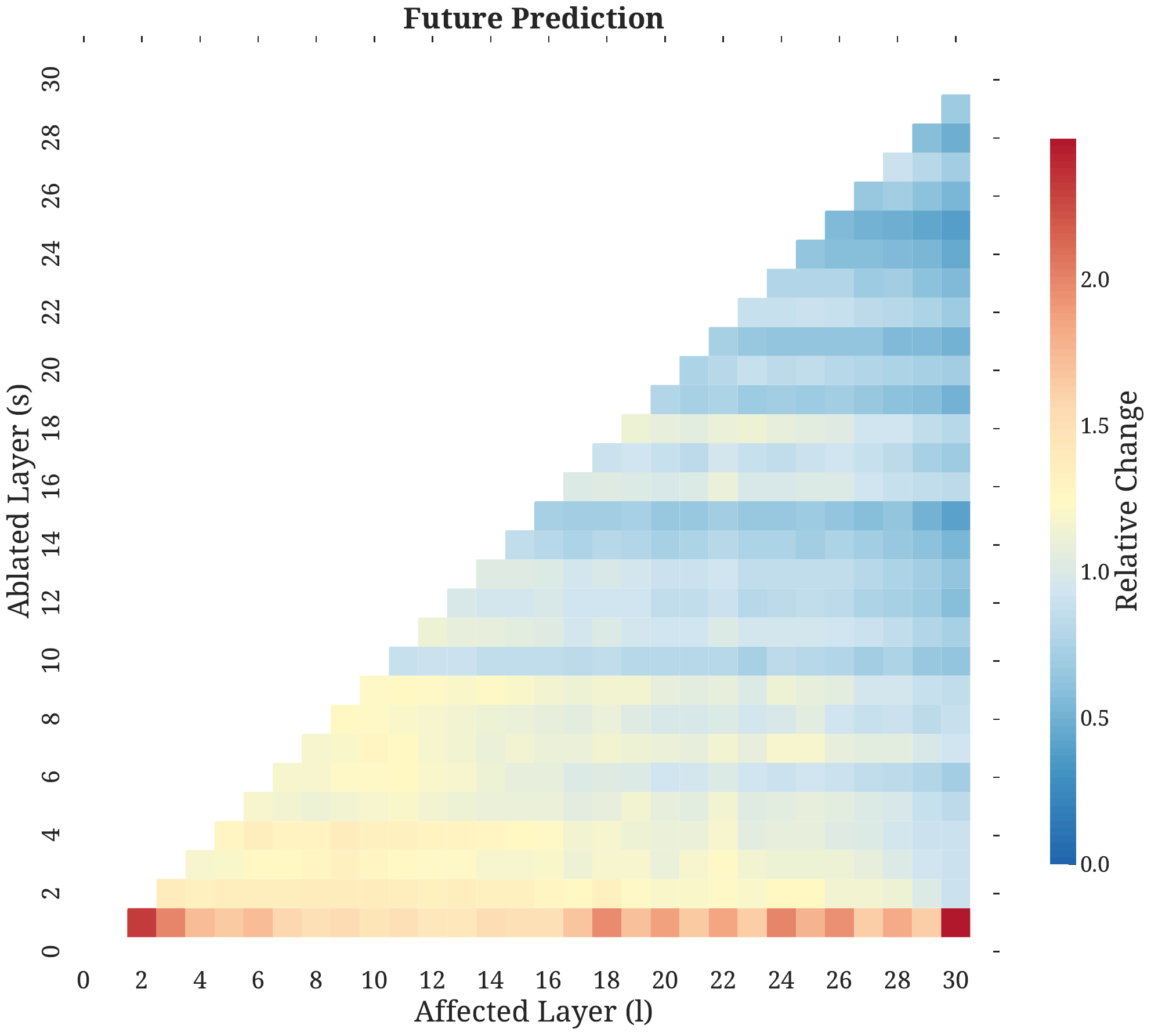}
    \caption{Prefill-only ablation}
\end{subfigure}

\begin{subfigure}[t]{0.48\textwidth}
    \centering
    \includegraphics[width=\linewidth]{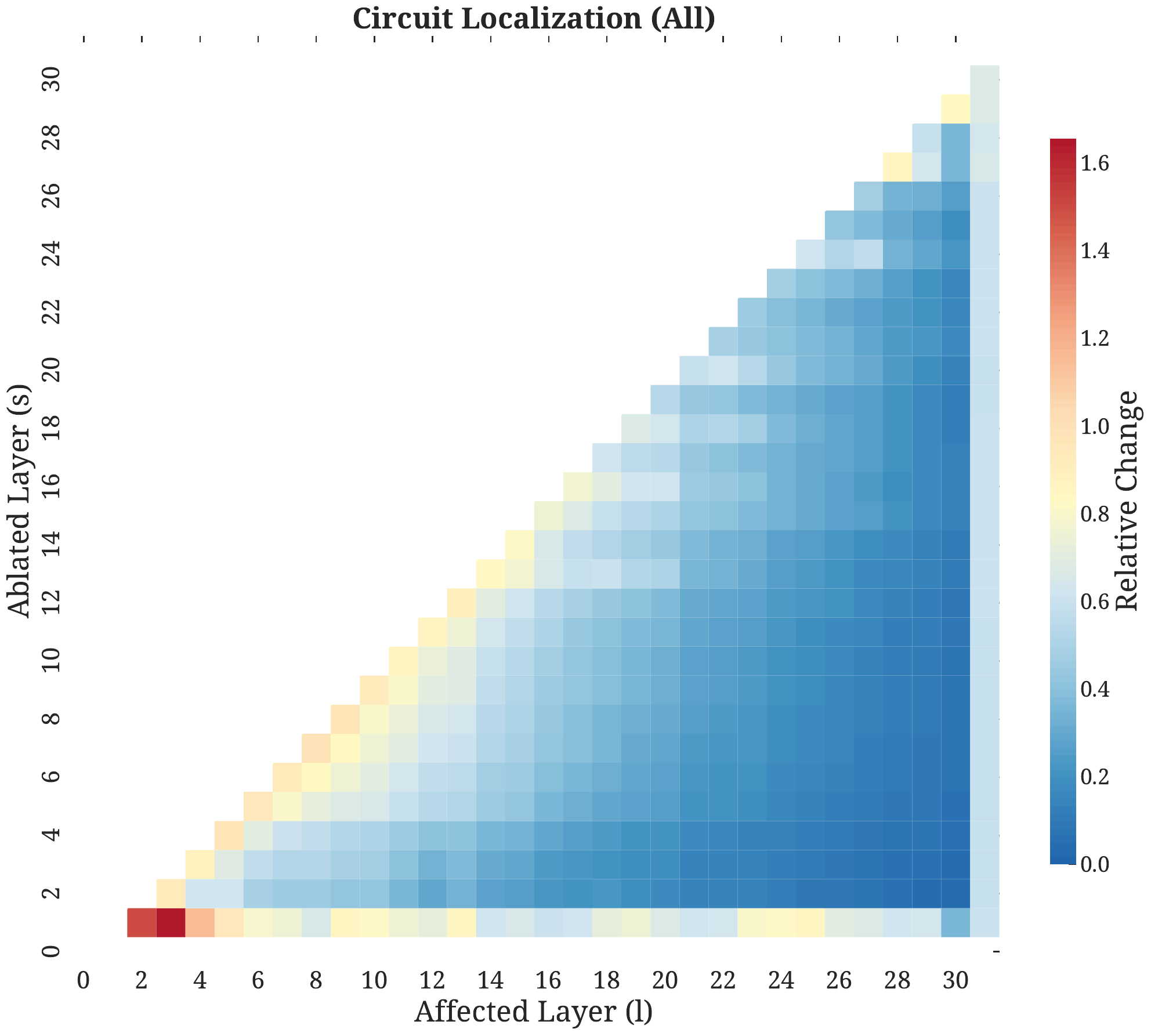}
    \caption{Circuit localization over all positions}
\end{subfigure}
\hfill
\begin{subfigure}[t]{0.48\textwidth}
    \centering
    \includegraphics[width=\linewidth]{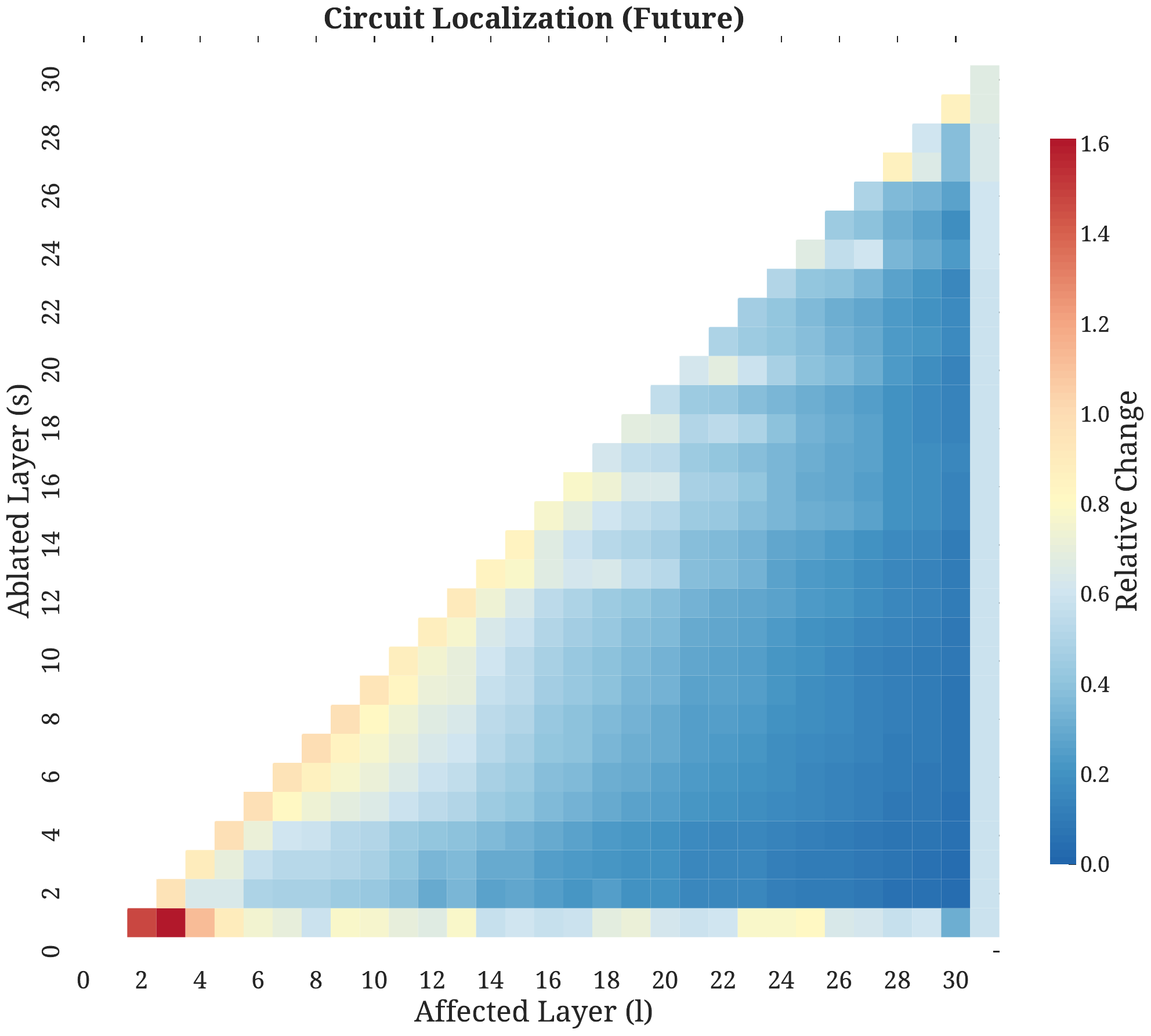}
    \caption{Circuit localization over future positions}
\end{subfigure}

\caption{
Layer-level intervention results in Llama3.1-8B-Base.
}
\label{fig:app_llama31_8b_layer_interventions}
\end{figure}

\paragraph{Llama3.1-8B-Base.}
Figure~\ref{fig:app_llama31_8b_layer_interventions} shows the intervention results for Llama3.1-8B-Base.

\noindent\textbf{Analysis.}
Llama3.1-8B-Base provides an important cross-family test because its localization and cosine profiles are more oscillatory than those of Qwen models. The intervention maps therefore help determine whether the internal transformation pattern is merely geometric or whether it has functional consequences for later computation.

The results support the latter interpretation. Intermediate layers produce stronger downstream disturbances than the outermost regions, indicating that later layers rely on information written in the internal computation stage. This is particularly meaningful for Llama3.1-8B-Base because its cosine profile contains a pronounced negative middle region. The intervention results suggest that this directional reorganization is not a superficial geometric artifact; perturbing the corresponding layers changes subsequent residual computations.

The circuit-localization maps are especially useful here. They show whether later layers depend on the source-layer contribution after the normal generated sequence has already been obtained. Stronger effects in internal regions indicate that Llama's later layers reuse information written by intermediate layers. Thus, even though Llama implements the layer-wise organization with a different local profile from Qwen, the causal structure remains similar: internal transformations have stronger downstream influence than outer-layer preservation or routing.

\begin{figure}[H]
\centering
\captionsetup{font=small,skip=3pt}
\captionsetup[subfigure]{font=small,skip=2pt}

\begin{subfigure}[t]{0.48\textwidth}
    \centering
    \includegraphics[width=\linewidth]{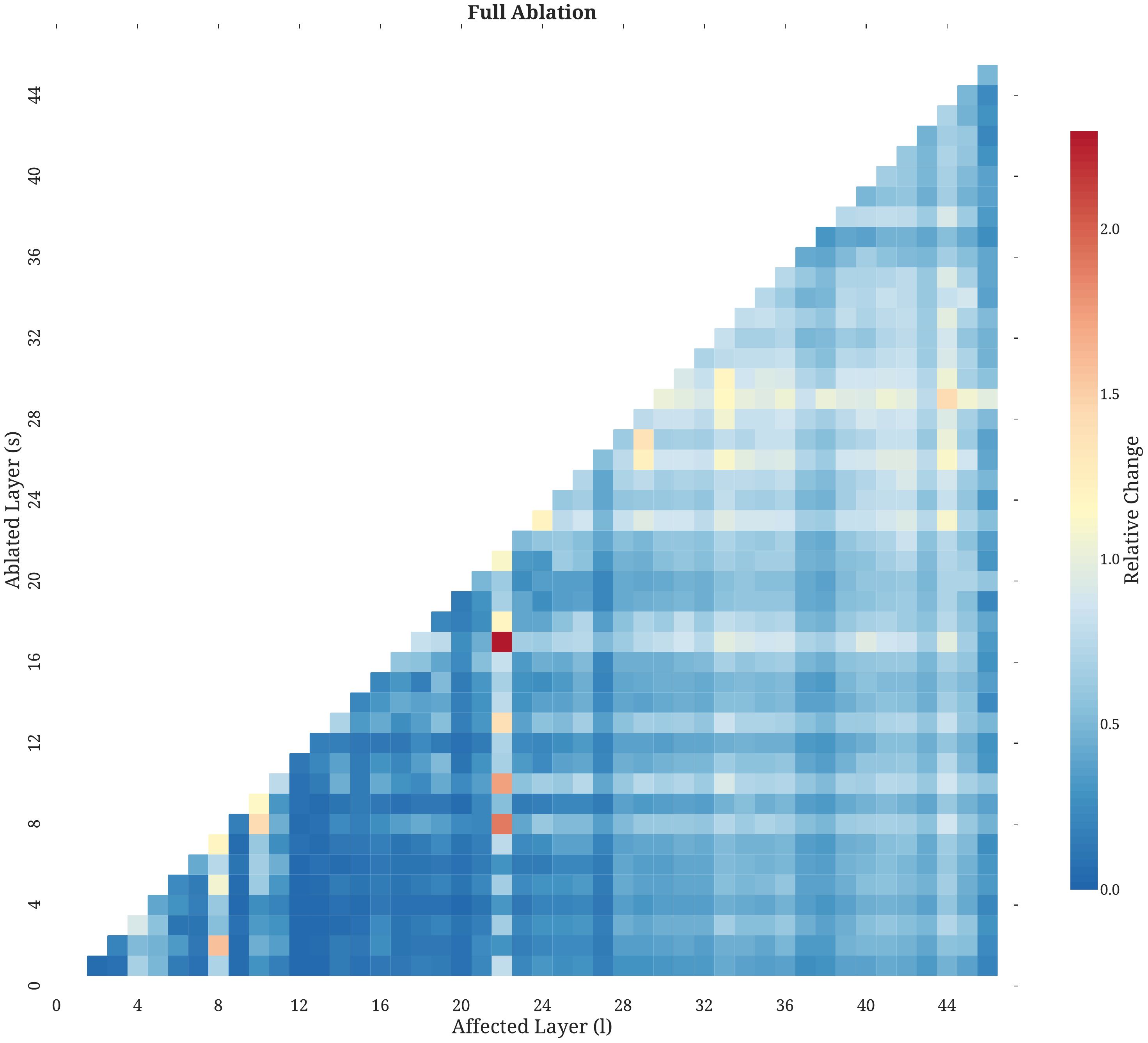}
    \caption{Full layer ablation}
\end{subfigure}
\hfill
\begin{subfigure}[t]{0.48\textwidth}
    \centering
    \includegraphics[width=\linewidth]{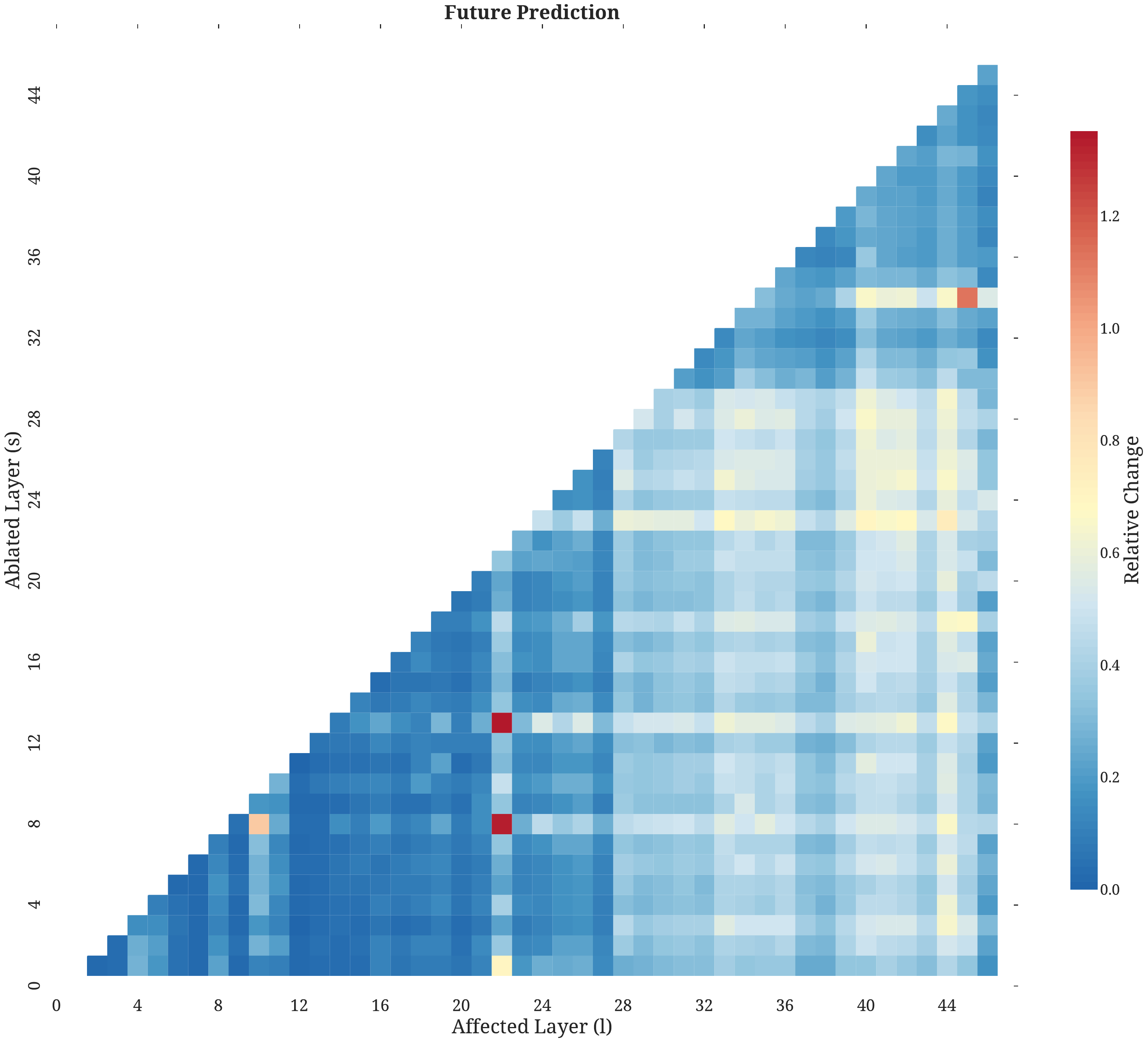}
    \caption{Prefill-only ablation}
\end{subfigure}

\begin{subfigure}[t]{0.48\textwidth}
    \centering
    \includegraphics[width=\linewidth]{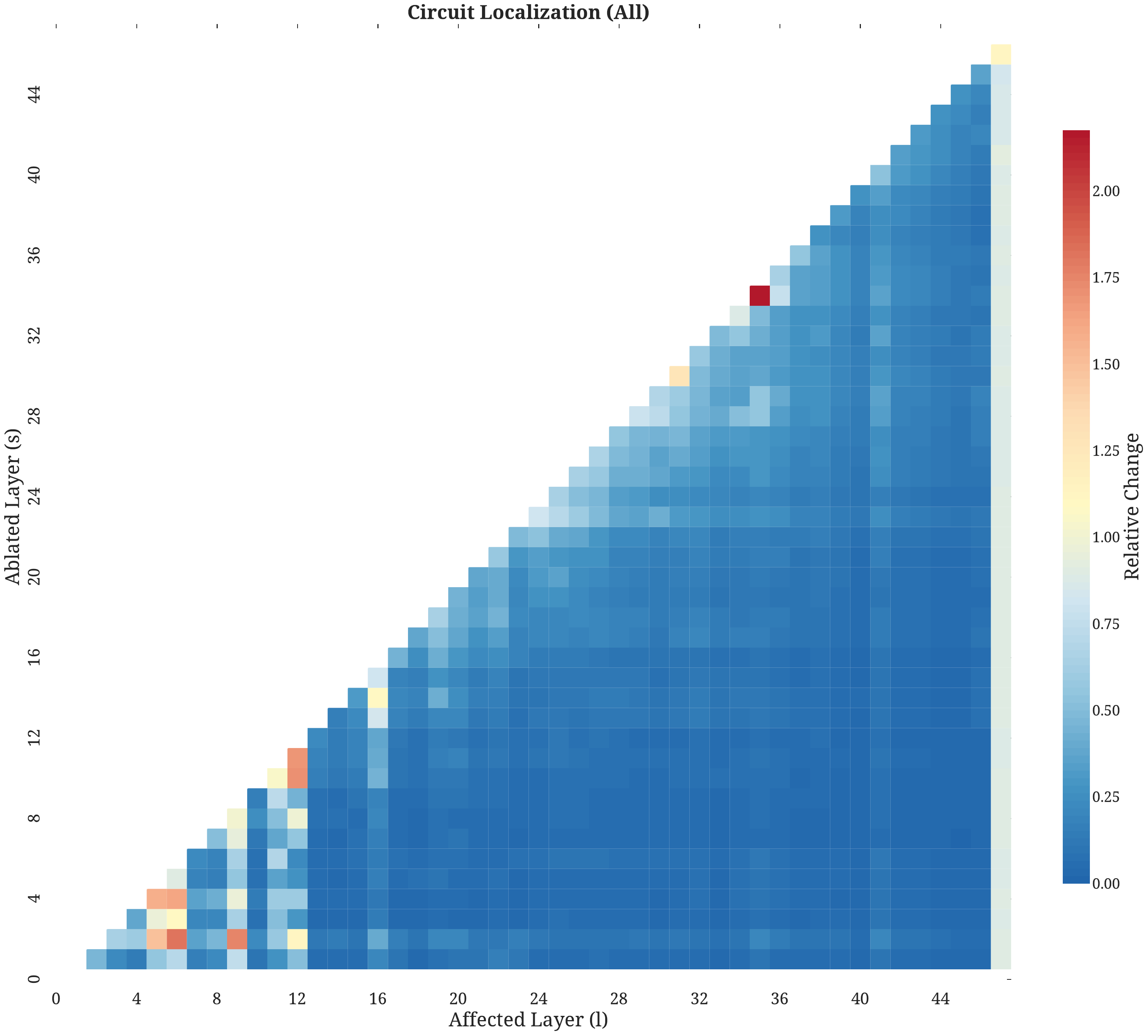}
    \caption{Circuit localization over all positions}
\end{subfigure}
\hfill
\begin{subfigure}[t]{0.48\textwidth}
    \centering
    \includegraphics[width=\linewidth]{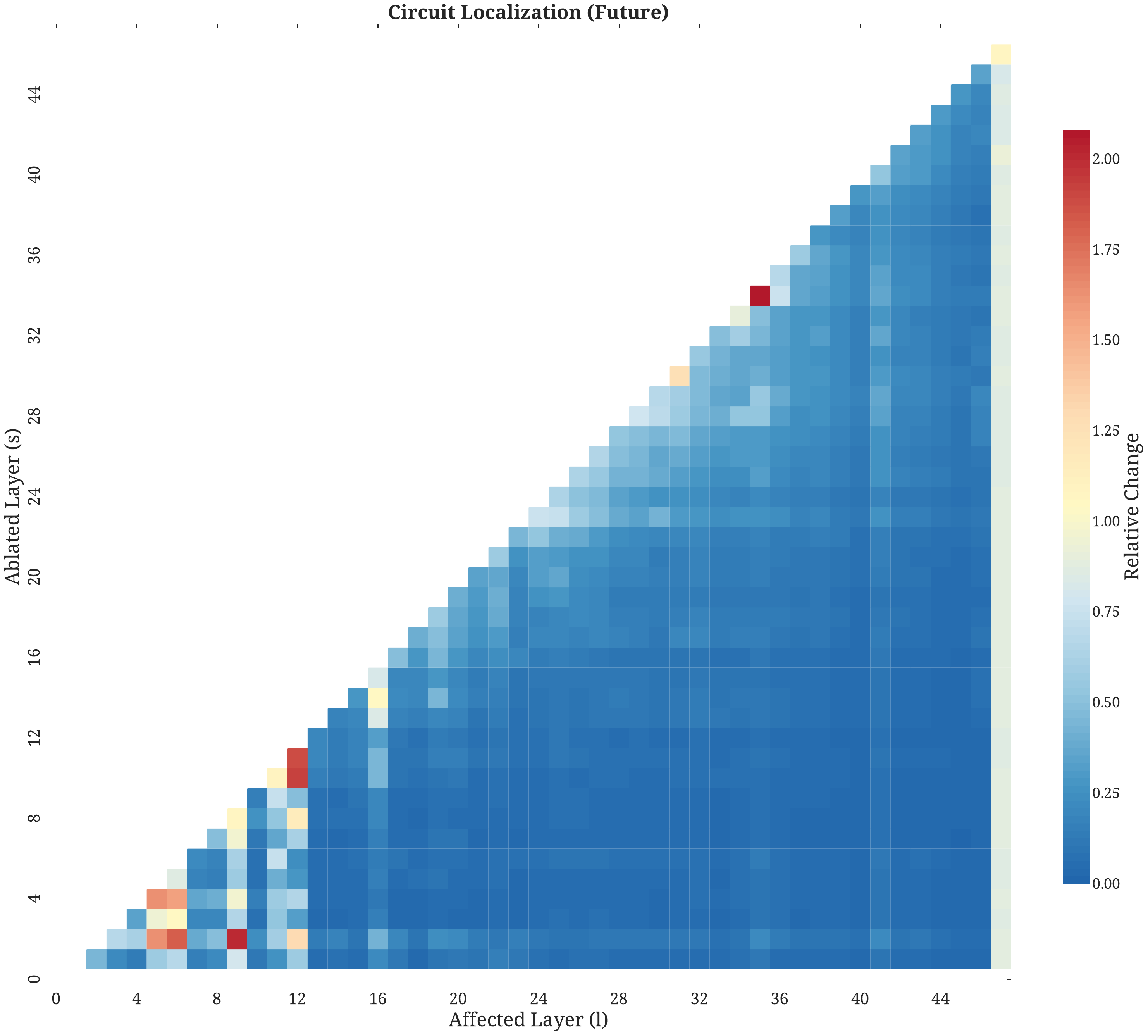}
    \caption{Circuit localization over future positions}
\end{subfigure}

\caption{
Layer-level intervention results in Gemma3-12B-Instruct.
}
\label{fig:app_gemma3_12b_instruct_layer_interventions}
\end{figure}

\paragraph{Gemma3-12B-Instruct.}
Figure~\ref{fig:app_gemma3_12b_instruct_layer_interventions} provides the layer-level intervention results for Gemma3-12B-Instruct.

\noindent\textbf{Analysis.}
Gemma3-12B-Instruct shows a broader and more locally variable intervention pattern. This is consistent with the earlier localization and representation analyses, where Gemma also exhibited a less sharply localized internal profile. The intervention results suggest that this variability does not eliminate the middle-stage organization; instead, it indicates that Gemma may distribute downstream-relevant transformations across several internal regions.

The distinction between full ablation and prefill-only ablation is important for this model. Full ablation measures how much the entire generation computation depends on a source layer, while prefill-only ablation tests whether prompt-stage information written by that layer continues to affect later decoding. If internal layers remain influential under prefill-only ablation, this indicates that they contribute information that persists across time rather than merely modifying the current hidden state.

The circuit-localization maps provide the strongest evidence of downstream dependence. By subtracting the source contribution from later-layer inputs, they ask whether later layers use that contribution. In Gemma3-12B-Instruct, stronger internal-layer effects indicate that the model's later computations depend on information written by the interior computation stage. This complements the main-text head-level ablation result on the same model: both layer-level and head-level interventions suggest that composition-oriented internal components have stronger functional influence than preservation-oriented outer components.

\paragraph{Summary.}
Across the four additional models, layer-level interventions support the same conclusion as the localization and representation analyses. The exact disturbance pattern varies across architectures and depths, but the strongest downstream effects tend to arise from internal source layers rather than from the outermost regions. Full layer ablation shows that removing these layers disrupts later residual updates; prefill-only ablation shows that their influence can propagate from prompt processing into later decoding; and circuit localization shows that later layers directly depend on information written by internal source layers.

These results strengthen the interpretation that the middle computation stage is functionally important, not merely statistically distinctive. The internal layers identified by composition-oriented localization and rule-level representation analyses also exert stronger causal influence on downstream computation.

\subsection{Cross-Model Head-Level Ablation}
\label{app:cross_model_head_ablation}

The main text reports the head-level ablation result on Gemma3-12B-Instruct. Here we provide the corresponding cumulative head-ablation results for additional models. Heads are ranked by the composition--preservation orientation score defined in Appendix~\ref{app:info_decomposition}. We compare three ablation trajectories: removing composition-oriented heads first, removing preservation-oriented heads first, and removing randomly selected heads. Faster accuracy degradation under composition-oriented ablation indicates that the heads identified by our interaction-based criterion have stronger functional influence on reasoning performance.

\begin{figure}[H]
\centering
\captionsetup{font=small,skip=3pt}

\includegraphics[width=0.82\textwidth]{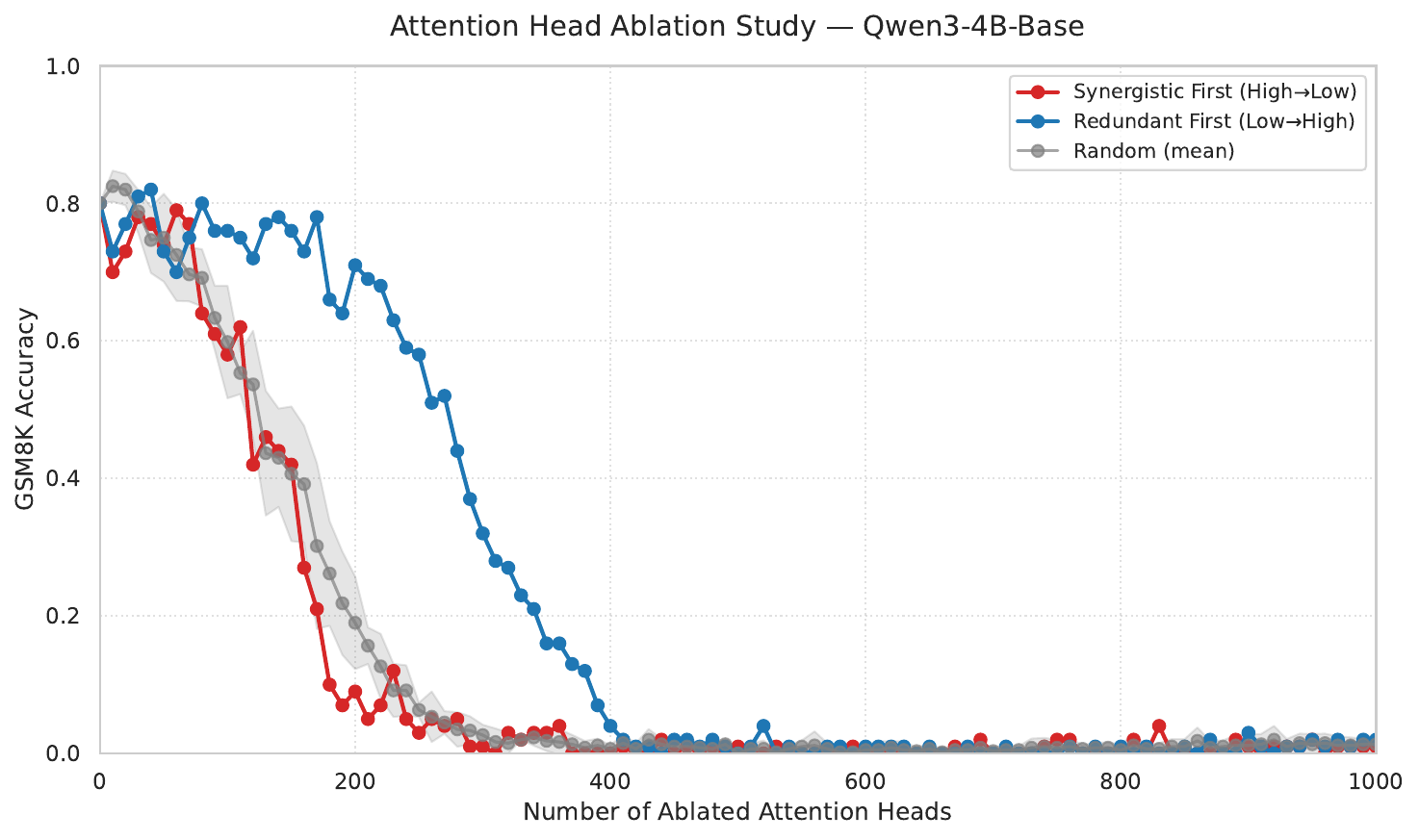}

\caption{
Head-level cumulative ablation in Qwen3-4B-Base.
}
\label{fig:app_qwen3_4b_head_ablation}
\end{figure}

\paragraph{Qwen3-4B-Base.}
Figure~\ref{fig:app_qwen3_4b_head_ablation} shows the cumulative head-ablation result for Qwen3-4B-Base.

\noindent\textbf{Analysis.}
Qwen3-4B-Base provides a useful small-scale test of the head-level intervention result. Because the model has fewer layers and fewer total attention heads than larger models, one might expect head removal to produce noisier curves. Nevertheless, the qualitative ordering remains informative: when composition-oriented heads are removed first, accuracy drops faster than under random ablation or preservation-oriented ablation.

This pattern suggests that the composition--preservation orientation score does not merely identify heads with unusual activation statistics. Instead, the ranking captures components that are more functionally involved in reasoning. In a smaller model, the redundancy available to compensate for head removal is more limited, so targeted removal of composition-oriented heads can produce a sharper degradation.

The preservation-oriented trajectory is also informative. If preservation-oriented heads were equally responsible for reasoning, removing them first would degrade accuracy at a comparable rate. The weaker effect of this trajectory suggests that these heads are more involved in routing or maintaining information than in constructing the rule-level transformations needed for multi-step reasoning.

\begin{figure}[H]
\centering
\captionsetup{font=small,skip=3pt}

\includegraphics[width=0.82\textwidth]{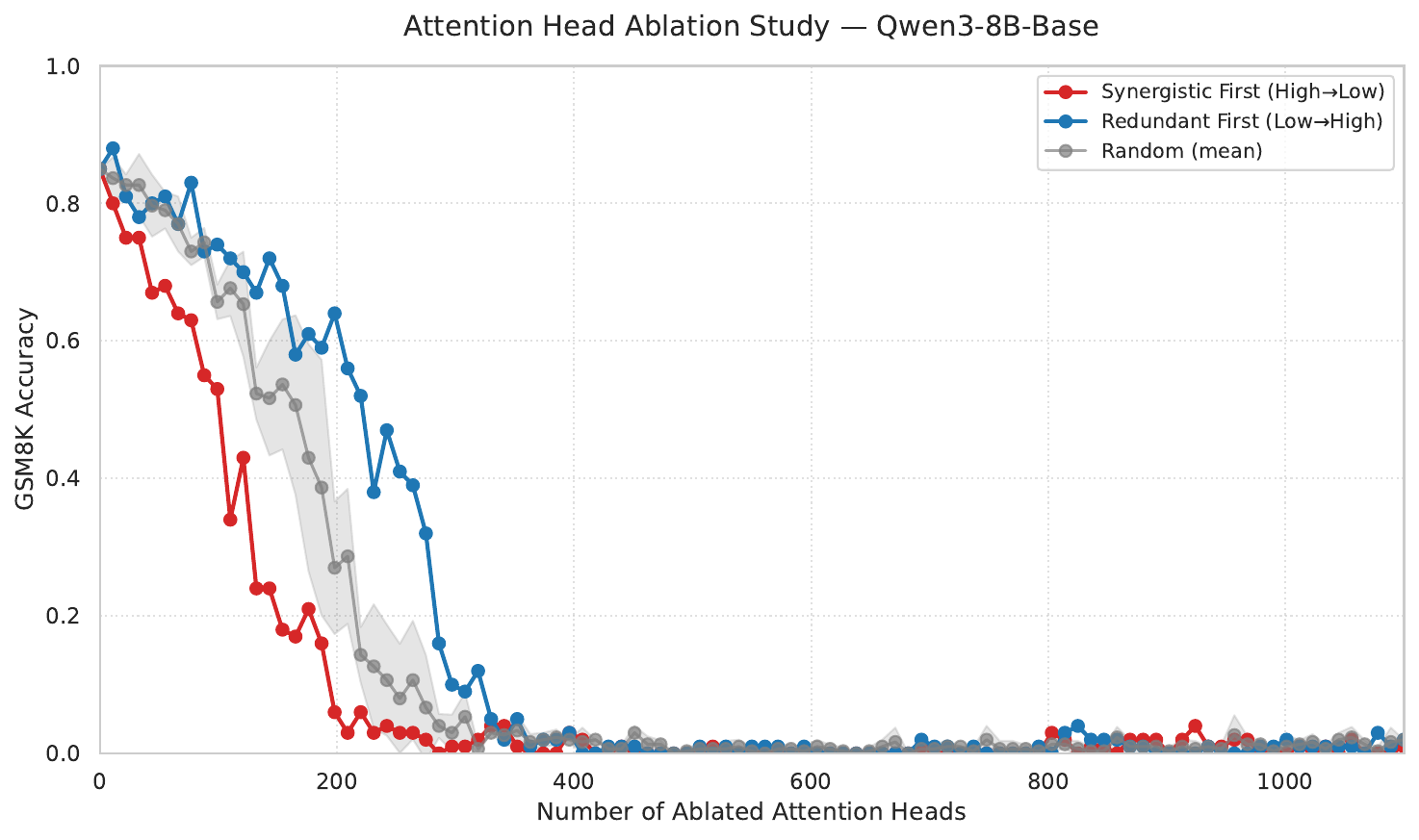}

\caption{
Head-level cumulative ablation in Qwen3-8B-Base.
}
\label{fig:app_qwen3_8b_head_ablation}
\end{figure}

\paragraph{Qwen3-8B-Base.}
Figure~\ref{fig:app_qwen3_8b_head_ablation} reports the same head-level ablation analysis for Qwen3-8B-Base, the main model used for the localization and representation experiments.

\noindent\textbf{Analysis.}
Qwen3-8B-Base is important because it connects the main localization and representation results to head-level functional intervention. The main text uses this model to show that composition-oriented heads concentrate in middle layers and that middle layers carry stronger rule-level representations. The ablation result here provides the corresponding causal test: removing the heads ranked as most composition-oriented produces a larger accuracy degradation.

The random trajectory serves as a baseline for the effect of removing the same number of heads without using the information-decomposition ranking. If the observed degradation were only caused by the number of removed heads, the targeted and random curves would be similar. The stronger degradation under composition-oriented removal therefore suggests that the ranking identifies heads with disproportionate functional importance.

The preservation-oriented trajectory provides the opposite control. These heads can still be important for maintaining information flow, but they are less damaging to remove early than the composition-oriented heads. This supports the paper's interpretation that reasoning performance is more sensitive to components involved in representation transformation than to components mainly associated with preservation.

\begin{figure}[H]
\centering
\captionsetup{font=small,skip=3pt}

\includegraphics[width=0.82\textwidth]{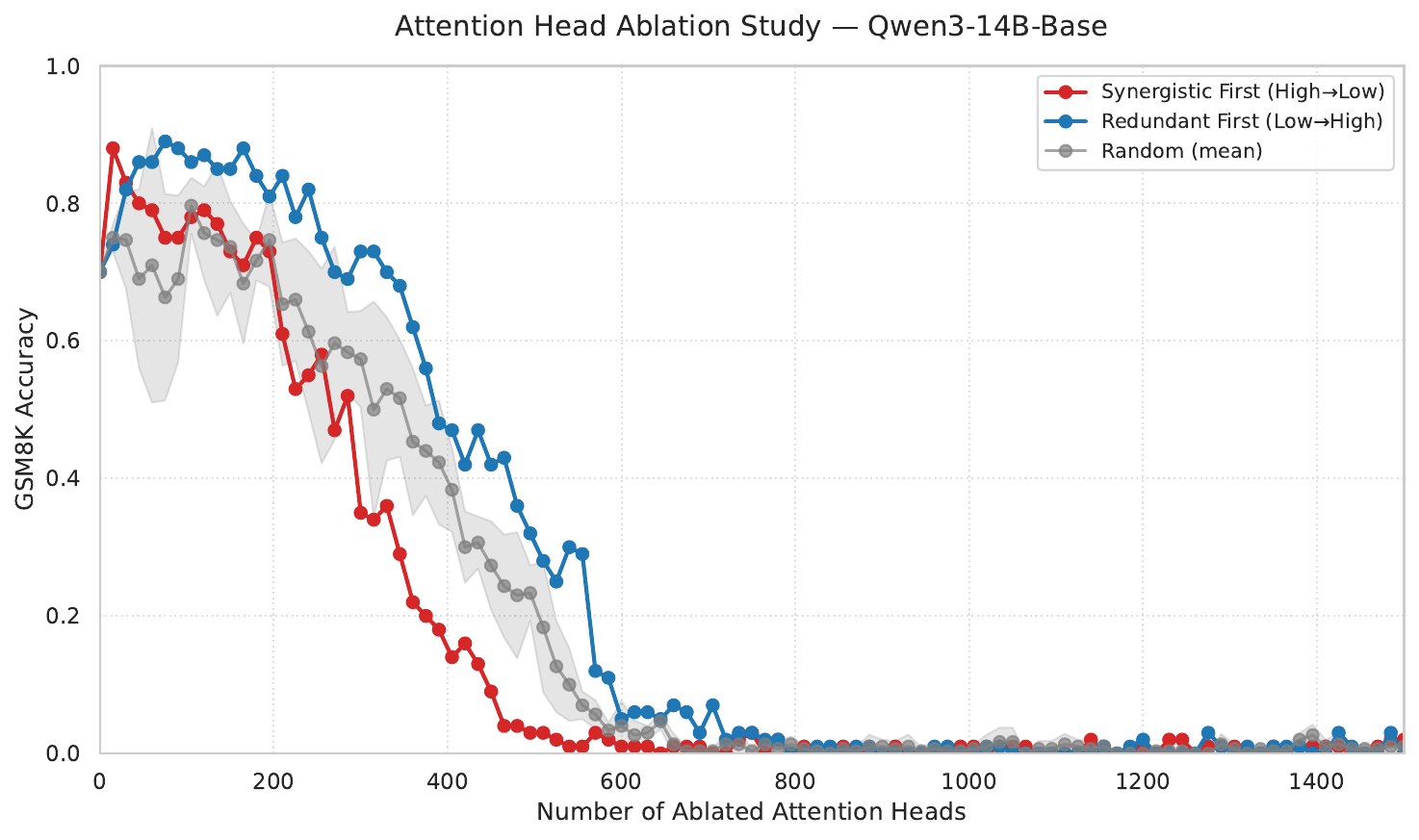}

\caption{
Head-level cumulative ablation in Qwen3-14B-Base.
}
\label{fig:app_qwen3_14b_head_ablation}
\end{figure}

\paragraph{Qwen3-14B-Base.}
Figure~\ref{fig:app_qwen3_14b_head_ablation} shows the cumulative head-ablation result for Qwen3-14B-Base.

\noindent\textbf{Analysis.}
Qwen3-14B-Base tests whether the head-level intervention result persists in a deeper and larger model. Larger models may contain more redundant heads and more distributed pathways, so one might expect cumulative ablation curves to degrade more gradually. Even under this expectation, the relative ordering among trajectories remains the key signal.

The composition-oriented-first trajectory provides evidence that the information-decomposition ranking identifies heads that are more important for reasoning than their preservation-oriented counterparts. The effect may be smoother than in smaller models because the 14B model has more capacity to compensate for the removal of individual heads. However, a consistent gap between targeted and random removal still indicates that the composition-oriented heads carry stronger functional weight.

This result also aligns with the earlier cross-model localization and representation analyses. In Qwen3-14B-Base, abstraction-related signals are spread across a broader internal depth range. The head-level ablation result suggests that this broader distribution does not make the ranking meaningless; rather, the most composition-oriented heads remain disproportionately important even when the model has more layers and more possible compensatory routes.

\begin{figure}[H]
\centering
\captionsetup{font=small,skip=3pt}

\includegraphics[width=0.82\textwidth]{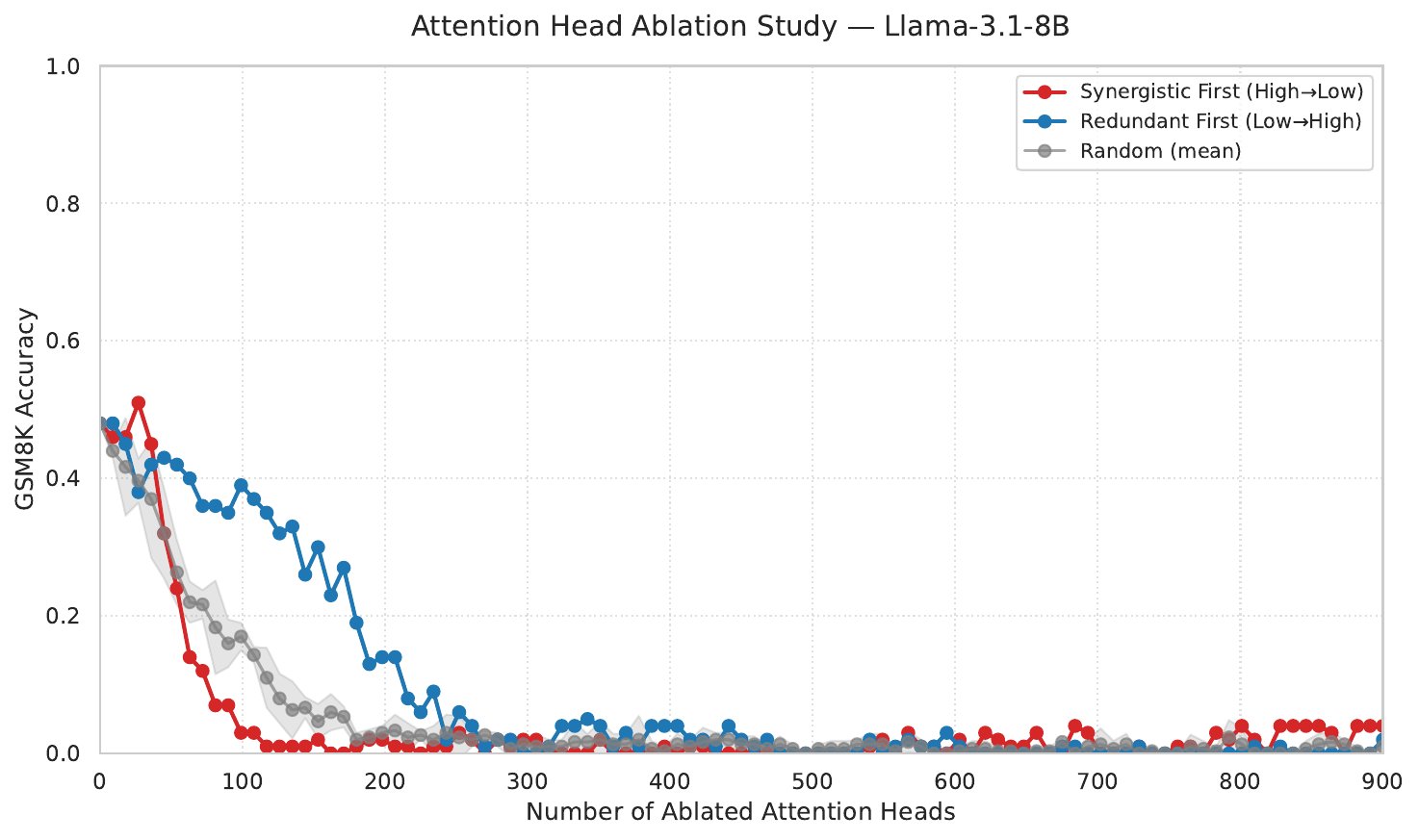}

\caption{
Head-level cumulative ablation in Llama3.1-8B-Base.
}
\label{fig:app_llama31_8b_head_ablation}
\end{figure}

\paragraph{Llama3.1-8B-Base.}
Figure~\ref{fig:app_llama31_8b_head_ablation} reports the head-level ablation result for Llama3.1-8B-Base.

\noindent\textbf{Analysis.}
Llama3.1-8B-Base provides the strongest cross-family test for the head-level ablation result. Its localization and cosine profiles differ from Qwen models, showing more interleaving and stronger directional reorganization in the middle layers. The ablation curve tests whether the same interaction-based criterion still identifies functionally important heads in this different architecture.

The key observation is the relative behavior of the three ablation trajectories. If composition-oriented heads are removed first and the model degrades faster than under random removal, then the ranking has functional meaning beyond the Qwen family. This indicates that the interaction-based orientation score captures a computational role that transfers across model architectures, even though the exact layer-wise profile differs.

The preservation-oriented trajectory again acts as a control. A weaker degradation under preservation-oriented removal suggests that these heads are less directly responsible for the reasoning-sensitive transformations measured by GSM8K accuracy. Together with the Llama localization and representation results, this supports the broader conclusion that internal composition-oriented components are more functionally important for abstract reasoning than components associated mainly with information preservation.

\paragraph{Summary.}
Across models, cumulative head ablation supports the causal interpretation of the composition--preservation orientation score. Removing composition-oriented heads first consistently produces stronger performance degradation than removing preservation-oriented heads first, and typically stronger degradation than random ablation. This indicates that the identified heads are not only statistically distinctive under information-decomposition analysis, but also functionally important for reasoning performance.

The result also complements the layer-level intervention analysis. Layer-level interventions show that middle-stage computations have stronger downstream influence, while head-level ablations show that the attention heads ranked as composition-oriented have stronger behavioral impact. Together, these two intervention views support the same conclusion: abstract reasoning relies more heavily on internal components involved in representation transformation than on components primarily associated with preservation and routing.

\section{Additional Analyses}
\label{app:additional_analyses}

This appendix reports additional network-level analyses of attention-head interactions. These analyses are not used as the main evidence in the paper, but provide an auxiliary visualization of how composition- and preservation-oriented interactions are organized among attention heads.

For each model, we construct undirected weighted graphs over attention heads. Each node corresponds to an attention head, and each edge weight is derived from the pairwise interaction score between two heads. We consider two graphs: a composition-oriented graph $G_{\mathrm{comp}}$ with edge weights given by $S_{\mathrm{comp}}(h_i,h_j)$, and a preservation-oriented graph $G_{\mathrm{pres}}$ with edge weights given by $S_{\mathrm{pres}}(h_i,h_j)$. For graph visualization, the same normalization and sparsification procedure is applied across models.

We quantify two graph-level properties. First, global efficiency measures how easily information can be exchanged over the weighted interaction graph. Let $d_{ij}$ denote the shortest-path distance between nodes $i$ and $j$, where edge distance is inversely related to edge weight. The global efficiency is
\begin{equation}
E(G)
=
\frac{1}{N(N-1)}
\sum_{i\neq j}
\frac{1}{d_{ij}} .
\label{app:eq:global_efficiency}
\end{equation}
Higher global efficiency indicates a more globally integrated interaction structure.

Second, modularity measures the degree to which the graph decomposes into relatively separated communities. Let $w_{ij}$ be the edge weight between nodes $i$ and $j$, $k_i=\sum_j w_{ij}$ be the weighted degree of node $i$, and $m=\frac{1}{2}\sum_{i,j}w_{ij}$. Given a community assignment $c_i$, the modularity is
\begin{equation}
Q(G)
=
\frac{1}{2m}
\sum_{i,j}
\left(
w_{ij}
-
\frac{k_i k_j}{2m}
\right)
\delta(c_i,c_j).
\label{app:eq:modularity}
\end{equation}
Higher modularity indicates a more community-structured interaction graph.

\begin{figure}[H]
\centering
\captionsetup{font=small,skip=3pt}
\captionsetup[subfigure]{font=small,skip=2pt}

\begin{subfigure}[t]{0.48\textwidth}
    \centering
    \includegraphics[width=\linewidth]{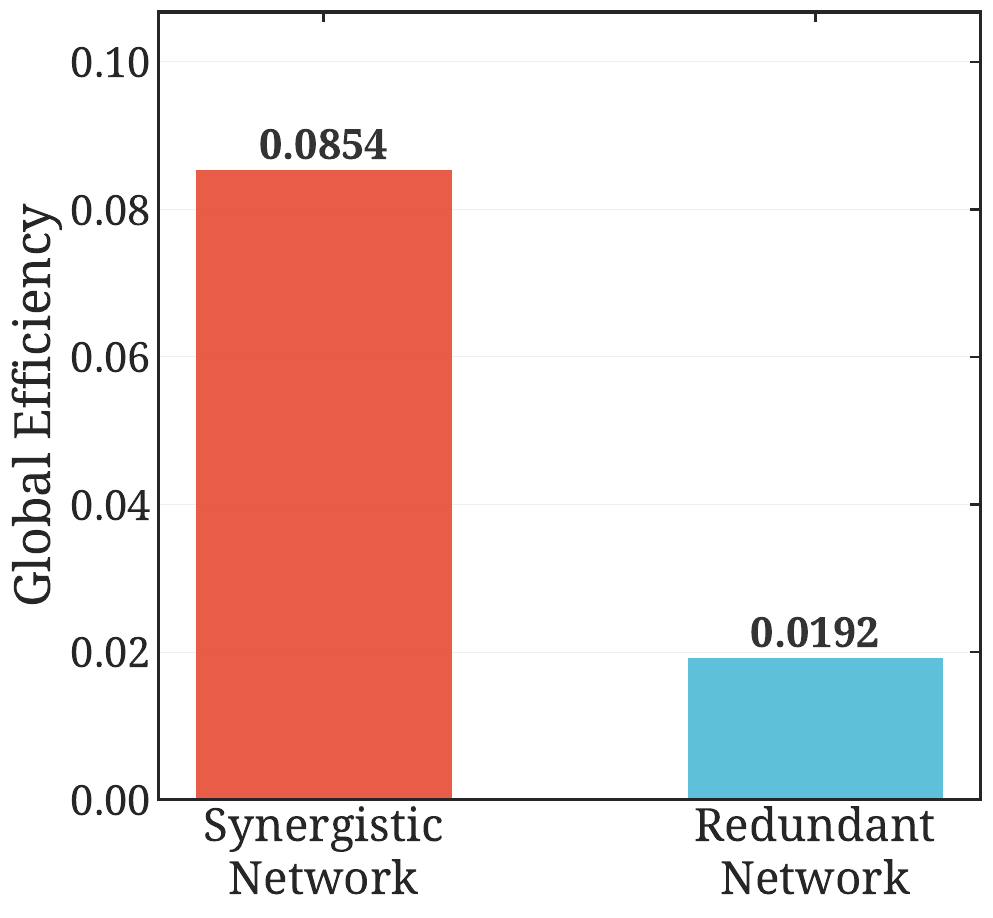}
    \caption{Global efficiency}
\end{subfigure}
\hfill
\begin{subfigure}[t]{0.48\textwidth}
    \centering
    \includegraphics[width=\linewidth]{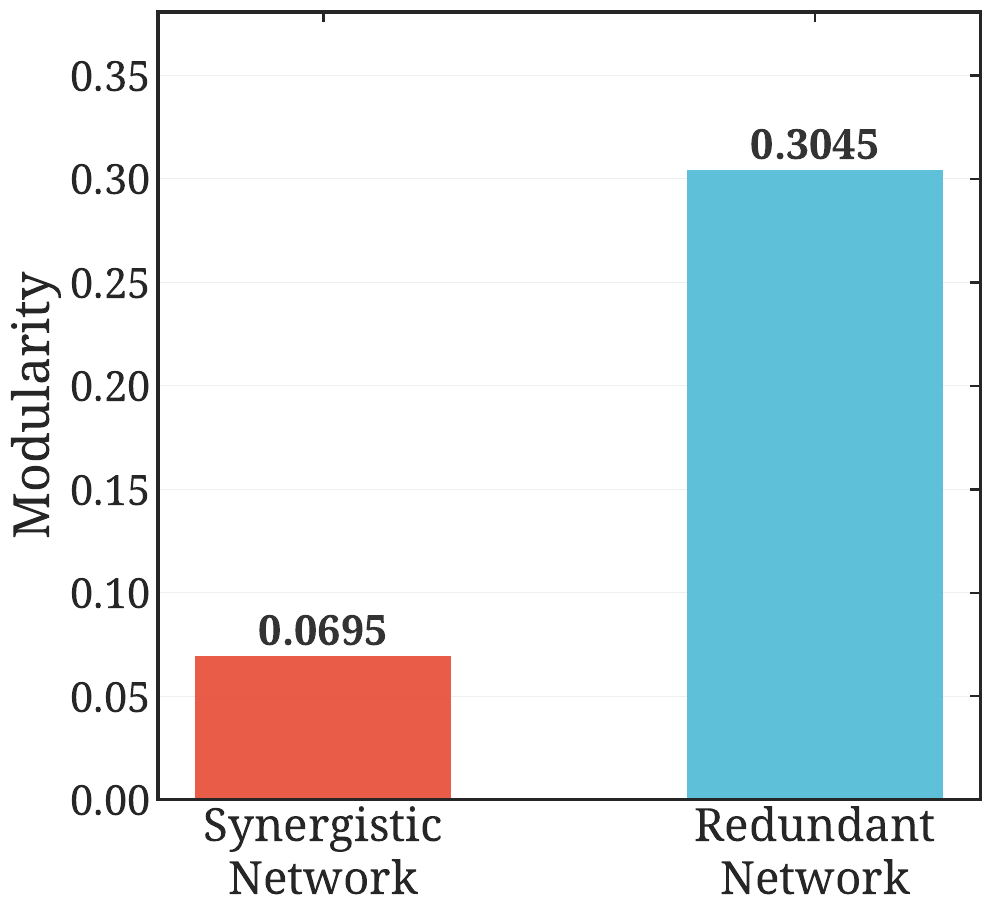}
    \caption{Modularity}
\end{subfigure}

\begin{subfigure}[t]{0.48\textwidth}
    \centering
    \includegraphics[width=\linewidth]{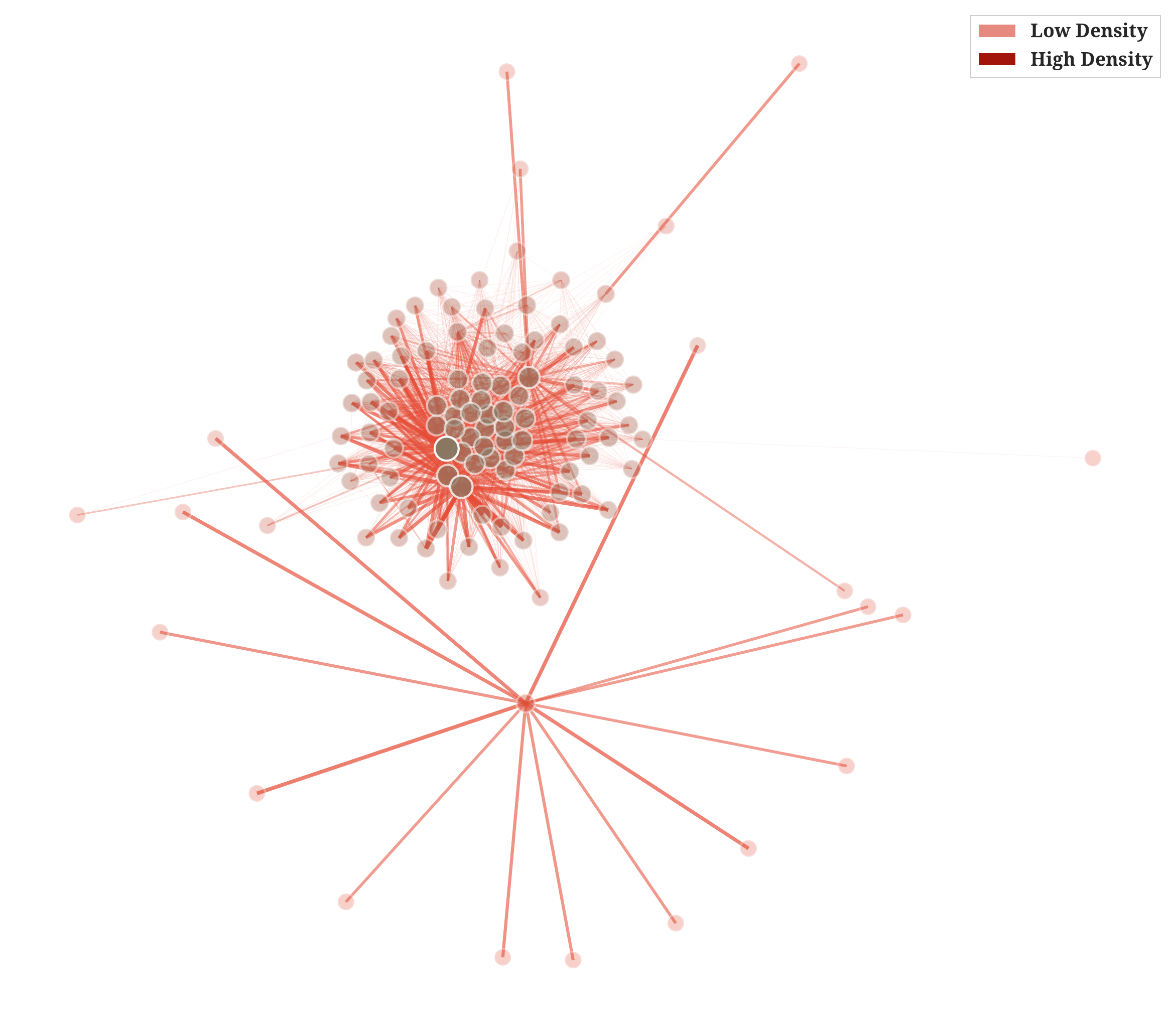}
    \caption{Composition-oriented graph}
\end{subfigure}
\hfill
\begin{subfigure}[t]{0.48\textwidth}
    \centering
    \includegraphics[width=\linewidth]{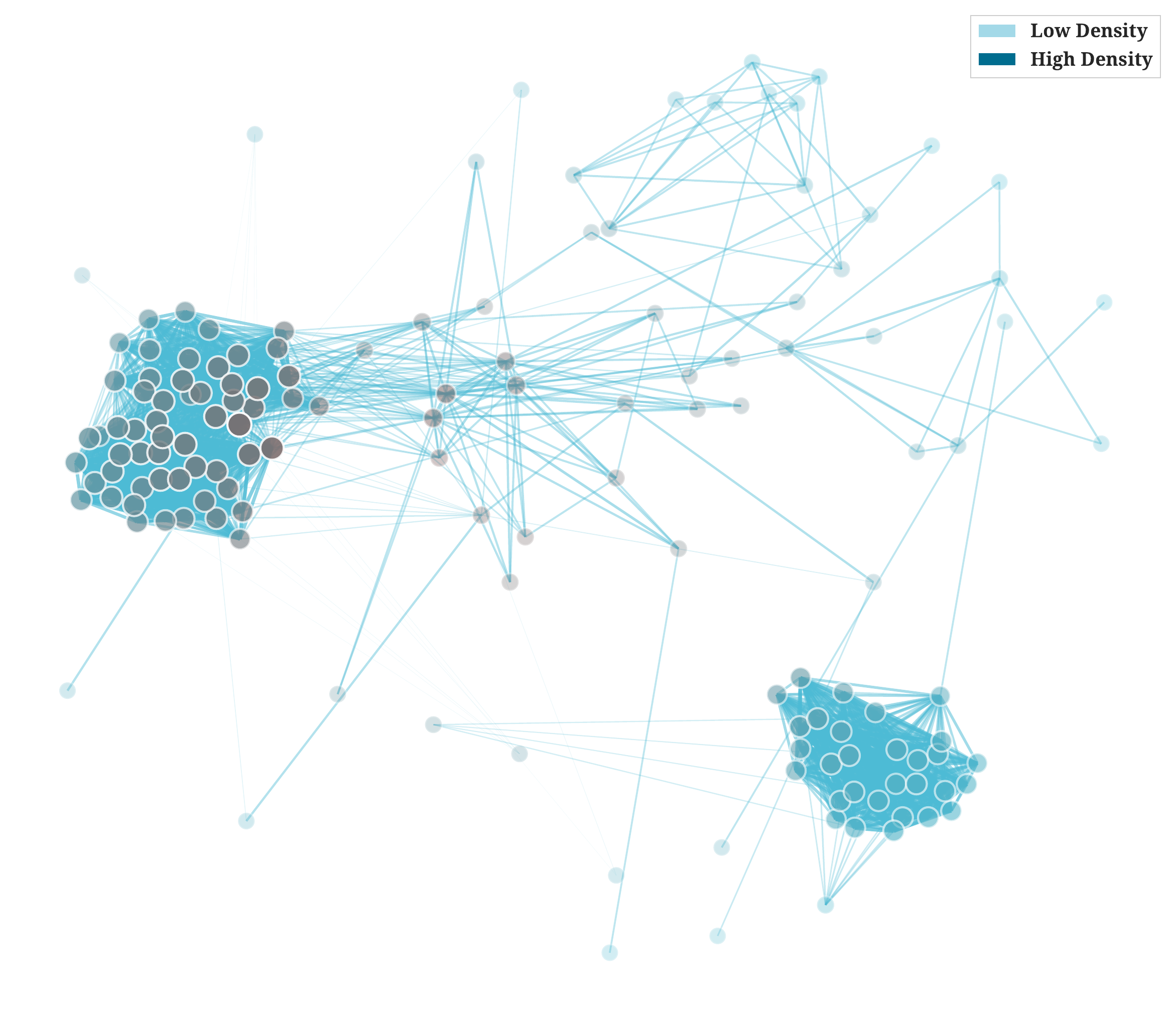}
    \caption{Preservation-oriented graph}
\end{subfigure}

\caption{
Additional network-level analysis for Qwen3-4B-Base.
}
\label{fig:app_qwen3_4b_network}
\end{figure}

\paragraph{Qwen3-4B-Base.}
Figure~\ref{fig:app_qwen3_4b_network} shows the network-level analysis for Qwen3-4B-Base.

\noindent\textbf{Analysis.}
Qwen3-4B-Base shows a relatively compact interaction structure, consistent with its smaller depth and parameter scale. The global-efficiency curve provides an auxiliary view of how broadly interaction information can propagate across heads, while the modularity curve reflects whether heads form more separated functional communities. The two graph visualizations suggest that composition-oriented interactions are more globally connected, whereas preservation-oriented interactions tend to form more locally clustered structures. This supports the main interpretation that composition-oriented components participate in representation transformation, while preservation-oriented components are more related to maintaining or routing information.

\begin{figure}[H]
\centering
\captionsetup{font=small,skip=3pt}
\captionsetup[subfigure]{font=small,skip=2pt}

\begin{subfigure}[t]{0.48\textwidth}
    \centering
    \includegraphics[width=\linewidth]{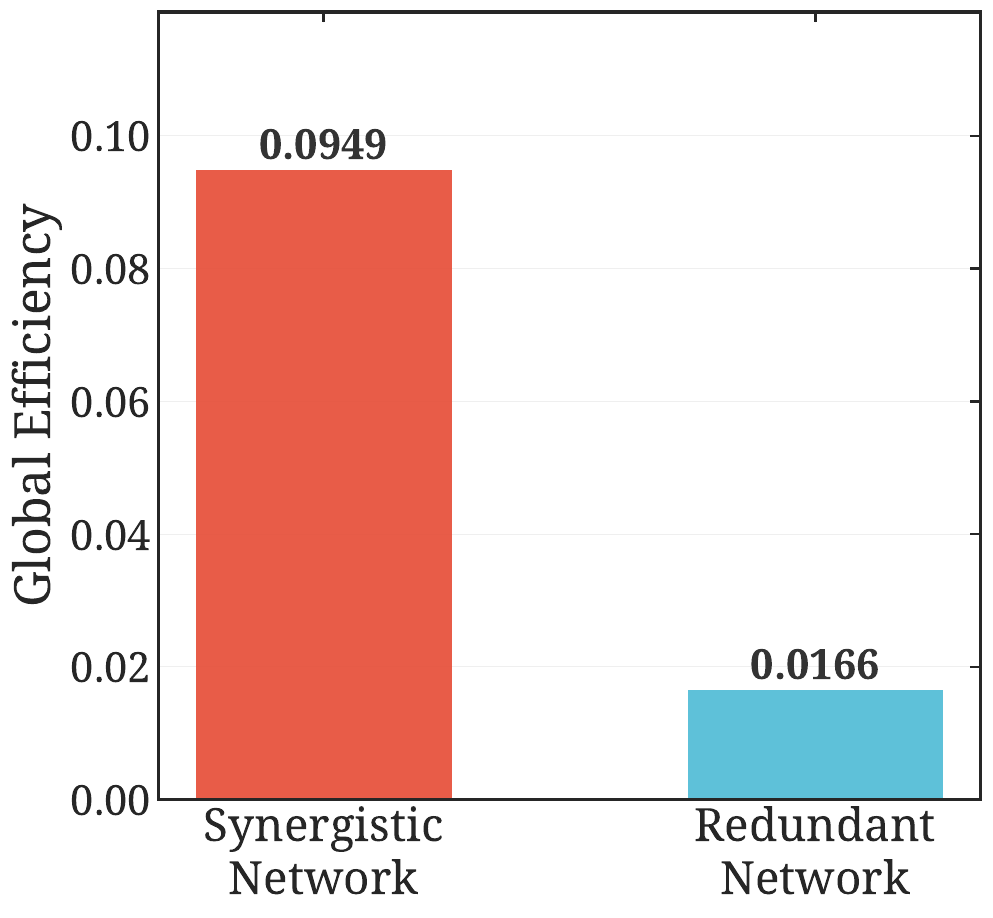}
    \caption{Global efficiency}
\end{subfigure}
\hfill
\begin{subfigure}[t]{0.48\textwidth}
    \centering
    \includegraphics[width=\linewidth]{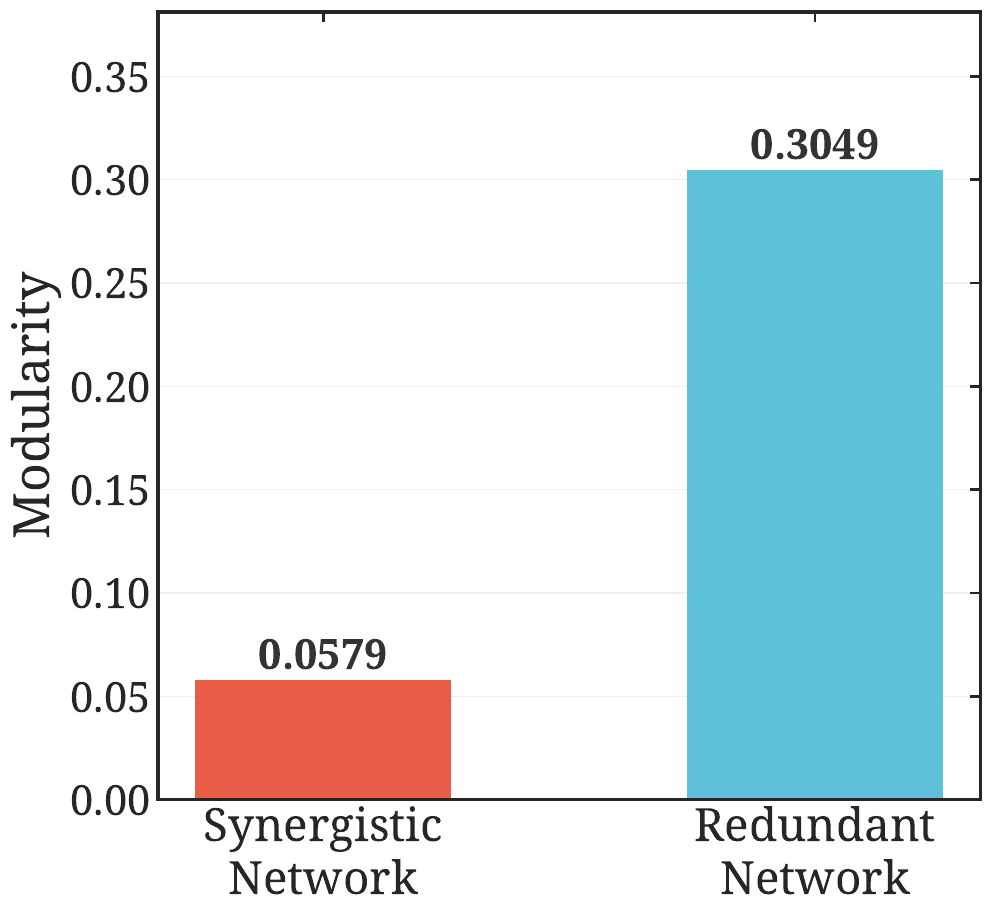}
    \caption{Modularity}
\end{subfigure}

\begin{subfigure}[t]{0.48\textwidth}
    \centering
    \includegraphics[width=\linewidth]{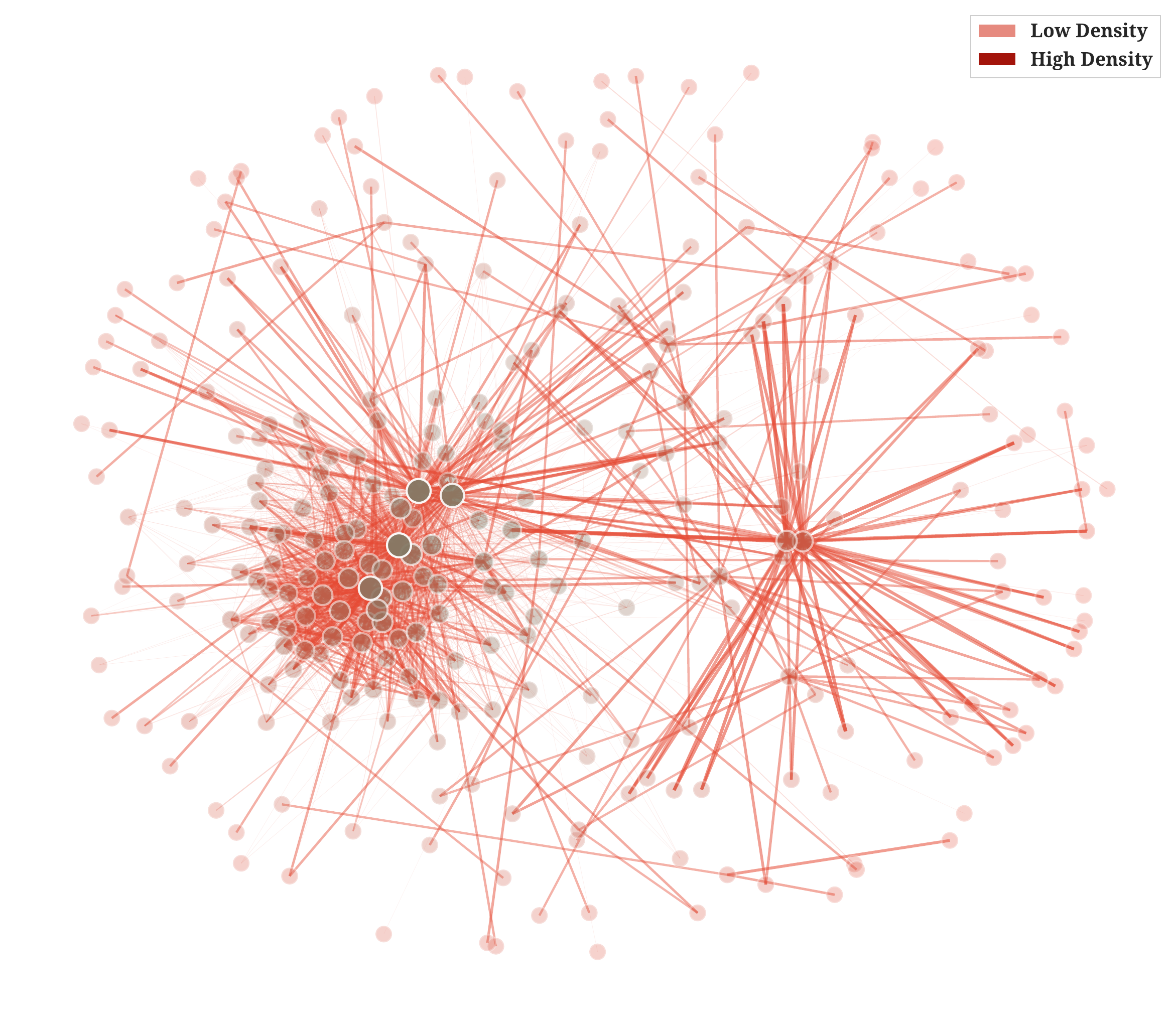}
    \caption{Composition-oriented graph}
\end{subfigure}
\hfill
\begin{subfigure}[t]{0.48\textwidth}
    \centering
    \includegraphics[width=\linewidth]{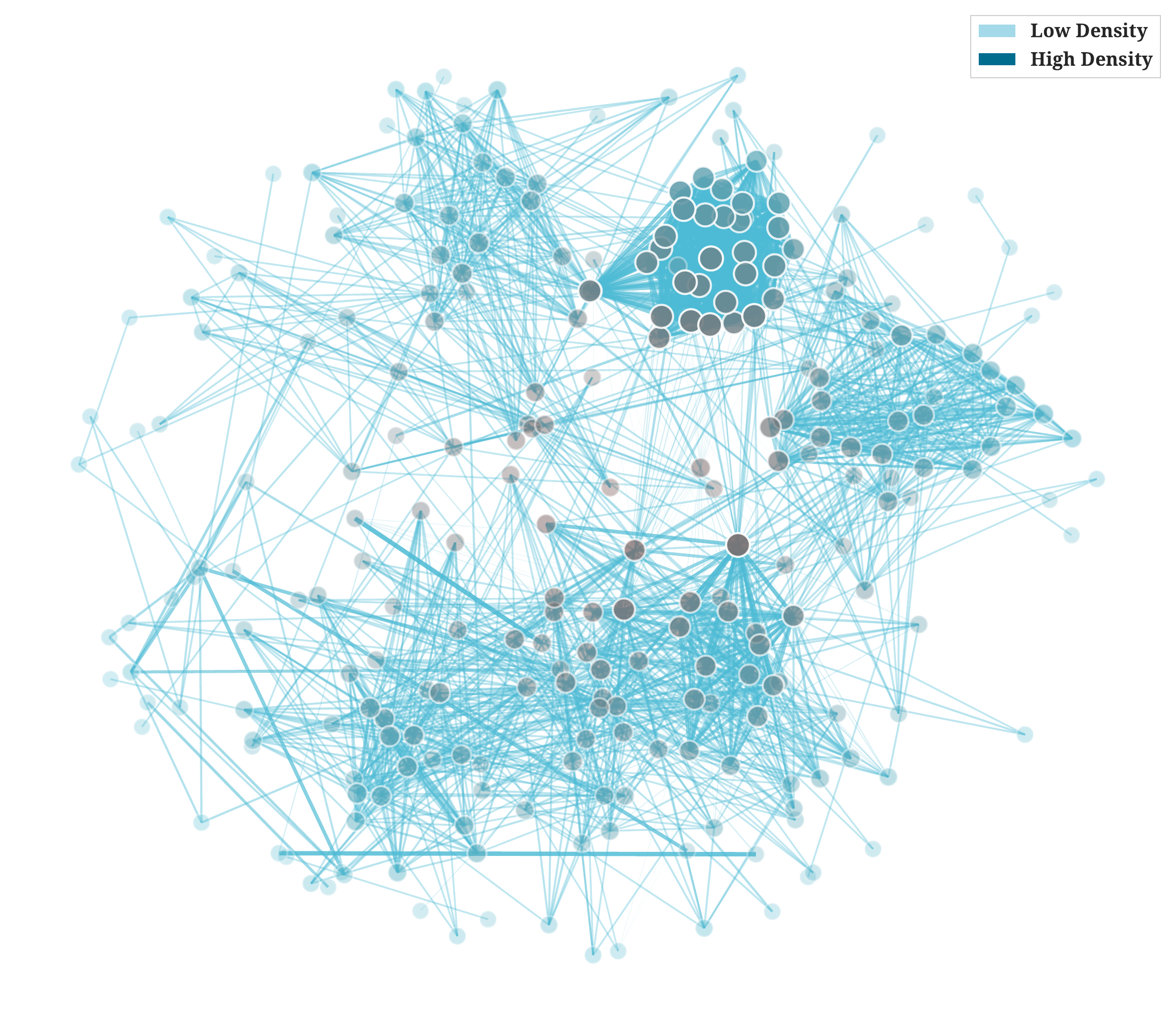}
    \caption{Preservation-oriented graph}
\end{subfigure}

\caption{
Additional network-level analysis for Qwen3-8B-Base.
}
\label{fig:app_qwen3_8b_network}
\end{figure}

\paragraph{Qwen3-8B-Base.}
Figure~\ref{fig:app_qwen3_8b_network} shows the same network-level analysis for the main Qwen3-8B-Base model.

\noindent\textbf{Analysis.}
For Qwen3-8B-Base, the network-level view is consistent with the main-text localization results. The composition-oriented graph exhibits a more integrated structure in the intermediate computation stage, suggesting that heads involved in representation transformation interact across a broader portion of the model. In contrast, preservation-oriented interactions appear more modular, indicating that information-preserving components may operate through more localized or separated groups. This graph-level pattern provides a complementary view of the same division of labor reported in the main experiments.

\begin{figure}[H]
\centering
\captionsetup{font=small,skip=3pt}
\captionsetup[subfigure]{font=small,skip=2pt}

\begin{subfigure}[t]{0.48\textwidth}
    \centering
    \includegraphics[width=\linewidth]{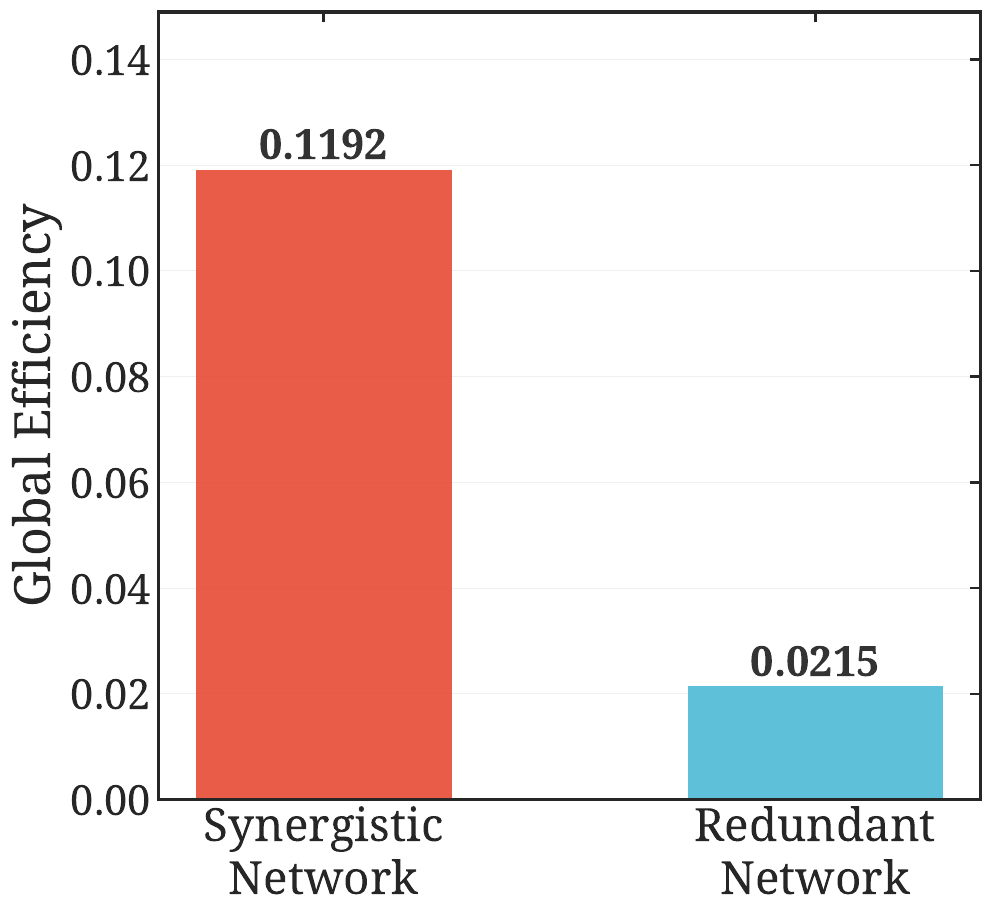}
    \caption{Global efficiency}
\end{subfigure}
\hfill
\begin{subfigure}[t]{0.48\textwidth}
    \centering
    \includegraphics[width=\linewidth]{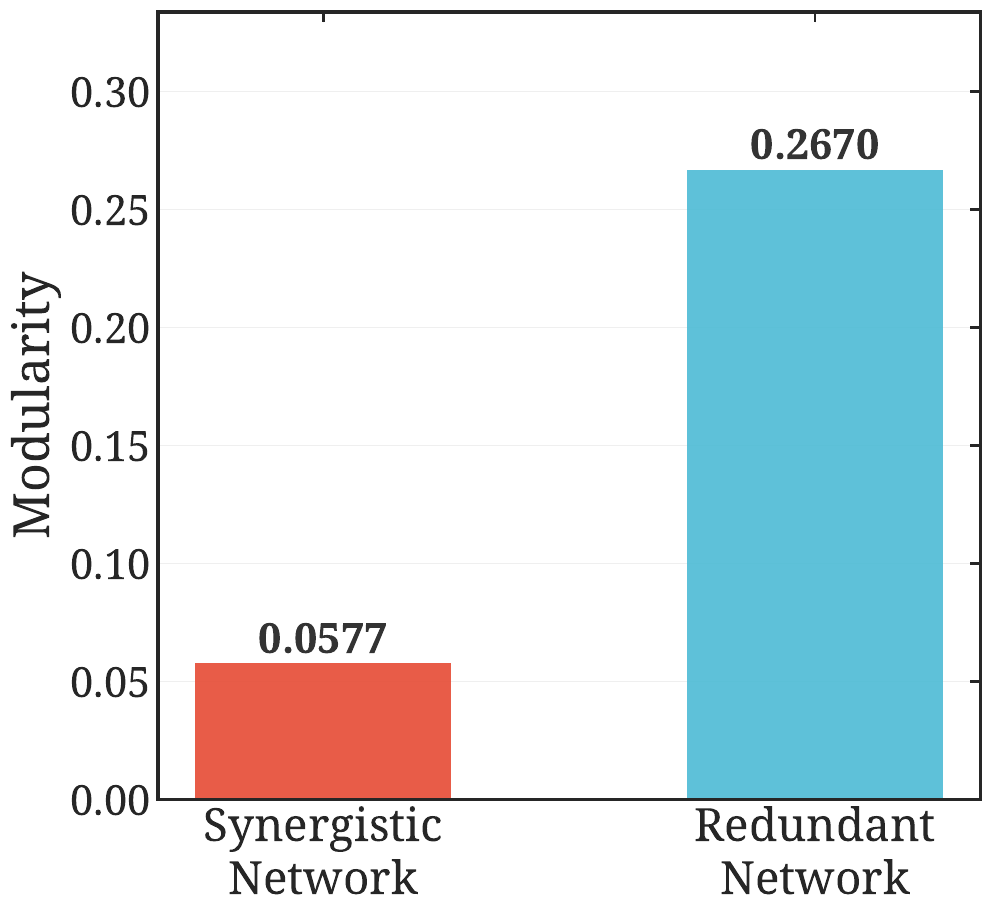}
    \caption{Modularity}
\end{subfigure}

\begin{subfigure}[t]{0.48\textwidth}
    \centering
    \includegraphics[width=\linewidth]{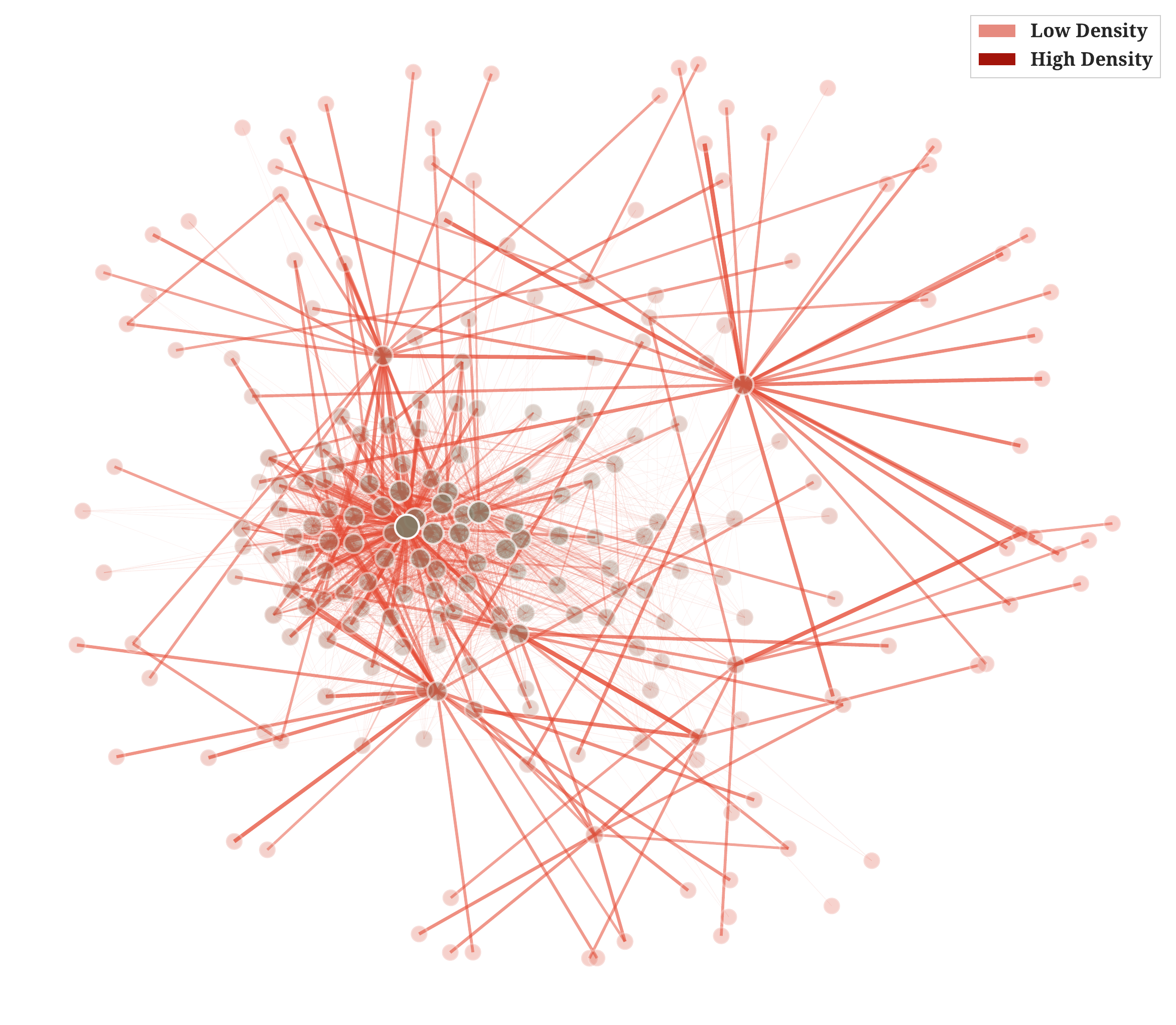}
    \caption{Composition-oriented graph}
\end{subfigure}
\hfill
\begin{subfigure}[t]{0.48\textwidth}
    \centering
    \includegraphics[width=\linewidth]{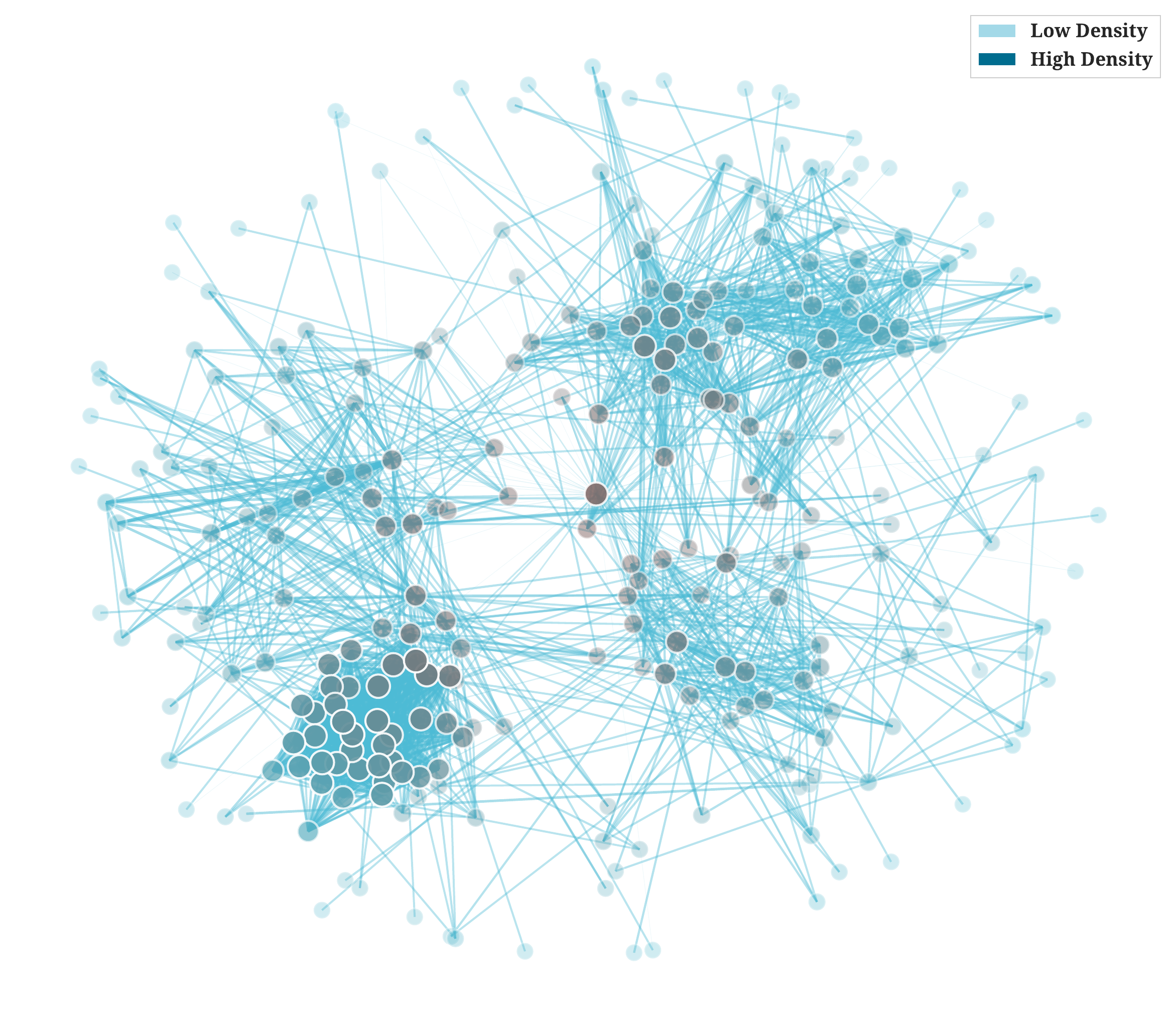}
    \caption{Preservation-oriented graph}
\end{subfigure}

\caption{
Additional network-level analysis for Qwen3-14B-Base.
}
\label{fig:app_qwen3_14b_network}
\end{figure}

\paragraph{Qwen3-14B-Base.}
Figure~\ref{fig:app_qwen3_14b_network} reports the network-level analysis for Qwen3-14B-Base.

\noindent\textbf{Analysis.}
Qwen3-14B-Base shows a broader and more distributed network structure than the smaller Qwen models. This is expected given its greater depth and number of heads. The global-efficiency and modularity trends suggest that the model can distribute interaction structure across a larger internal region while still preserving the distinction between globally integrated composition-oriented interactions and more community-structured preservation-oriented interactions. Thus, the graph-level analysis agrees with the earlier observation that larger models may stretch the middle computation stage over a wider depth interval.

\begin{figure}[H]
\centering
\captionsetup{font=small,skip=3pt}
\captionsetup[subfigure]{font=small,skip=2pt}

\begin{subfigure}[t]{0.48\textwidth}
    \centering
    \includegraphics[width=\linewidth]{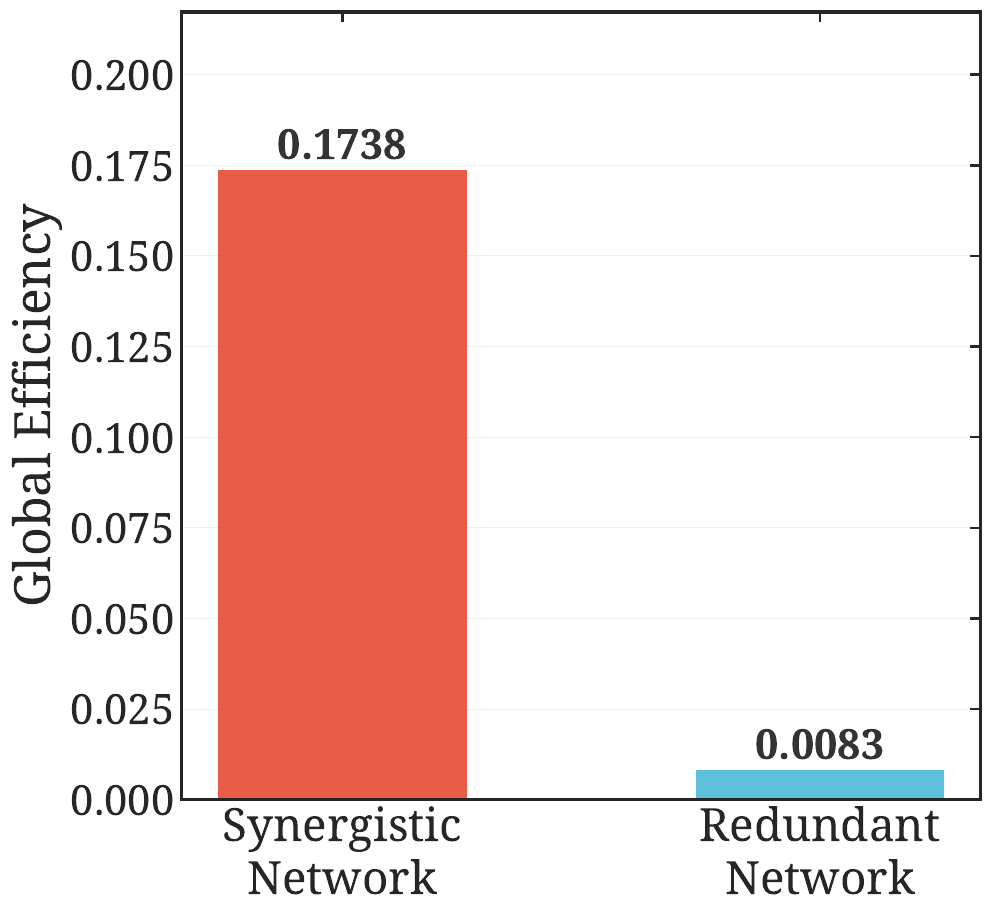}
    \caption{Global efficiency}
\end{subfigure}
\hfill
\begin{subfigure}[t]{0.48\textwidth}
    \centering
    \includegraphics[width=\linewidth]{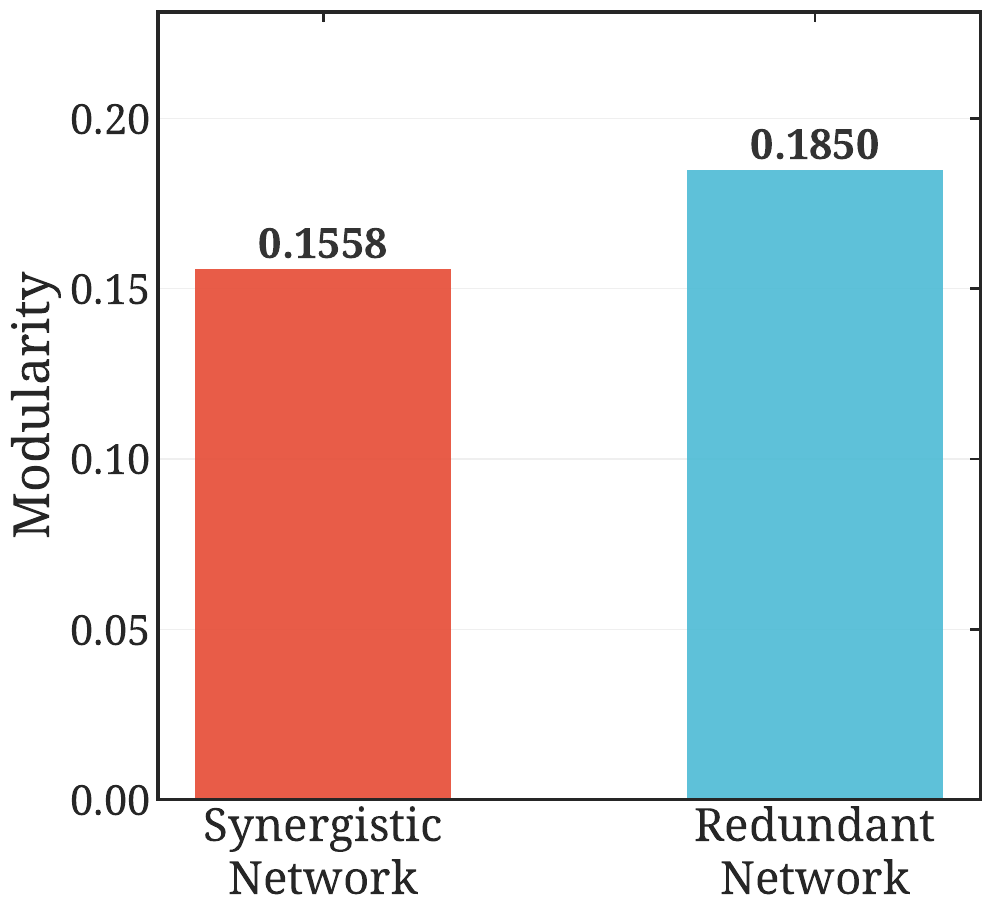}
    \caption{Modularity}
\end{subfigure}

\begin{subfigure}[t]{0.48\textwidth}
    \centering
    \includegraphics[width=\linewidth]{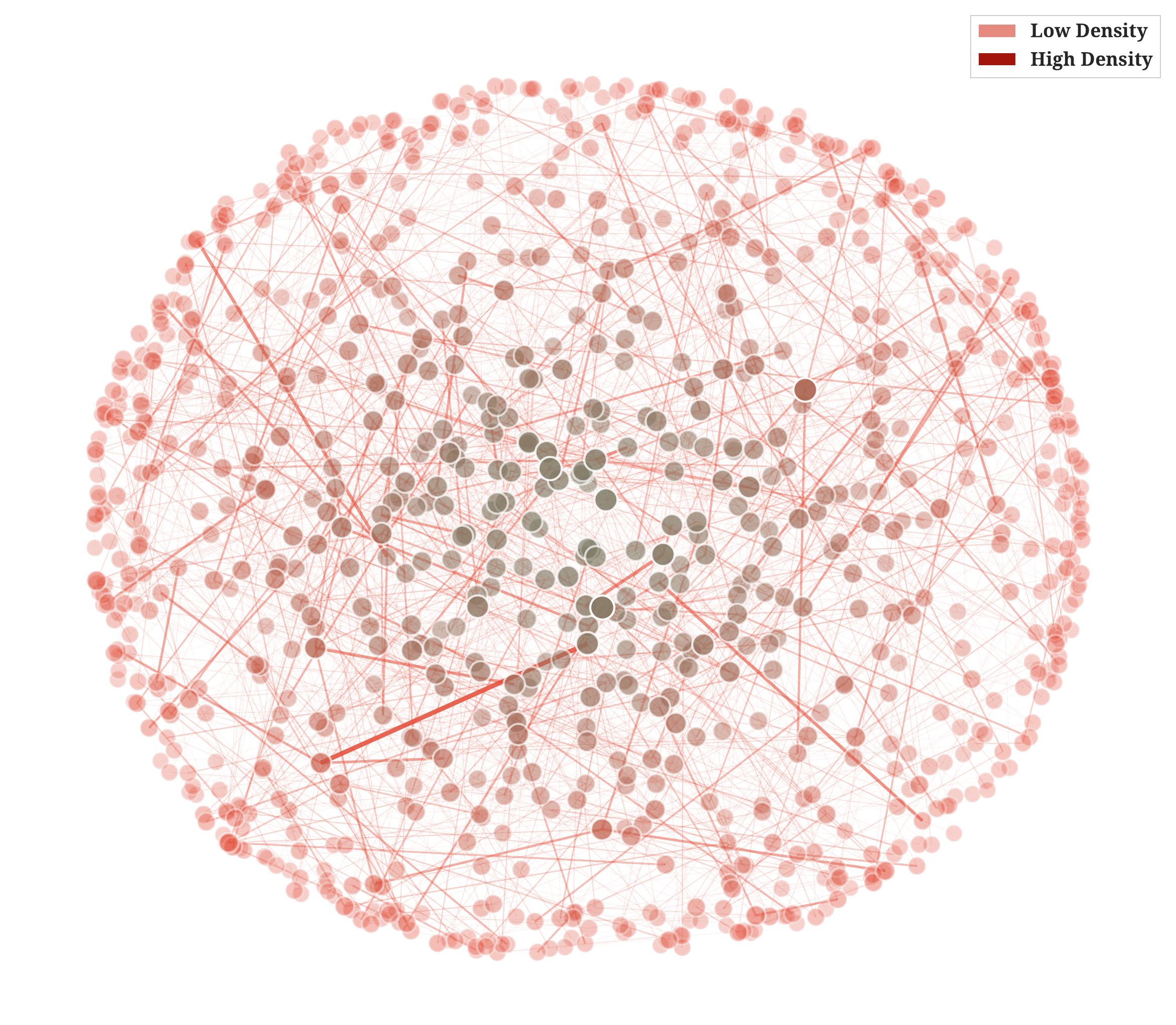}
    \caption{Composition-oriented graph}
\end{subfigure}
\hfill
\begin{subfigure}[t]{0.48\textwidth}
    \centering
    \includegraphics[width=\linewidth]{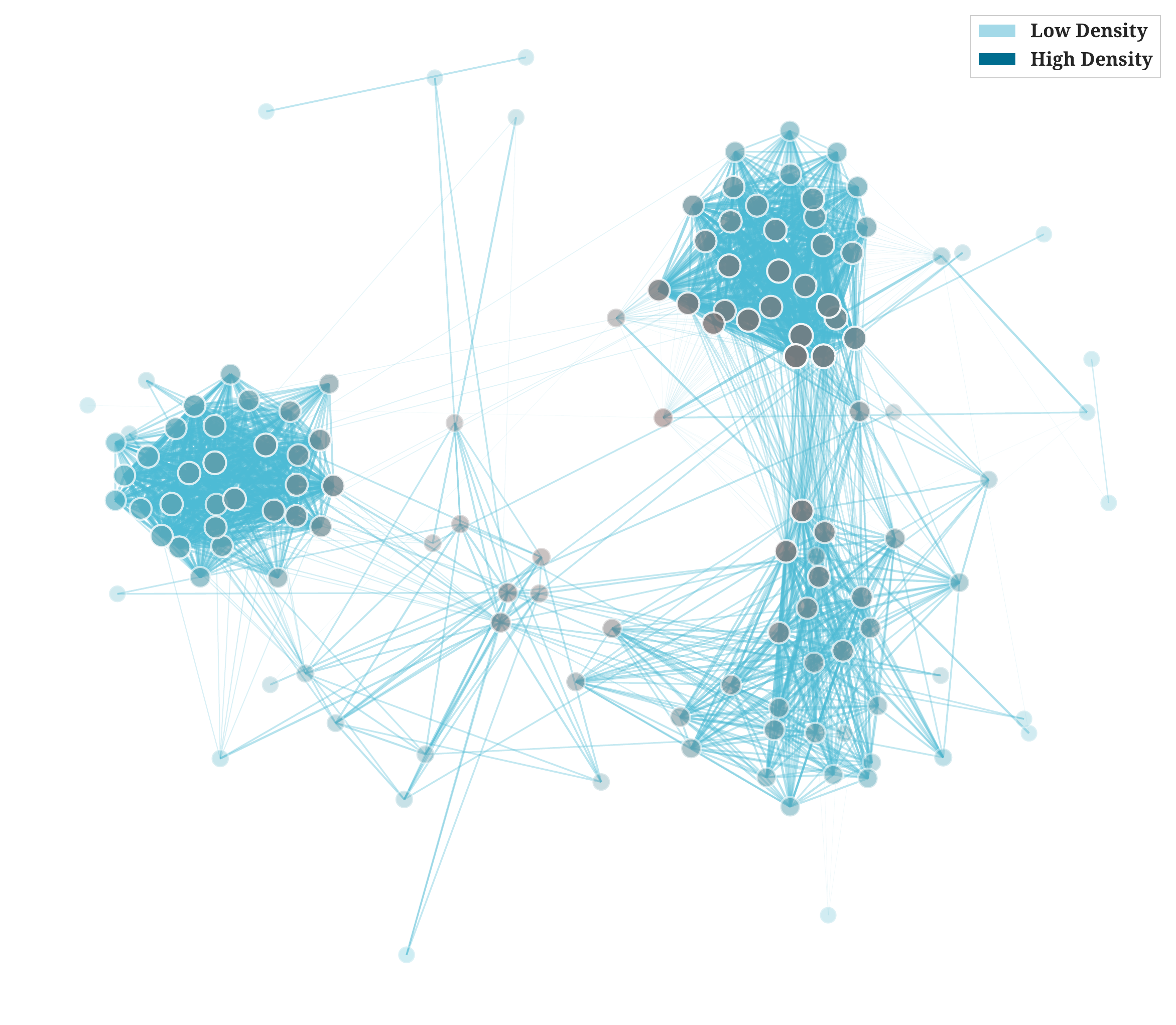}
    \caption{Preservation-oriented graph}
\end{subfigure}

\caption{
Additional network-level analysis for Llama3.1-8B-Base.
}
\label{fig:app_llama31_8b_network}
\end{figure}

\paragraph{Llama3.1-8B-Base.}
Figure~\ref{fig:app_llama31_8b_network} provides the network-level analysis for Llama3.1-8B-Base.

\noindent\textbf{Analysis.}
Llama3.1-8B-Base shows a more architecture-specific network structure. This is consistent with the earlier localization and cosine-similarity results, where Llama exhibited stronger oscillation and directional reorganization in intermediate layers. The composition-oriented graph provides an auxiliary view of these transformations: interactions are not uniformly distributed but are organized through a subset of internally connected heads. The preservation-oriented graph, by contrast, shows a more clustered organization, suggesting that information-maintaining functions are more locally structured. This supports the cross-family generality of the composition--preservation distinction.

\begin{figure}[H]
\centering
\captionsetup{font=small,skip=3pt}
\captionsetup[subfigure]{font=small,skip=2pt}

\begin{subfigure}[t]{0.48\textwidth}
    \centering
    \includegraphics[width=\linewidth]{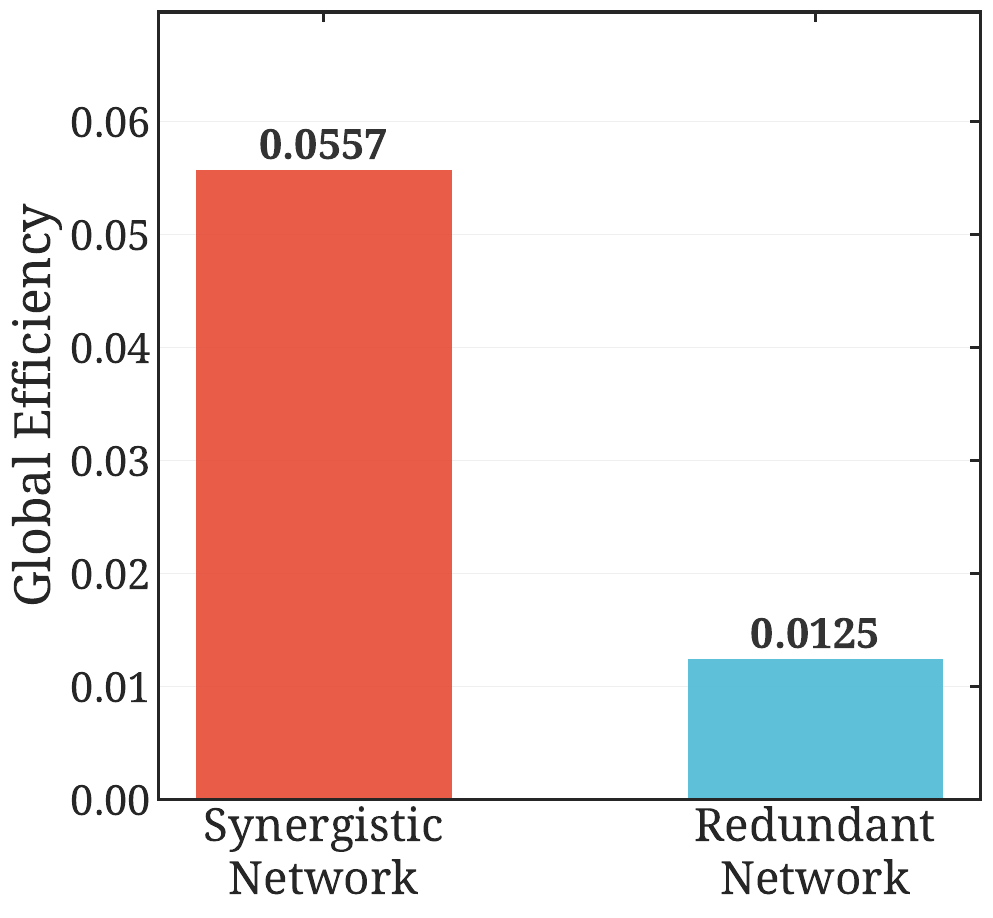}
    \caption{Global efficiency}
\end{subfigure}
\hfill
\begin{subfigure}[t]{0.48\textwidth}
    \centering
    \includegraphics[width=\linewidth]{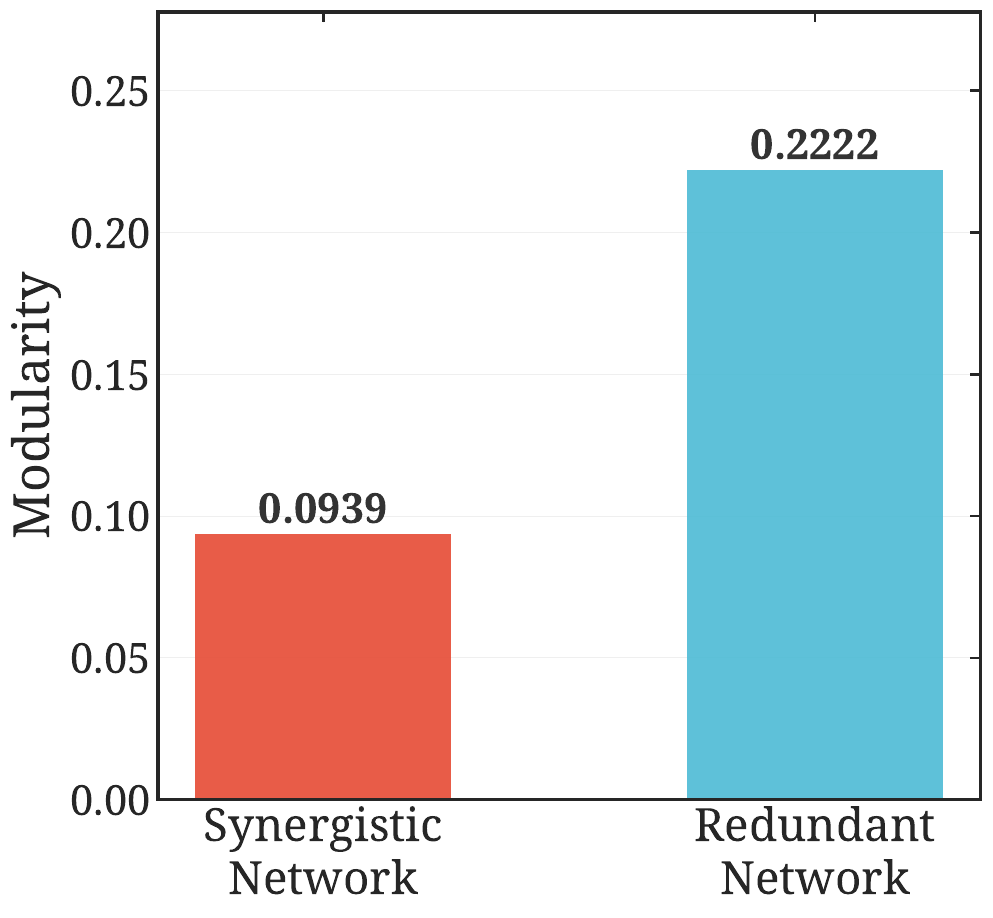}
    \caption{Modularity}
\end{subfigure}

\begin{subfigure}[t]{0.48\textwidth}
    \centering
    \includegraphics[width=\linewidth]{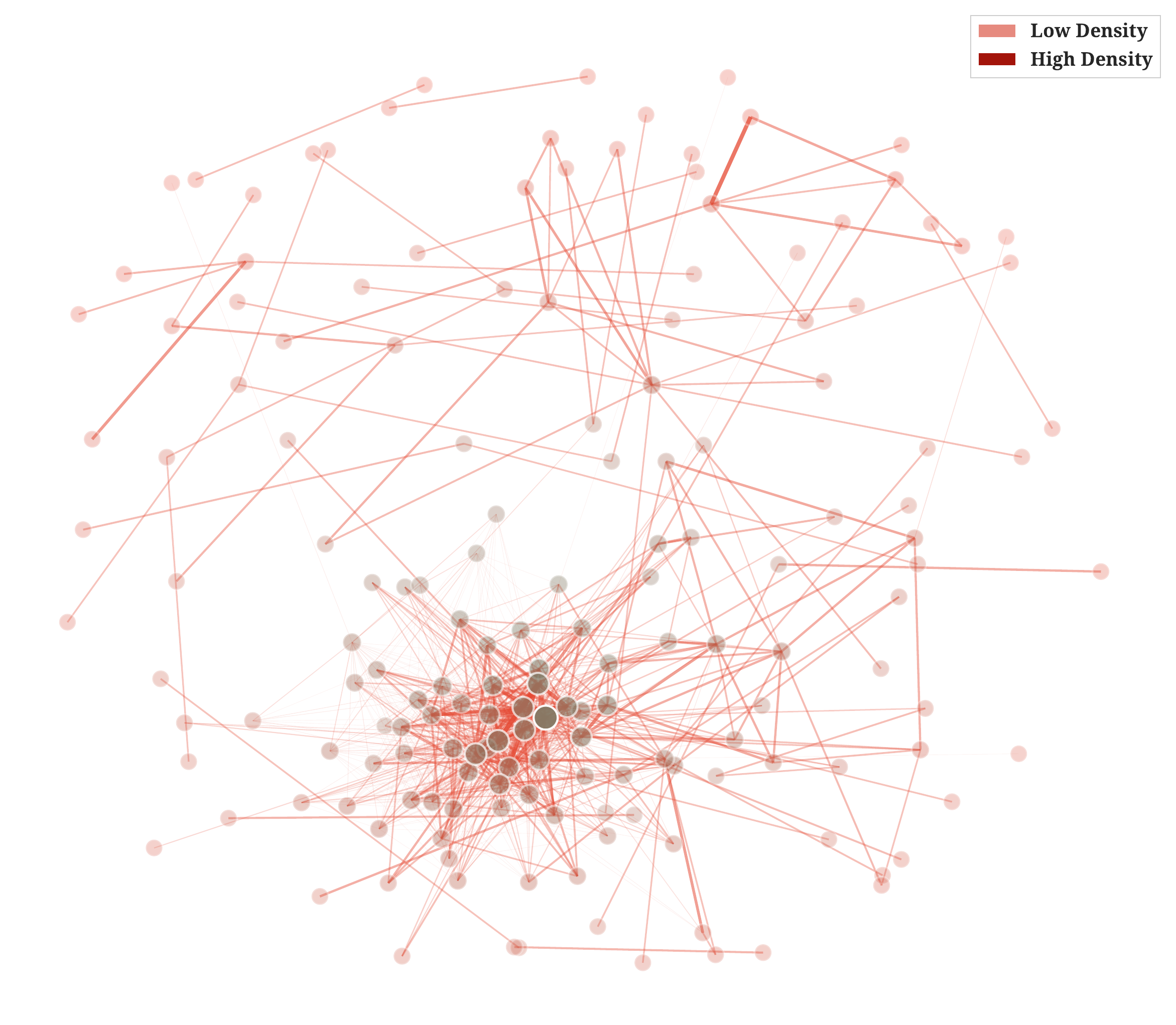}
    \caption{Composition-oriented graph}
\end{subfigure}
\hfill
\begin{subfigure}[t]{0.48\textwidth}
    \centering
    \includegraphics[width=\linewidth]{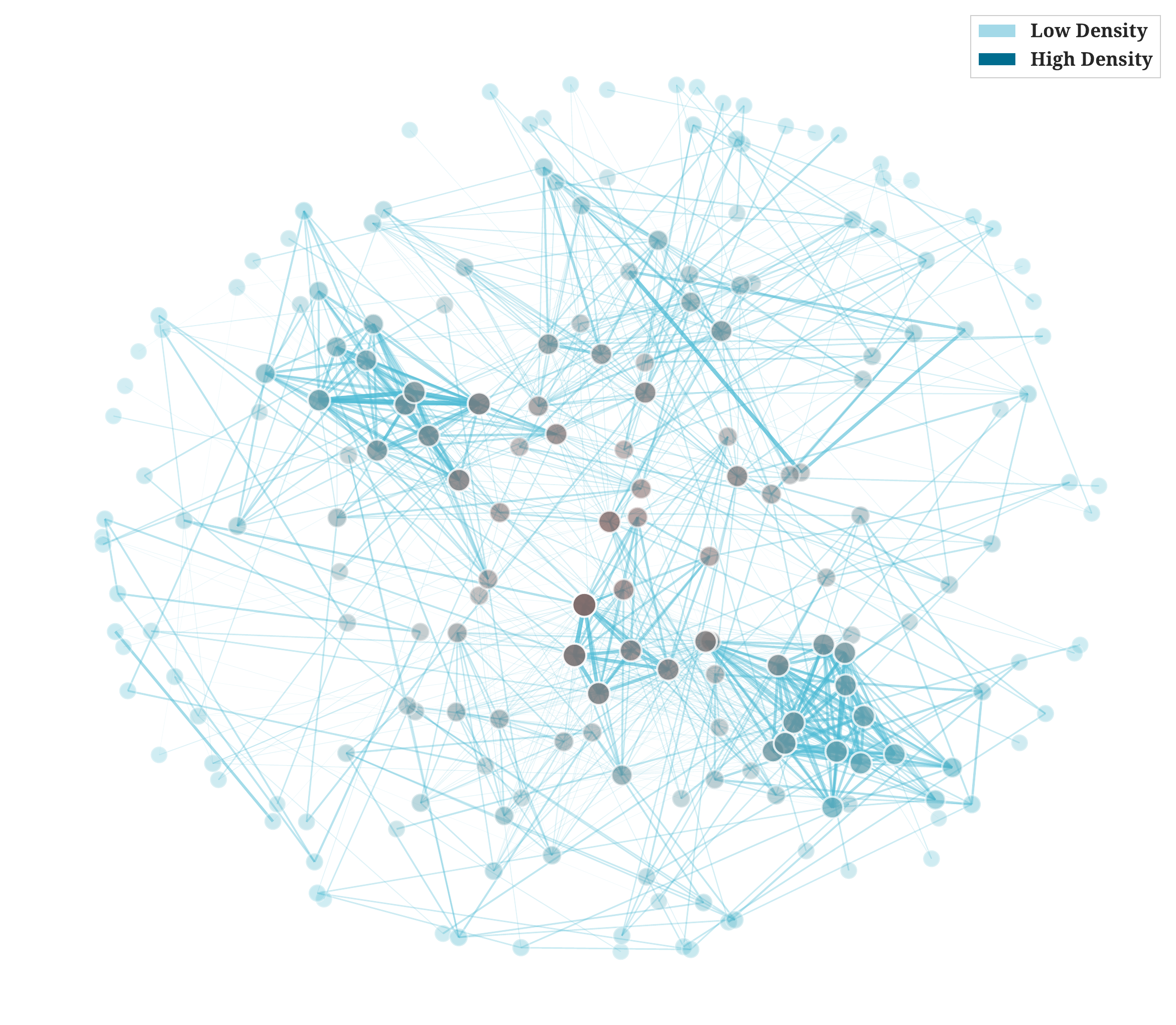}
    \caption{Preservation-oriented graph}
\end{subfigure}

\caption{
Additional network-level analysis for Gemma3-12B-Instruct.
}
\label{fig:app_gemma3_12b_instruct_network}
\end{figure}

\paragraph{Gemma3-12B-Instruct.}
Figure~\ref{fig:app_gemma3_12b_instruct_network} shows the network-level analysis for Gemma3-12B-Instruct.

\noindent\textbf{Analysis.}
Gemma3-12B-Instruct exhibits a broader and more locally variable network organization. This matches the earlier layer-wise results, where Gemma showed more oscillatory interaction and cosine profiles. The global-efficiency and modularity measures suggest that its composition-oriented interactions are still organized differently from preservation-oriented interactions, but the separation is less sharply expressed than in Qwen models. This may reflect the effect of instruction tuning or model-family differences. Nevertheless, the graph-level analysis remains consistent with the broader conclusion: composition-oriented dynamics are associated with more integrated internal interactions, while preservation-oriented dynamics are more compatible with modular information routing.

\paragraph{Summary.}
These network-level analyses provide an auxiliary view of the interaction structure among attention heads. Across models, composition-oriented graphs tend to be more globally integrated, whereas preservation-oriented graphs tend to show stronger modular or clustered organization. We do not use these graph properties as the primary evidence for the paper's claims, because they depend on visualization and graph-construction choices. Instead, they serve as a complementary perspective that is consistent with the main localization, representation, and intervention results.

\end{document}